\documentclass{article} %
\usepackage{iclr2027_conference,times}

\usepackage{amsmath,amsfonts,bm}

\def\eqref#1{equation~\ref{#1}}

\def\1{\bm{1}}

\DeclareMathAlphabet{\mathsfit}{\encodingdefault}{\sfdefault}{m}{sl}
\SetMathAlphabet{\mathsfit}{bold}{\encodingdefault}{\sfdefault}{bx}{n}

\usepackage{hyperref}
\hypersetup{pdftitle={Capability Scaling-Down Laws for LLM Compression},
  pdfauthor={Xueqi Cheng, Liang Wu, Kelly Wan, Liangjie Hong, Yushun Dong}}
\usepackage{url}
\usepackage{amsmath,amssymb,amsthm}
\usepackage{booktabs}
\usepackage{tabularx}
\usepackage{multirow}
\usepackage{microtype}
\usepackage{graphicx}
\usepackage[font=small,labelfont=bf]{caption}
\usepackage{subcaption}
\usepackage{xcolor}
\usepackage{enumitem}
\usepackage{rotating}
\usepackage{array}
\usepackage{placeins}
\usepackage{float}
\renewcommand{\topfraction}{0.95}\renewcommand{\bottomfraction}{0.95}
\renewcommand{\textfraction}{0.05}\renewcommand{\floatpagefraction}{0.92}
\newcommand{\nSeries}{5}
\newcommand{\nFamilies}{4}

\newcommand{\preliminary}[1]{\footnote{Preliminary evidence: #1.}}
\usepackage{etoolbox}
\AtBeginEnvironment{tabular}{\raggedright\let\\\tabularnewline}

\newcommand{\dw}{\delta w}

\newcommand{\gbar}[1]{\bar{g}_{#1}}
\newcommand{\Lc}[1]{L_{#1}}

\newcommand{\probeset}[1]{\mathcal{P}_{#1}}

\newcommand{\dens}{d}
\newcommand{\bits}{b}

\ExplSyntaxOn
\tl_new:N \l__split_old_tl
\tl_new:N \l__split_new_tl
\cs_new_protected:Npn \splitReadTable #1
  {
    \file_get:nnN {#1.tex} {} \splitTable
    \tl_set:Nx \splitTable {\tl_to_str:N \splitTable}
  }
\cs_new_protected:Npn \retargetreference #1#2
  {
    \tl_set:Nx \l__split_old_tl {\tl_to_str:n {#1}}
    \tl_set:Nx \l__split_new_tl {\tl_to_str:n {#2}}
    \exp_args:NNV \tl_if_in:NnTF \splitTable \l__split_old_tl
      {\exp_args:NNVV \tl_replace_all:Nnn \splitTable \l__split_old_tl \l__split_new_tl}
      {\PackageError{appendix-split}{Reference~patch~failed}{Check~the~input~table.}}
  }
\cs_new_protected:Npn \splitUseTable
  {
    \tl_set_rescan:Nno \splitTable {} {\splitTable}
    \splitTable
  }
\ExplSyntaxOff
\newenvironment{retargetreferences}[1]{%
  \begingroup
  \renewcommand{\input}[1]{%
    \splitReadTable{##1}%
    #1%
    \splitUseTable
  }%
}{\endgroup}

\title{Capability Scaling-Down Laws for\\ LLM Compression}

\author{Xueqi Cheng$^{1}$, Liang Wu$^{2}$, Kelly Wan$^{2}$, Liangjie Hong$^{2}$, Yushun Dong$^{1}$ \\
\textnormal{$^{1}$Florida State University \quad \texttt{\{xc25,yd24f\}@fsu.edu}} \\
\textnormal{$^{2}$Nokia Applied Research \quad \texttt{\{liang.wu,kelly.wan,liangjie.hong\}@nokia.com}}
}

\iclrfinalcopy %
\begin{document}

\maketitle
\lhead{Preprint}
\vspace{-12pt}

\begin{abstract}
LLM compression reduces inference costs and memory requirements, but selecting a method and configuration remains largely empirical because comparable resource reductions can produce different capability losses. We present a systematic investigation of capability scaling-down laws for LLM compression across pruning, quantization, and distillation. Our framework measures capability loss in mathematics, code generation, and question answering, and relates these measurements to model size, training stage, compression settings, data availability, and training exposure. We develop simple predictive relations and evaluate their accuracy, measurement efficiency, and generalization to unseen configurations and model states. The results identify response patterns that can be shared across capabilities and differences that require separate treatment. Sharing the density response across pruning levels halves the configuration measurements needed to fit a pruning predictor: on new Pythia states, on pre-registered OLMo-2 test states and under Wanda pruning, the compact relation matches a regression fitted with all measurements on math and code to within 0.020 nats per token, with coefficients refitted for each setting. Controlled distillation experiments show that the cost of heavy data reuse recurs across question-answering distributions, while the net benefit depends on the evaluation distribution. We further evaluate the decision value of these predictions by comparing numerical selection with configuration medians and fixed method priorities. Independent evaluations across two model families show that selection captures most of the available cross-method benefit for question answering within the tested candidate sets, where a fixed method priority attains the same regret, with smaller opportunities for mathematics and code. These results clarify the predictive scope of capability scaling-down laws and their use in compression method selection. Our code is publicly available at:
\url{https://github.com/LabRAI/scaling_down_law}.
\end{abstract}

\section{Introduction}
\label{sec:intro}

LLM compression aims to reduce inference costs and memory requirements while preserving the capabilities needed for deployment. These requirements vary across applications: software engineering agents rely on code generation \citep{jimenez2024swebench,yao2023react,schick2023toolformer,liu2024agentbench}, whereas document question-answering systems require reasoning over retrieved evidence \citep{yang2018hotpotqa,ho2020constructing,trivedi2022musique}. Domain applications narrow them further, as in LLM-assisted seizure diagnosis \citep{li2026llm} and typhoon track forecasting \citep{li2025typhoformer}, or energy systems, where a Mamba state space model enhanced with graph neural networks predicts consumption from the grid topology and temporal dynamics \citep{yu2026energymamba} and diffusion guided by anomaly semantics generates synthetic consumption data \citep{jiang2026synenergy}. However, the selection and configuration of compression methods, including pruning \citep{lecun1989optimal,frantar2023sparsegpt,sun2023simple}, quantization \citep{dettmers2022llmint8,frantar2022gptq,lin2023awq}, and distillation \citep{hinton2015distilling,gu2023minillm,agarwal2023gkd}, remain largely empirical because comparable resource reductions can produce substantially different capability losses. Identifying configurations that satisfy workload-specific requirements therefore often entails repeated compression and evaluation of every candidate.

Predictive compression laws seek to reduce this evaluation burden by estimating post-compression loss from model characteristics and intervention settings \citep{frantar2023scaling,kumar2024scaling,panferov2025unified,zhang2024scaling,busbridge2025distillation}. Recent laws resolve individual tasks for one method at a time \citep{sengupta2025compression,zhou2026task}, and complementary studies document heterogeneous compression effects for individual tasks and quantizers \citep{jaiswal2023compressing,liu2025quantization}. Less is known about how many measurements a capability-level prediction needs, which parts of a fitted relation carry over to another model family or compression algorithm, and when a numerical loss prediction changes the choice among compression methods. A configuration that preserves overall performance may still degrade a required capability, and existing predictions offer limited guidance on which method and level will satisfy a workload.

In this paper, we systematically investigate how compression response structure can reduce the measurements needed to develop capability scaling-down laws. Using a common framework for pruning, quantization, and distillation, we examine which aspects of the response can be shared across configurations and which depend on model state, training conditions, or evaluation distribution. We then evaluate the resulting relations through independent tests of predictive accuracy, measurement efficiency, and transfer. Finally, we assess when numerical loss prediction provides additional value for compression selection beyond simple method preferences.

Overall, our contributions are summarized below:
\begin{itemize}
\item \textbf{A systematic investigation of capability scaling-down laws.} A common framework for pruning, quantization, and distillation examines how model state, compression configuration, training data, and evaluation distribution shape capability loss and its prediction.
\item \textbf{Measurement-efficient predictive relations.} Sharing the response structure across pruning densities halves the configuration measurements needed to fit a predictor, within 0.020 nats per token of the full-grid regression on math and code, and the saving reproduces on a second model family and under a second pruning criterion: the predictive form and fitting strategy carry over, while the coefficients are estimated for each setting.
\item \textbf{Compression selection and decision analysis.} Independent selection tests separate the value of comparing methods from the additional value of numerical loss prediction: on question answering a stable method priority attains the regret of numerical selection, and configuration-level estimates add a smaller benefit on mathematics.
\end{itemize}

\section{Problem Formulation and Experimental Design}
\label{sec:setup}

\subsection{Capability Loss and Prediction Inputs}
\label{sec:loss}

We measure mathematical reasoning (Math), code generation (Code) and multi-hop question answering
(QA) through fixed probe sets $\probeset{c}$ drawn once from held-out splits of MATH-500
\citep{hendrycks2021math,lightman2023verify}, MBPP \citep{austin2021program} and 2WikiMultihopQA
\citep{ho2020constructing}; each fixes one evaluation distribution, implied in $L_c$. The observable
is the
token-weighted conditional cross-entropy of the reference completion for a model with parameters $w$,
\begin{equation}
\Lc{c}(w)\;=\;\frac{1}{T_c}\sum_{(x,y)\in\probeset{c}}\;\sum_{t=1}^{|y|}
-\log p_w\!\left(y_t\mid x,y_{<t}\right),\qquad
T_c=\sum_{(x,y)\in\probeset{c}}|y|,
\label{eq:taskloss}
\end{equation}
with prompt tokens masked and the same 64 probes per capability within a model. The reference model
$M_0$ is the checkpoint before the intervention, and responses are within-model differences in
native nats per token: the change in capability loss $\Delta L_c=L_c(\text{compressed})-L_c(M_0)$ for pruning and
quantization, and for distillation the signed transfer response
$\delta_c=L_c(S_{\mathrm{KD}})-L_c(S_0)$ of a trained student against the initial student $S_0$ it
started from, not against the teacher. Absolute levels from different tokenizers are never placed on one scale. We use capability loss to denote reference-completion cross-entropy on an evaluation distribution associated with a capability; this continuous endpoint supports the study of compression responses and predictive relations, and task accuracy and execution success are evaluated separately.

\subsection{Models and Compression Methods}
\label{sec:models}

We study pruning, quantization, and distillation through controlled implementations that expose their main intervention variables: weight density, numerical precision and group size, and training data and exposure. Pythia checkpoints \citep{biderman2023pythia} provide a setting in which model size and training stage can be varied separately, while Gemma students support controlled comparisons of data availability and reuse. We use additional model families, a second pruning criterion, and independent evaluation distributions to examine which findings extend beyond these initial settings.

The implementations are fixed by protocol (Appendix~\ref{sec:app_details}): global magnitude pruning
to a retained density $d$, with Wanda \citep{sun2023simple} as the second criterion; symmetric round-to-nearest quantization per output channel at bit-width
$b$, or with groups of $g$ weights sharing one scale; and LoRA fine-tuning \citep{hu2022lora} of an
existing student on teacher traces. Pythia development states span three sizes at three training stages; held-out Pythia states up to 6.9B and OLMo-2 checkpoints \citep{olmo2025olmo2} test transfer, and a heterogeneous panel of twelve public models from 0.6B to 32B establishes the phenomena of Section~\ref{sec:observed}. The inputs admitted to each predictor are the parameter count $N_0$ of the transformer weight
matrices of $M_0$, the pretraining token count $D_0$ where disclosed, and the reference capability
losses $L_{0,c}=L_c(M_0)$ measured once; for distillation the initial student plays this role. Each prediction adds a predicted change to the reference loss,
\begin{equation}
\widehat L_{m,c}\;=\;L^{\mathrm{obs}}_{0,c}\;+\;
\widehat\Delta_{m,c}(\mathbf x,\boldsymbol\theta_m),\qquad
\mathbf x=(N_0,D_0,L_{0,c}),
\label{eq:interface}
\end{equation}
with $\boldsymbol\theta_m$ the setting of method $m$: density, bit-width and group size, or pool
and budget.

\subsection{Fitting and Evaluation}
\label{sec:eval}

Each relation is fit on development states and settings only and judged on unseen settings and
states, with forms, coefficients and predictions fixed beforehand; each prediction keeps the
status it was given, whatever the outcome. Error is the mean absolute error of $\widehat\Delta$ in nats per token per capability on the same
cells as the same-input alternatives and the source-free references, and the strongest baseline chosen inside the development folds is the comparator.
Two families of predictor recur: a \emph{source-conditioned} relation reads descriptors of the
initial model state, a \emph{source-free} one predicts the response from the compression
configuration alone; both predict a change to which the reference loss $L_{0,c}$ is added, so
source-free means independent of the initial state, not free of information about it.
This design separates variation in the initial model state from variation in the compression
configuration, which is what lets Section~\ref{sec:observed} attribute a response to one or the other;
Appendix~\ref{sec:app_registration} lists the revision at which each prediction was fixed, and the fold rules.

\section{Empirical Analysis of Capability Loss}
\label{sec:observed}

This section describes the response structure that the predictors of Section~\ref{sec:laws} build
on, measured on the heterogeneous panel and the controlled Pythia and Gemma panels of
Section~\ref{sec:models}. Section~\ref{sec:resp_capabilities} shows which response shapes pruning and
quantization share across capabilities and model states, and Section~\ref{sec:why} shows how training
data, exposure and the evaluation distribution shape the response to distillation and what that asks
of its predictors.

\subsection{Responses to Pruning and Quantization}
\label{sec:resp_capabilities}\label{sec:state_config}

\begin{figure}[t]
\centering
\includegraphics[width=\textwidth]{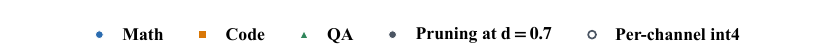}\\[1pt]
\includegraphics[width=\textwidth]{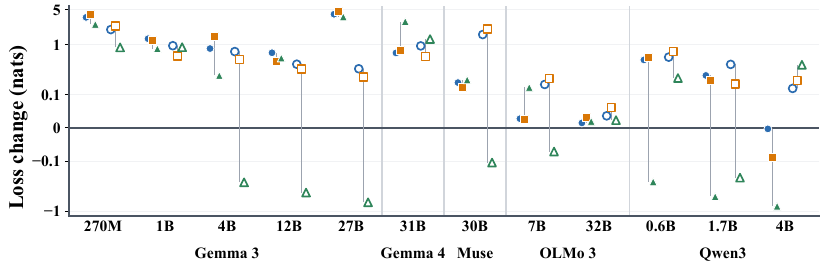}
\caption{Capability loss change from each model's dense loss across twelve public models, under pruning to $d=0.7$ (filled markers, left in each model's slot) and per-channel int4 (open markers, right), on a symmetric logarithmic axis that is linear within $\pm0.1$ nats; grey bars span each model's three responses under each method. The ordering of capabilities changes across series, and at the same setting the magnitudes differ by two orders of magnitude across all twelve models.}
\label{fig:heterogeneity}
\end{figure}

\begin{figure}[t]
\centering
\includegraphics[width=\textwidth]{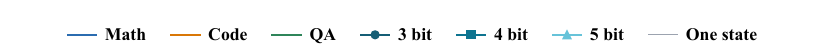}\\[1pt]
\begin{subfigure}[t]{1.8in}\centering\includegraphics[width=\linewidth]{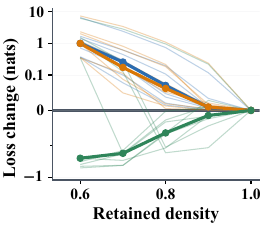}\caption{Pruning}\end{subfigure}\hfill
\begin{subfigure}[t]{1.8in}\centering\includegraphics[width=\linewidth]{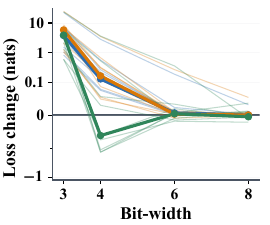}\caption{Per-channel quantization}\end{subfigure}\hfill
\begin{subfigure}[t]{1.8in}\centering\includegraphics[width=\linewidth]{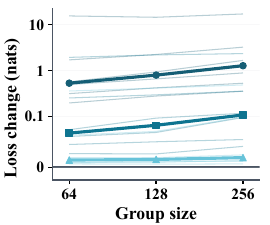}\caption{Grouped quantization}\end{subfigure}
\caption{Compression responses on the controlled Pythia panel, the structure the predictors of Section~\ref{sec:laws} build on: thin lines are the development states and bold lines their median, on a symmetric logarithmic axis that is linear within $\pm0.1$ nats. (a) Loss change against retained density on the nine development states: mathematics and code rise along one shape with a state-dependent scale, while question answering follows no shared curve. (b) Loss change against bit-width under per-channel round-to-nearest. (c) Mathematics loss change against group size at 3, 4 and 5 bits on the six grouped-quantization development states.}
\label{fig:pythia_responses}
\end{figure}

Compression changes capabilities unevenly and in no fixed order (Fig.~\ref{fig:heterogeneity}), which is
the first structure a predictor has to capture. On the controlled Pythia panel, mathematics and code
share one shape in density with a capability-specific scale, whereas the question-answering response
changes sign across density at every state (Fig.~\ref{fig:pythia_responses}). The two
compression axes also carry different shapes. Under pruning, once the measured first-order alignment
term is subtracted, the residual before the cliff is a power law in the deleted capability mass, with an
exponent that transfers across families. Under quantization the response is instead a threshold in bit
width whose position depends on the group size, so an interpolation over the measured grid predicts it
where a smooth surface does not. Fig.~\ref{fig:pythia_responses} shows both shapes on the development
states: pruning losses rise steeply once density falls below 0.8, per-channel losses are large at 3 bits
and small from 4 bits on, and at 3 bits the grouped response grows with the group size while at 4 and 5
bits it stays near zero. Both accounts read measurements of the compressed model, so they explain rather than predict
(Appendix~\ref{sec:app_mech}).

Model state sets the scale of the pruning response for mathematics and code, and the quantization response depends on the full configuration. On Pythia checkpoints whose size and training stage vary independently, parameter count and
pretraining tokens enter the pruning amplitude for mathematics and code and carry those capabilities to unseen densities of states that
entered the fit; the same inputs over-predict question answering, whose density response is not
proportional to any shared curve. A per-setting regression on parameter count, pretraining tokens and reference loss
reduces the mean absolute error from 0.41 to 0.25 nats for mathematics and from 0.52 to 0.29
for code, and raises it from 0.50 to 0.81 for question answering. Under quantization the configuration acts through two axes at once, the threshold in bit width
moving with the group size, so a relation needs the configuration in full and the initial state only
where it helps. Because the pruning shape is shared across states, a predictor can learn a state's scale
from a few densities and take the shape from the others, which is the measurement saving that
Section~\ref{sec:efficiency} tests.

\subsection{Effects of Training Data and Evaluation Distribution}
\label{sec:why}

\begin{figure}[t]
\centering
\includegraphics[width=\textwidth]{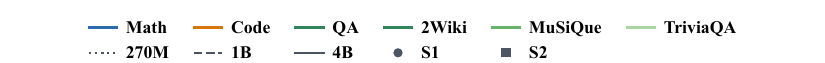}\\[1pt]
\begin{subfigure}[t]{1.35in}\centering\includegraphics[width=\linewidth]{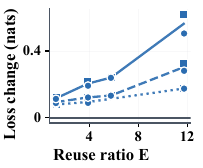}\caption{Math}\label{fig:resp_a}\end{subfigure}\hfill
\begin{subfigure}[t]{1.35in}\centering\includegraphics[width=\linewidth]{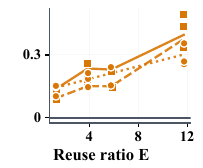}\caption{Code}\label{fig:resp_b}\end{subfigure}\hfill
\begin{subfigure}[t]{1.35in}\centering\includegraphics[width=\linewidth]{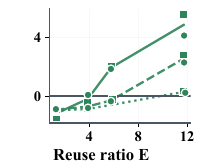}\caption{QA on 2Wiki}\label{fig:resp_c}\end{subfigure}\hfill
\begin{subfigure}[t]{1.35in}\centering\includegraphics[width=\linewidth]{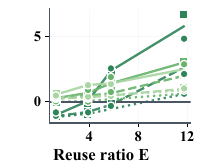}\caption{QA, three distributions}\label{fig:resp_d}\end{subfigure}
\caption{Distillation transfer responses under one protocol: loss change in nats per native token
from each student's initial state against the reuse ratio $E=T/D_U$, at the matched budget nearest
200k supervised tokens, each panel on its own scale; dotted 270M, dashed 1B, solid 4B, markers the
two pool seeds. Panel (d) rescores the same trajectories at that budget on three question-answering distributions:
a fresh 2Wiki sample disjoint from the probe, MuSiQue and TriviaQA.}
\label{fig:responses_v2}
\end{figure}

The cost of heavy data reuse recurs across question-answering distributions, while the net benefit of
distillation depends on the evaluation distribution. At a matched budget $T$ near 200k supervised tokens
(Fig.~\ref{fig:responses_v2}), a reuse ratio $E=T/D_U$ near twelve costs question answering $+0.29$,
$+2.55$ and $+4.85$ nats on students of 100M, 700M and 3.2B non-embedding parameters, and nine times
more independent data converts the same budget into a gain near one nat on each. Rescored on a fresh
2Wiki sample, on MuSiQue and on TriviaQA at the same budget, all three distributions worsen at high
reuse and improve as reuse falls, but only the fresh 2Wiki sample crosses into a gain. The reuse response also fixes an empirical data requirement: in a pre-specified experiment of eighteen
trajectories on the 1B and 4B students (Appendix Fig.~\ref{fig:dreq}), the independent data at which a
loss constraint is met lies inside the tested pools on the fresh 2Wiki sample for both students at the
two larger budgets, and we report it as measured intervals.

The training recipe is a further condition to consider when extending these relations. With every
other setting held fixed (Appendix Fig.~\ref{fig:lr_pilot}), the loss increase on mathematics and code
grows by an order of magnitude between learning rates of $5\times10^{-5}$ and $1\times10^{-4}$; at the
lower rate the 1B student is almost unharmed on those two capabilities and question answering improves
by $0.79$ nats, so the responses of Fig.~\ref{fig:responses_v2} are measured under one fixed recipe.
These responses motivate predictors that account jointly for data availability and training exposure,
with student state and evaluation distribution determining their range of transfer.

\section{Capability Scaling-Down Laws}
\label{sec:laws}

Each predictor is built from the response structure of Section~\ref{sec:observed}: the observed
structure fixes the form, the inputs of Section~\ref{sec:models} fill it, and a validation target
decides how Section~\ref{sec:general} judges it. The observed structure differs by method: pruning
shares one density shape across states, quantization acts through the full configuration, and
distillation separates independent data, training exposure and evaluation distribution. The initial
state enters a relation only for capabilities where it lowered development error
(Section~\ref{sec:state_config}); elsewhere the relation predicts the response from the
configuration alone.

\subsection{Pruning}
\label{sec:law_prune}

Mathematics and code share one density shape whose scale depends on the model state
(Section~\ref{sec:state_config}), so we write their response as a state-dependent amplitude times a
density term shared across states,
\begin{equation}
\widehat{\Delta L}_c(\mathbf x,d)=A_c(\mathbf x)\,\Big(\frac{1-d}{0.3}\Big)^{\gamma_c},
\qquad
A_c(\mathbf x)=\beta_c\cdot\big(1,\tilde{\log N_0},\tilde L_{0,c},\tilde{\log D_0}\big),
\label{eq:power}
\end{equation}
with five parameters per capability and inputs standardized on the development set. Because the
density term is shared, a development grid can keep every state while measuring fewer densities per
state, and the validation target is the error at half the development measurements against the
full-grid regression on states in no fit (Section~\ref{sec:efficiency}). The form is fitted on Pythia
160M--1.4B at three pretraining stages over densities 0.6--0.9 and holds at new stages of sizes inside
that range; Appendix~\ref{sec:app_selbound} gives its selection among forms of the same complexity. At sizes outside it, and for question answering, whose density response follows no
shared curve, the predictor is the pre-specified development median curve
$L_{0,c}+\widetilde f_c(d)$, which reads only the reference loss and which the selection rule of
Section~\ref{sec:selection} calls on every new state. On another family or pruning criterion the form
and its fitting strategy are reused and the coefficients are estimated anew (Section~\ref{sec:efficiency}).

\subsection{Quantization}
\label{sec:law_quant}

Under quantization the response is a threshold in bit width whose position moves with the group size
(Section~\ref{sec:resp_capabilities}), so the predictors work on the measured configuration grid
rather than on a smooth surface. Per-channel quantization is predicted per bit width, by the per-bit
source regression on states in the fit and the per-bit median on new ones. Grouped quantization adds
the group size $g$ as a second axis: for mathematics and code on seen states the relation is a
piecewise interpolation over the measured $(b,g)$ grid, linear in $\log_2 g$, and otherwise it is the
per-configuration development median. Both are fitted on the Pythia states over 3--8 bits and groups
of 32--512 weights and validated at unseen group sizes of seen states and on a new state
(Section~\ref{sec:unseen_settings}). Here a complete description of the
configuration and coverage of its grid matter more than a parametric form, and the configuration
medians are the strong baseline that any quantization relation has to beat.

\subsection{Distillation}
\label{sec:law_distill}

Predicting the loss after distillation requires separating the amount of independent data, the
cumulative training exposure and the evaluation distribution (Section~\ref{sec:why}). The teacher
supplies the training targets and is fixed within the evaluated protocol, and the relations read
supervised completion tokens $T$, unique pool tokens $D_U$ and their ratio to predict the signed
transfer response of Section~\ref{sec:loss} as
\begin{equation}
\widehat\delta_c=a_c\log(1+E)\ (\text{math, code}),\qquad \widehat\delta_{\mathrm{QA}}=u\,(a+b\,u+q\,v),
\label{eq:kdforms}
\end{equation}
with $u=\log(1+T/T_{\mathrm{ref}})$ and $v=\log(D_U/D_{\mathrm{ref}})$ for reference constants fixed on
the development set: one parameter for mathematics and code, which vanish at zero budget, and three for question
answering. Pre-specified structural checks neither reject nor establish an additive joint form in
budget and pool (Appendix~\ref{sec:app_selbound}), so the forms rest on the development comparison:
leave-one-run-out selected the logarithm of the reuse ratio for mathematics and code and the joint
budget and pool form only for question answering (Appendix Table~\ref{tab:distill_confirm}), a local
approximation at the tested budgets and pools. The validation target is new pools and budgets of the
development students (Section~\ref{sec:unseen_settings}). The relations describe the Gemma-3 270M and
1B students under one recipe, with pools to 158k tokens, budgets of 50k--200k and question answering
scored on 2Wiki. There question answering improves under both training recipes tested while mathematics
and code move by hundredths to tenths of a nat; at high reuse and on the other question-answering distributions the response
reverses (Section~\ref{sec:why}), and the 4B student and reuse near fourteen lie outside this domain.

\section{Predictive Performance and Generalization}
\label{sec:general}

\subsection{Measurement Efficiency}
\label{sec:efficiency}

\begin{figure}[t]
\centering
\includegraphics[width=\textwidth]{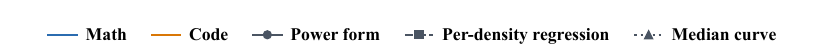}\\[1pt]
\begin{subfigure}[t]{1.8in}\centering\includegraphics[width=\linewidth]{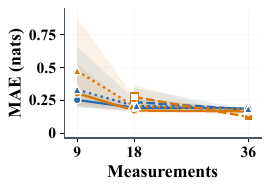}\caption{Pythia, error against budget}\end{subfigure}\hfill
\begin{subfigure}[t]{1.8in}\centering\includegraphics[width=\linewidth]{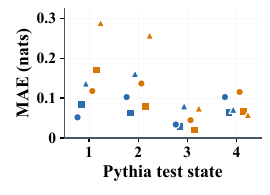}\caption{Pythia, four new states}\end{subfigure}\hfill
\begin{subfigure}[t]{1.8in}\centering\includegraphics[width=\linewidth]{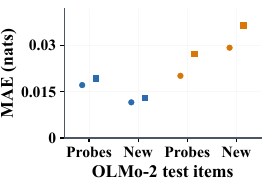}\caption{OLMo-2, half against full}\end{subfigure}
\caption{Measurement efficiency of the compact pruning form on mathematics and code, in nats per native
token: error on new Pythia states against development measurements per capability; error per state in
the pre-specified Pythia confirmation (states 1 to 4: 160M at step 80k, 410M at 112k, 1.4B and 1B at
48k), the power form at half budget and the others at full budget; and, on four OLMo-2 test states, the
form fitted on 12 configurations (circles) against the per-density regression fitted on 24 (squares).}
\label{fig:efficiency}
\end{figure}

Sharing response structure across densities reduces the measurements needed to fit the pruning predictor while retaining coverage of model states. This finding is reproduced in three evaluations (Table~\ref{tab:efficiency-body}, Fig.~\ref{fig:efficiency}): magnitude pruning on Pythia, magnitude pruning on OLMo-2, and Wanda pruning on Pythia. With coefficients refitted on each development panel, the compact form uses half the configuration measurements and yields math and code errors within 0.020 nats per token of the full-grid regression. The reduced grids require 59\% of the full-grid GPU time on OLMo-2 and 70\% under Wanda, including reference measurements, loading, and calibration. These results support reuse of the predictive form and fitting strategy, with coefficients estimated for each setting.
\begin{table}[tb]
\normalfont
\centering
\small
\caption{Measurement efficiency of the compact pruning form in three evaluations, mean absolute error in nats per token on four test states in no fit. The half grid keeps every development state at half the densities, 18 of 36 configurations per capability on Pythia and 12 of 24 on OLMo-2; the full-grid per-density regression is the pre-specified comparator. On OLMo-2 and under Wanda every form was pre-specified before the test; on the Pythia magnitude panel the forms other than the compact form are retrospective refits.}
\label{tab:efficiency-body}
\begin{tabular*}{\textwidth}{@{\extracolsep{\fill}}>{\raggedright\arraybackslash\hspace{0pt}}p{\dimexpr 0.270000000\textwidth-3.780000000\tabcolsep\relax}>{\raggedright\arraybackslash\hspace{0pt}}p{\dimexpr 0.090000000\textwidth-1.260000000\tabcolsep\relax}>{\raggedright\arraybackslash\hspace{0pt}}p{\dimexpr 0.106700000\textwidth-1.493800000\tabcolsep\relax}>{\raggedright\arraybackslash\hspace{0pt}}p{\dimexpr 0.106700000\textwidth-1.493800000\tabcolsep\relax}>{\raggedright\arraybackslash\hspace{0pt}}p{\dimexpr 0.106700000\textwidth-1.493800000\tabcolsep\relax}>{\raggedright\arraybackslash\hspace{0pt}}p{\dimexpr 0.106700000\textwidth-1.493800000\tabcolsep\relax}>{\raggedright\arraybackslash\hspace{0pt}}p{\dimexpr 0.106600000\textwidth-1.492400000\tabcolsep\relax}>{\raggedright\arraybackslash\hspace{0pt}}p{\dimexpr 0.106600000\textwidth-1.492400000\tabcolsep\relax}@{}}
\toprule
& & \multicolumn{2}{>{\centering\arraybackslash\hspace{0pt}}p{\dimexpr 0.213400000\textwidth-0.987600000\tabcolsep\relax}}{Pythia, magnitude} & \multicolumn{2}{>{\centering\arraybackslash\hspace{0pt}}p{\dimexpr 0.213400000\textwidth-0.987600000\tabcolsep\relax}}{OLMo-2, magnitude} & \multicolumn{2}{>{\centering\arraybackslash\hspace{0pt}}p{\dimexpr 0.213200000\textwidth-0.984800000\tabcolsep\relax}@{}}{Pythia, Wanda} \\
\cmidrule(lr){3-4}\cmidrule(lr){5-6}\cmidrule(lr){7-8}
Form & Grid & Math & Code & Math & Code & Math & Code \\
\midrule
Compact power form & Half & 0.073 & 0.104 & 0.017 & 0.020 & 0.023 & 0.026 \\
Quadratic strength form & Half & 0.115 & 0.142 & 0.026 & 0.045 & 0.026 & 0.048 \\
Strength only & Half & 0.098 & 0.166 & 0.018 & 0.029 & 0.031 & 0.046 \\
Median density curve & Half & 0.145 & 0.212 & 0.026 & 0.039 & 0.029 & 0.046 \\
Per-density regression & Half & 0.108 & 0.136 & 0.025 & 0.035 & 0.024 & 0.032 \\
\midrule
Per-density regression & Full & 0.060 & 0.084 & 0.019 & 0.027 & 0.025 & 0.027 \\
\bottomrule
\end{tabular*}
\end{table}

Three conditions qualify this finding. The reduction is in density measurements per state: every
reduced grid keeps all development states, so the saving comes from sharing the density response
rather than from dropping states; at 9 measurements the per-density regression cannot be fitted at all
(Fig.~\ref{fig:efficiency}a). In a retrospective subsampling of the development panel, the compact
form's median error falls from 0.25 to 0.20 nats on mathematics and from 0.30 to 0.17 on code between 9
and 18 measurements and barely moves between 18 and 36, so half the grid already sits where further
measurements stop paying (Appendix Table~\ref{tab:a9}). Where the response is near zero, complex prediction does not beat zero
change: under Wanda the 1B and 1.4B test states move by at
most 0.06 nats and no predictor improves on zero change there (Appendix~\ref{sec:app_a11}). The errors
are compared on independent panels and are not tests of statistical equivalence. Among forms of
similar complexity fitted on the same half grid, the compact form has the lowest mathematics and code
error in all three evaluations. Three kinds of transfer are thus separated: the research
method, with its inputs, measurement budget and independent validation, applies to all three
compression methods; the predictive form and fitting strategy reproduce across two families and two
pruning criteria; and the coefficients are estimated for each setting, since without refitting the
Pythia coefficients err by 0.4 to 1.9 nats on OLMo-2 and the magnitude-pruning coefficients by 0.19 to
0.27 nats under Wanda.

\subsection{Prediction at Unseen Configurations and Model States}
\label{sec:unseen_settings}\label{sec:unseen_states}

\begin{table*}[t]\normalfont
\centering\footnotesize\fontsize{8}{9.5}\selectfont\linespread{0.85}\selectfont
\setlength{\tabcolsep}{1.5pt}
\renewcommand{\arraystretch}{0.88}
\setlength{\abovecaptionskip}{4pt}
\caption{\footnotesize\fontsize{8.5}{10}\selectfont Prediction at equal development budget. Test: configurations or states unseen by the relation. Delivered relation: predictor recommended on all evidence; a dagger marks selection after seeing test results. Error: test mean absolute error in nats per token, per capability (QA: question answering). Gain: strongest development baseline error minus delivered error; positive favours the delivered relation. Baselines are chosen inside the development folds and scored on the same cells. $n$, on each method line: development configuration measurements per capability. Appendix Table~\ref{tab:main-prediction-candidates} names baselines and pre-specified candidates; Fig.~\ref{fig:generalization} compares each row with its baseline.}
\label{tab:main-prediction-v2}
\begin{tabular*}{\textwidth}{@{\extracolsep{\fill}}llrrrrrr@{}}
\toprule
& & \multicolumn{3}{c}{Error} & \multicolumn{3}{c}{Gain over baseline} \\
\cmidrule(lr){3-5}\cmidrule(lr){6-8}
Test & Delivered relation & Math & Code & QA & Math & Code & QA \\
\midrule
\multicolumn{8}{@{}l@{}}{\textit{Pruning, $n=84$}} \\
Three new checkpoints & Median density curve$^{\dagger}$ & 0.277 & 0.214 & 0.221 & $-$0.047 & $+$0.008 & $+$0.000 \\
\midrule
\multicolumn{8}{@{}l@{}}{\textit{Quantization, $n=54$}} \\
Group sizes 32 and 512 & Interpolation; median for QA$^{\dagger}$ & 0.065 & 0.116 & 0.458 & $+$0.268 & $+$0.562 & $+$0.000 \\
A new 1.4 billion stage & Development median$^{\dagger}$ & 0.088 & 0.153 & 0.141 & $+$0.262 & $+$0.410 & $+$0.000 \\
\midrule
\multicolumn{8}{@{}l@{}}{\textit{Distillation, $n=100$}} \\
Six new pools, 270 million student & Reuse; budget and pool for QA & 0.074 & 0.019 & 0.515 & $-$0.009 & $+$0.004 & $+$0.095 \\
Six new pools, 1 billion student & Reuse; budget and pool for QA & 0.057 & 0.049 & 0.463 & $-$0.023 & $+$0.005 & $-$0.013 \\
\bottomrule
\end{tabular*}
\end{table*}

At equal development budget and with every prediction fixed before measurement, the relations that
read the initial state help at unseen configurations of seen states, while on new model states the
source-free medians predict as well or better (Table~\ref{tab:main-prediction-v2}; pre-specified candidates in Appendix
Table~\ref{tab:main-prediction-candidates}; Fig.~\ref{fig:generalization} compares each task with its
baseline). In grouped quantization, with 54
development and 21 pre-specified test cells, the piecewise interpolation is the most accurate
predictor of unseen group sizes at the development states for mathematics and code, cutting the error
of the strongest development baseline from 0.333 to 0.065 nats on mathematics and from 0.678 to 0.116
on code. For distillation,
at six new pools and three new budgets, every selected form improves on zero change, with a detected
gain over the development baseline for code on both students and for question answering on the
smaller one. On three Pythia checkpoints in no fit, the pre-specified pruning power form matches the
per-density regression on mathematics with a quarter of the parameters, and trails the median on code
and every source-free curve on question answering; outside the tested density range every
source-conditioned form errs by 0.6 to 4.5 nats. At 2.8B and 6.9B, beyond the development range, the
pre-specified median curve predicts pruning and per-bit quantization more accurately (at 6.9B and step 112k,
0.05, 0.07 and 0.15 nats against 0.23, 0.19 and 0.96 for the source regression), and on a new
grouped quantization state it wins for mathematics and question answering, while the interpolation keeps its lead
on code.

\section{Compression Method Selection}
\label{sec:selection}

How precise must a loss prediction be to change a compression choice? We answer this under nominal
storage constraints by comparing, on the same candidates, selection with parametric predictions (the
source-conditioned relations of Section~\ref{sec:laws}), with empirical predictions (development
configuration medians), and with a fixed method priority that uses no numerical prediction. Given a reference model, a capability and a budget, the decision is
\begin{equation}
(m^{*},\boldsymbol\theta_m^{*})=\arg\min_{m,\;\boldsymbol\theta_m\in\mathcal F_m(B)}\;
\widehat L_{m,c}(\mathbf x,\boldsymbol\theta_m),
\label{eq:selection}
\end{equation}
where $\mathcal F_m(B)$ holds the settings of method $m$ whose nominal storage, counted in stored matrix parameters, falls within $B$. The final rule calls
the pruning and quantization relations of Section~\ref{sec:laws} on a seen state and their
source-free medians on a new one; for distillation it calls a regression on the student state for mathematics and pre-specified constants otherwise (Appendix Table~\ref{tab:locked_rule}); a workload needing several capabilities replaces $\widehat L_{m,c}$ by $\max_c[\widehat L_c(M)-L_c(M_0)]$.

\begin{figure}[t]
\centering
\includegraphics[width=\textwidth]{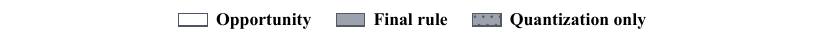}\\[1pt]
\begin{subfigure}[t]{1.35in}\centering\includegraphics[width=\linewidth]{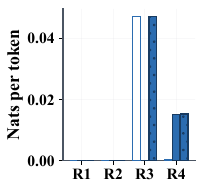}\subcaption{Mathematics}\end{subfigure}\hfill
\begin{subfigure}[t]{1.35in}\centering\includegraphics[width=\linewidth]{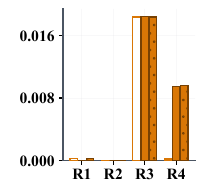}\subcaption{Code}\end{subfigure}\hfill
\begin{subfigure}[t]{1.35in}\centering\includegraphics[width=\linewidth]{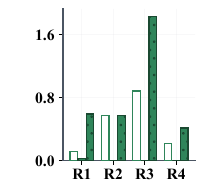}\subcaption{Question answering}\end{subfigure}\hfill
\begin{subfigure}[t]{1.35in}\centering\includegraphics[width=\linewidth]{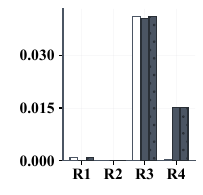}\subcaption{Largest increase}\end{subfigure}
\caption{Independent validation of the final rule on four references that supplied it no outcome
(R1--R4, Appendix Table~\ref{tab:s3-identity}): opportunity over the best quantization candidate and regret of the final rule and of quantization only, means in nats per native token over the sixteen paired budgets.}
\label{fig:rule_main}
\end{figure}

On four references that supplied it no outcome, the final rule captures most of the cross-method opportunity for question answering. Each reference carries twelve candidates at seventeen storage budgets (Appendix Table~\ref{tab:s3-identity}). Opportunity is the loss of the best feasible quantization candidate minus that of the best feasible candidate of any method, and a policy's regret is the loss of its choice minus that best candidate. On the 64 cells at which both policies find a feasible candidate the rule selects the new student in 63, with regret 0.000 nats on three references and 0.027 on the fourth against 0.41 to 1.84 for quantization only, meeting the pre-specified criterion (Fig.~\ref{fig:rule_main}); on mathematics, code and the largest loss increase the opportunity is at most 0.047 nats and the rule matches quantization within 0.0015. On 384 fresh question-answering items scored after the round, the rule's choice has a lower loss than the quantization-only choice on three references and is level on the fourth, while a generation readout puts the same students ahead on exact match on three, behind on one and no better than their initial students. Termination and the evaluation target are separate factors: a supervised end marker restores termination and raises the Gemma student's exact match, while the Pythia disagreement between loss and exact match persists (Appendix~\ref{sec:app_eos}), so a selector has to state whether it optimises loss or exact match. The quantization-only regret splits into selection error inside quantization and cross-method opportunity: on Pythia 410M its 0.59 nats are 0.47 and 0.12, and the rule's choice is 0.09 nats better than the best quantization candidate. 

A retrospective comparison of the three policies on the same candidates (Appendix
Table~\ref{tab:s3-ablation}) shows how fine a prediction selection needs. On question answering the
fixed method priority, the development medians and the final rule reach the same regret, 0.014 nats on
the Pythia pair and 0.000 on the Gemma pair, against 0.58 and 1.12 for quantization only: the benefit
rests on a stable difference between methods rather than on fine loss margins. On mathematics the
medians beat the priority by 0.024 nats on the Gemma pair, the smaller benefit that
configuration-level estimates add. The medians are read from the development panel alone, so this benefit needs no measurement
of the target model. The parametric relations improve on neither, so their contribution
lies in measurement efficiency (Section~\ref{sec:efficiency}) rather than in selection.

\section{Related Work}
\label{sec:related}

\paragraph{Scaling Laws for LLM Compression.}
Compression laws take sparsity, precision, training format, the pre-pruning loss and post-training
tokens, or the teacher loss as inputs and predict an aggregate loss
\citep{frantar2023scaling,kumar2024scaling,panferov2025unified,zhang2024scaling,busbridge2025distillation},
following pretraining laws in which loss follows power laws in scale and downstream behaviour tracks
log-likelihood
\citep{hestness2017deep,kaplan2020scaling,hoffmann2022training,schaeffer2023emergent,gadre2024language,bhagia2024establishing,lourie2025scaling},
with a fitted form that depends on the protocol \citep{li2025misfitting,besiroglu2024chinchilla}.
Closest to ours, \citet{sengupta2025compression} predict a task's post-pruning accuracy from its unpruned accuracy and the pruning ratio, transferring the exponent to unseen models with or without one calibrating measurement, and \citet{zhou2026task} fit task-stratified quantization laws that join model size, bit width, group size and calibration-set size, validated at unseen sizes. Capability-specific responses and the joint action of bit width and group size are therefore established; this paper adds how much information and how many measurements a capability-loss prediction needs, separated into unseen configurations and model-state transfer across three methods, which parts of a fitted relation carry over to another model family and pruning criterion, and the prediction's decision value.

\paragraph{Compression Effects and Method Selection.}
We study controlled implementations of pruning, post-training quantization and
distillation
\citep{lecun1989optimal,han2015deep,frankle2019lottery,frantar2023sparsegpt,sun2023simple,ma2023llmpruner,ashkboos2024slicegpt,men2024shortgpt,dettmers2022llmint8,frantar2022gptq,xiao2022smoothquant,lin2023awq,chee2023quip,shao2023omniquant,dettmers2023qlora,hinton2015distilling,sanh2019distilbert,hsieh2023distilling,mukherjee2023orca,gu2023minillm,agarwal2023gkd,xu2024survey},
whose capability losses are documented as uneven for individual tasks and quantizers, largest on
knowledge-intensive tasks and harder reasoning problems
\citep{jaiswal2023compressing,liu2025quantization,ouyang2024lowbit,hong2024decoding,sengupta2025compression}.
Distilling from a hosted model's outputs is also how model extraction attacks replicate a model,
which recent defenses detect from the distribution of API traffic \citep{liu2026embarrassingly},
divert toward low-transferability knowledge \citep{dai2026let} or answer with certified ownership
verification \citep{shen2026credit}.
Comparisons between methods measure each candidate on the target model, finding that quantization
retains more than pruning at equal ratio \citep{jaiswal2023compressing} or placing sparsity and
precision on one capacity scale \citep{panferov2025unified}, and controlled re-evaluations find that
reported efficiency gains depend on the evaluation setup \citep{zhang2026rethinking}; repeated
tokens are worth less than new ones \citep{muennighoff2023scaling}, an exchange rate our reuse response measures per capability
(Section~\ref{sec:observed}). Model routing chooses among models per query by predicting each
model's utility before seeing its answer \citep{cheng2026latentrouter}, but for compression none states
which method a model, capability and budget should receive without measuring every candidate;
Section~\ref{sec:selection} tests that decision on independent candidates.

\section{Conclusion}
\label{sec:learned}

In this paper, we present a systematic investigation of capability scaling-down laws for LLM compression across various settings. Through a common framework and controlled experiments, we characterize how capability losses vary with model state, compression configuration, and training data, and identify shared response patterns and capability differences. Building on these findings, we develop predictive relations and demonstrate that compact forms can reduce measurement requirements while maintaining useful predictive accuracy. We further introduce and independently evaluate a compression selection method that compares predicted capability losses under nominal storage constraints. Together, these contributions clarify the information needed to predict compression outcomes and provide an empirical foundation for developing capability scaling-down laws, reducing repeated evaluation, and aligning compression choices with application-specific capability requirements.

\phantomsection\label{sec:learned_end}

\section*{Reproducibility Statement}
Every number is produced by a script from a fixed result file; the public repository holds the
measurement suite, the recorded results, the table and figure generators, and tests of every printed
value. Section~\ref{sec:setup} and Appendices~\ref{sec:app_details} and~\ref{sec:app_registration} give
the protocols and when each prediction was fixed.

\section*{Ethics Statement}
The study uses public models and benchmark data, involves no human subjects or personal data, and has
no direct negative use that we foresee.

\section*{AI Use Statement}
We used generative AI tools to polish the writing and to help implement experiment code, checked by unit
tests and by regenerating every reported number from the saved results; the research questions, design,
analysis and scientific writing are the authors' own. The distillation data are teacher traces from a
commercial language model, the object of study (Section~\ref{sec:setup}); we used generative AI for no
other data, proofs or translation, and take full responsibility for the paper.

\phantomsection\label{sec:refs_start}
\bibliography{references_preprint}
\bibliographystyle{iclr2027_conference}

\appendix
\newpage
\raggedbottom
\makeatletter
\setlength{\@fptop}{0pt}\setlength{\@fpsep}{10pt plus 4pt}\setlength{\@fpbot}{0pt plus 1fil}
\makeatother
\renewcommand{\topfraction}{0.95}\renewcommand{\bottomfraction}{0.95}\renewcommand{\textfraction}{0.03}\renewcommand{\floatpagefraction}{0.6}
\setcounter{topnumber}{4}\setcounter{bottomnumber}{4}\setcounter{totalnumber}{6}
\section*{Appendix}
\suppressfloats[t]

\noindent The appendix gives the experimental settings, the fitted models and their
selection, the pre-specified prediction results, the mechanism analyses and the selection evidence
behind the main text. Complete result tables and auxiliary analyses are in the supplementary
material.

\section{Experimental Details}
\label{sec:app_details}

This appendix records the probes, the compression protocols and the controlled panel in the form
used throughout the paper, so that each later section can be read against them.

\subsection{Probes and loss computation}
\label{sec:app_probes}

Each capability is measured on a fixed probe set of prompt and reference pairs, drawn once with a
fixed seed from held-out benchmark splits that are disjoint from all training and recovery sources
used in the paper. The primary benchmarks are MATH-500 \citep{hendrycks2021math,lightman2023verify}
for mathematical reasoning, MBPP \citep{austin2021program} for code generation, and 2WikiMultihopQA
\citep{ho2020constructing} for multi-hop question answering. For each capability we draw 128 probes
and split them by index parity into an estimation half, used only for gradient and Fisher
statistics, and a measurement half of 64 probes on which each reported loss is computed, so that
mechanism statistics and evaluations do not share sampling noise. All reported results use the
64-probe measurement half unless a 512-probe re-measurement is stated.

The secondary benchmarks of the measurement checks (Appendix~\ref{sec:app_measure}) are GSM8K
\citep{cobbe2021gsm8k}, HumanEval \citep{chen2021codex} and HotpotQA \citep{yang2018hotpotqa} on the
pruning panel, where no training is involved, and SVAMP \citep{patel2021svamp}, HumanEval and
TriviaQA \citep{joshi2017triviaqa} where distilled models are scored, chosen so that no benchmark
from which distillation traces were derived is used to score a distilled model; MuSiQue
\citep{trivedi2022musique} serves as a second multi-hop distribution. A sample-level audit found no
overlap between any measurement probe and any training pool. The models are Pythia
\citep{biderman2023pythia}, Gemma 3 \citep{gemma2025gemma3} and Qwen3 \citep{qwen2025qwen3}, adapted
with LoRA \citep{hu2022lora}; the size-and-stage panel follows the observational use of released
checkpoints \citep{ruan2024observational,polo2024tinybenchmarks}.

\subsection{Compression protocols}
\label{sec:app_compress}

The three operations are fixed by protocol. Pruning is global magnitude pruning over every weight matrix of the language model, the attention and MLP projections together with the embedding and output-head matrices, which removes the smallest weights under one threshold to a target retained density $\dens$; biases and norms are untouched, and the prune scope recorded for each model is its full matrix parameter count. The global threshold is estimated from a seeded sample of
absolute weights, which lets the same procedure scale to 30B-parameter models, and each density is
applied to the same dense reference weights, so the densities are nested, and each pruned set contains the previous one.

Per-channel quantization fake-quantizes the weights per output channel with symmetric
round-to-nearest at bit-width $\bits$: integers are clamped to $[-q_{\max},q_{\max}]$ with
$q_{\max}=2^{\bits-1}-1$, the per-channel scale is the maximum absolute weight of the channel divided by
$q_{\max}$, and activations stay in bfloat16. The dense weights are restored between bit-widths.
Because the clipping range is fixed, the step size from 4 to 5 bits shrinks by $7/15$ and the
squared step ratio between adjacent bit-widths is $(q_{\max}(\bits)/q_{\max}(\bits+1))^2$ and
not a constant $1/4$. A squared step ratio becomes a loss ratio only under assumptions about the
error distribution and the local response that we do not test, so the $4^{-\bits}$ shape is one
candidate among others and not a derived reference. Nominal storage ratios are computed from
bit-widths; no memory or latency was measured.

The grouped variant of the same quantizer (Section~\ref{sec:law_quant}) splits each row of a
weight matrix into contiguous groups of $g$ input weights, pads the last group with zeros, which do
not affect the group maximum, and applies the same symmetric round-to-nearest with
$q_{\max}=2^{b-1}-1$ per group; $g$ equal to the row width reproduces the per-channel quantizer
exactly. The configurations are $b\in\{3,4,5\}$ and $g\in\{64,128,256\}$, with nominal storage
computed from $b$ and $g$ under the same convention as the per-channel case, so budgets are
comparable across the two quantizers.

Distillation, the third operation, fine-tunes students on black-box teacher traces, a frontier API teacher
answering GSM8K, HotpotQA and CodeAlpaca prompts with 600 traces per domain, under the full-trace
recipe: LoRA adapters of rank 16 and scaling 32 on the query, key, value, output, gate, up and down
projections, AdamW at learning rate $10^{-4}$ unless stated, an effective batch of 16 sequences, and
a cosine schedule with 3\% warmup. The training mode and the trainable parameter count are recorded
in each run manifest. Token accounting distinguishes processed tokens, which count each input token seen including repetitions, unique pool tokens, and supervised completion tokens. The data pool of $U$
examples per domain is either the first $U$ rows of the trace file, the historical protocol, or a
random sample drawn with a pool-sampling seed independent of the training seed; pools are
identified by the sorted hashes of their samples and pairwise overlaps are reported. The historical
schedule spanned a fixed number of epochs, so its cosine horizon varied with pool size; the
controlled matrix of Section~\ref{sec:law_distill} uses a schedule defined on the same absolute
processed-token budget for each pool, and the two protocols are not mixed in any fit or any table of this paper.

\subsection{The controlled Pythia panel}
\label{sec:app_pythia}

\begin{table}[tb]
\normalfont
\centering\footnotesize\setlength{\tabcolsep}{4pt}\renewcommand{\arraystretch}{1.15}
\caption{Cohorts of models and states, compression interventions, and study roles. Rows distinguish development cohorts, pre-specified prediction tests, and descriptive \mbox{post-hoc} measurements; each cohort is scored with the same 64 probes per capability. Densities and reuse counts are dimensionless, bit widths are in bits, and budgets are in tokens. The 2.8B Pythia checkpoints are excluded because their available revisions loaded identical weights, one state under three names.}
\label{tab:cohorts}
\begin{tabular*}{\textwidth}{@{\extracolsep{\fill}}>{\raggedright\arraybackslash\hspace{0pt}}p{\dimexpr 0.176923077\textwidth-1.061538462\tabcolsep\relax}>{\raggedright\arraybackslash\hspace{0pt}}p{\dimexpr 0.307692308\textwidth-1.846153846\tabcolsep\relax}>{\raggedright\arraybackslash\hspace{0pt}}p{\dimexpr 0.253846154\textwidth-1.523076923\tabcolsep\relax}>{\raggedright\arraybackslash\hspace{0pt}}p{\dimexpr 0.261538462\textwidth-1.569230769\tabcolsep\relax}@{}}
\toprule
Cohort & Models and states & Interventions & Role in the study \\
\midrule
Heterogeneous panel &
Twelve public checkpoints from \nSeries{} series and \nFamilies{} families: Qwen3 (0.6B, 1.7B, 4B),
Gemma-3 (270M to 27B), Gemma-4 (31B), Muse (30B), OLMo-3 (7B, 32B); two further Qwen3 checkpoints
(8B, 14B) measured prospectively at four densities. &
Global magnitude pruning at ten densities in $[0.3,\allowbreak 0.9]$; per-channel round-to-nearest quantization
at 8, 6, 5, 4 and 3 bits. &
Establishes the phenomena and the limits of cross-family transfer; screens the measurement noise. \\
\addlinespace
Controlled Pythia panel &
Development states: 160M, 410M and 1.4B at steps 16k, 64k and 143k (nine states). Held-out states:
160M, 410M and 1.4B at step 96k; 1B at steps 32k, 96k and 112k; 6.9B at steps 32k and 112k. &
Pruning at four development densities (0.9 to 0.6) with tests at 0.75, 0.65 and 0.55; quantization at
8, 6, 4 and 3 bits with a test at 5 bits. &
Development set for pruning and quantization; tests along the source axis and the configuration
axis. \\
\addlinespace
Distillation, controlled &
The nine Pythia development states as students; one fixed teacher pool of 600 traces per domain;
Low-rank adaptation adapters under one recipe. &
A fixed distillation recipe. &
Tests source transfer of the signed response $\delta_c$. \\
\addlinespace
Distillation, pools &
Gemma-3-1B student; pools of 75 and 600 traces per domain for fitting and 225 for testing; three
pool seeds and two training seeds. &
Processed-token budget $T$ and reuse ratio $E=T/D_U$. &
Tests transfer to an unseen trace pool. \\
\addlinespace
Distillation, heterogeneous &
Gemma-3 students from 270M to 12B, OLMo-3-7B and Qwen3 0.6B to 4B; two teachers accessed through an application programming interface; full traces compared with answer-only traces. &
Varied distillation recipes. &
Describes measured responses. \\
\bottomrule
\end{tabular*}
\end{table}

The panel uses the public Pythia checkpoints at sizes 160M, 410M and 1.4B and pretraining steps
16{,}000, 64{,}000 and 143{,}000, which correspond to 33.6B, 134.2B and 299.9B processed tokens at
the published batch of $1024\times2048$ tokens per step. The size coordinate $N_0$ counts transformer-matrix parameters only, the query, key, value, output and MLP matrices of each layer, computed from the architecture configuration, and the training-token coordinate is
$D_0=\text{step}\times2{,}097{,}152$. The sources added for the prospective tests, listed in Table~\ref{tab:cohorts}, entered no fit or model selection before their predictions were pre-specified, and they are reported as pre-specified tests.

Before any of these checkpoints entered a fit, the weight identity of each one was verified.
SHA-256 hashes of the safetensors blobs of all 23 Pythia revisions used, 30 unique blobs, show that
each revision with pruning or quantization measurements resolves to a distinct weight set, with one
exception: the Pythia-2.8B revisions published for steps 16k, 64k and 143k load identical learned
parameters. Those revisions were excluded from the panel, and the 2.8B repeatability check
(Appendix~\ref{sec:app_results}) is one state measured twice. The three checkpoints of the pruning confirmation panel have shard hashes that differ
from all other revisions of the same size, and their dense capability vectors differ from those of the neighbouring stages of the same size.

The parameter conventions follow from the architectures. All models in
the heterogeneous panel and the controlled panel are dense
transformers (Section S1 of the supplement); no sparsely activated model
entered any fit, so total and per-token active parameter counts
coincide throughout, and the reference-model definition of
Section~\ref{sec:loss} is exercised only on dense architectures. The laws use the transformer-matrix count $N_0$ of the decoder blocks, excluding embeddings, head, biases and norms; the pre-specified
distillation source term used the metadata parameter count of the student recorded when the predictions were fixed,
which for Gemma-3-4B includes 0.42B vision and projector parameters
and counts tied embeddings twice; these pre-specified covariates are reported as recorded when the predictions were fixed.

\begin{figure}[tb]
\centering
\includegraphics[width=\textwidth]{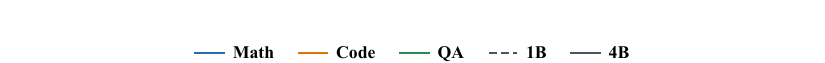}\\[1pt]
\begin{subfigure}[t]{2.7in}\centering\includegraphics[width=\linewidth]{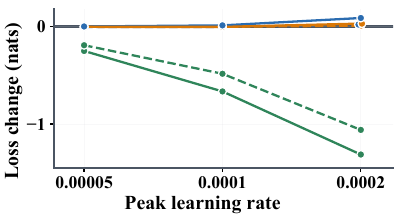}\caption{Final loss change against peak learning rate}\end{subfigure}
\caption{Learning-rate pilot for the 1B (dashed) and 4B (solid) students at one pool: final-checkpoint
loss change per capability against the configured peak learning rate; positive values indicate a
loss above that of the initial student at the same pool and budget.}
\label{fig:lr_pilot}
\end{figure}

\section{Measurement Validity}
\label{sec:app_measure}

The capability-specificity, same-capability agreement, independent multi-hop distribution, and
pre-specified eight-state scope checks are reported in Section S3 of the supplement.

The remaining two checks concern the link between loss and behavior and the choice of reference
condition. Loss-to-accuracy links were fit on
GSM8K accuracy, MBPP pass@1, and TriviaQA exact match and token F1
across seven pruned models; all four outcomes are compatible with one
cross-family sigmoid in the loss, with no resolved family-specific
term. A lower QA loss does not by itself imply better QA behavior: the
four pruned OLMo-3-32B cells whose QA loss lies below dense show no
increase in exact match. This audit covers pruning only and is not repeated for the other two methods, whose public panels
are smaller.

The reference condition was also varied. For a subset of models the
direct-answer and given-reasoning-answer conditional distributions were
measured in addition to the full-solution reference; they differ from
it for math and QA and coincide for code, where the reference is the
function body, so the numbers in this paper are specific to the
full-solution reference and should not be read as evidence of robustness to the reference format.

The loss is the conditional cross-entropy of the reference completion
given the prompt, summed over completion tokens and divided by the
number of completion tokens, as in Eq.~\ref{eq:taskloss}. Prompts and
references are tokenized with the tokenizer of each model and truncated to
1024 tokens; prompt tokens are masked out of the loss. All within-model
responses $\Delta L_c$ are differences in native-token nats
on identical token sequences. Because native-token units differ across
tokenizers, cross-family comparisons of absolute levels use per-byte
normalization or are reported separately, and no cross-tokenizer
quantity is treated as a common numerical scale. Negative responses are
retained as measured and are not clipped at zero.

\begin{figure}[!t]
\centering
\includegraphics[width=5.5in]{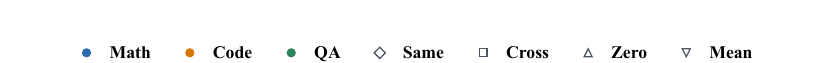}\\[1pt]
\begin{subfigure}[t]{1.8in}\centering\includegraphics[width=\linewidth]{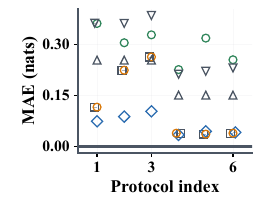}\subcaption{Math transfer MAE}\end{subfigure}\hfill
\begin{subfigure}[t]{1.8in}\centering\includegraphics[width=\linewidth]{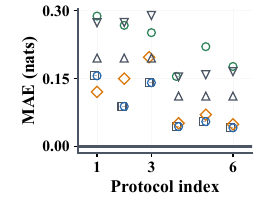}\subcaption{Code transfer MAE}\end{subfigure}\hfill
\begin{subfigure}[t]{1.8in}\centering\includegraphics[width=\linewidth]{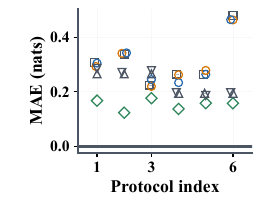}\subcaption{QA transfer MAE}\end{subfigure}\\[3pt]
\begin{subfigure}[t]{1.8in}\centering\includegraphics[width=\linewidth]{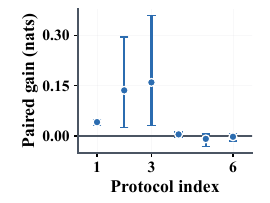}\subcaption{Math paired gain}\end{subfigure}\hfill
\begin{subfigure}[t]{1.8in}\centering\includegraphics[width=\linewidth]{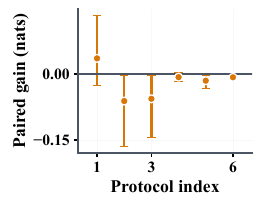}\subcaption{Code paired gain}\end{subfigure}\hfill
\begin{subfigure}[t]{1.8in}\centering\includegraphics[width=\linewidth]{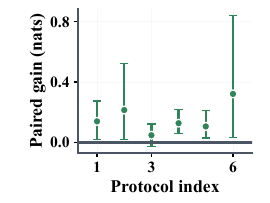}\subcaption{QA paired gain}\end{subfigure}

\caption{Measurement support for capability specificity. (a)--(c) Out-of-fold MAE of predicting the secondary-benchmark response of a capability from the primary-benchmark response of the same capability (diamond) or of another capability, against zero and train-mean controls, on the six-model, five-density pruning panel; protocols 1--3 hold out models, densities, or both, and 4--6 repeat them at $d\ge0.75$. (d)--(f) Paired same-over-cross gain with 95\% model-bootstrap intervals over models.}
\label{fig:measurement}
\end{figure}

\section{Heterogeneous-Model Results}
\label{sec:app_hetero}

This section reports the twelve-model panel that Fig.~\ref{fig:heterogeneity} summarises, first
under pruning, then under quantization and distillation, and closes with what the loss endpoint
does and does not establish about the behavior of the compressed models.

\begin{table}[tb]
\normalfont\centering\footnotesize\setlength{\tabcolsep}{3.5pt}
\caption{The table compares the pruning loss change across models and capabilities. Loss changes are in nats per native token; density is dimensionless. Ranking cells count least-affected and most-affected outcomes out of the set of densities measured for each model.}
\label{tab:panel_prune}
\begin{tabular*}{\textwidth}{@{\extracolsep{\fill}}>{\raggedright\arraybackslash\hspace{0pt}}p{\dimexpr 0.210000000\textwidth-2.940000000\tabcolsep\relax}>{\raggedright\arraybackslash\hspace{0pt}}p{\dimexpr 0.080000000\textwidth-1.120000000\tabcolsep\relax}>{\raggedright\arraybackslash\hspace{0pt}}p{\dimexpr 0.080000000\textwidth-1.120000000\tabcolsep\relax}>{\raggedright\arraybackslash\hspace{0pt}}p{\dimexpr 0.080000000\textwidth-1.120000000\tabcolsep\relax}>{\raggedright\arraybackslash\hspace{0pt}}p{\dimexpr 0.110000000\textwidth-1.540000000\tabcolsep\relax}>{\raggedright\arraybackslash\hspace{0pt}}p{\dimexpr 0.130000000\textwidth-1.820000000\tabcolsep\relax}>{\raggedright\arraybackslash\hspace{0pt}}p{\dimexpr 0.150000000\textwidth-2.100000000\tabcolsep\relax}>{\raggedright\arraybackslash\hspace{0pt}}p{\dimexpr 0.160000000\textwidth-2.240000000\tabcolsep\relax}@{}}
\toprule
Model & \multicolumn{3}{>{\centering\arraybackslash\hspace{0pt}}p{\dimexpr 0.240000000\textwidth--0.640000000\tabcolsep\relax}}{Loss change at density 0.7} & {Math at density 0.5} & Threshold density & {QA least; most affected} & Lowest QA change (density) \\
\cmidrule(lr){2-4} & Math & Code & QA & & & & \\
\midrule
Qwen3-0.6 billion & +0.49 & +0.56 & -0.26 & +4.72 & 0.60 & 3/\allowbreak 3; 0/\allowbreak 3 & -0.26 (0.70)\\
\midrule
Qwen3-1.7 billion & +0.24 & +0.19 & -0.51 & +3.60 & 0.60 & 6/\allowbreak 6; 0/\allowbreak 6 & -0.90 (0.62)\\
\midrule
Qwen3-4 billion & -0.00 & -0.09 & -0.79 & +1.31 & 0.50 & 5/\allowbreak 5; 0/\allowbreak 5 & -1.96 (0.55)\\
\midrule
Qwen3-8 billion & +0.03 & +0.10 & -0.41 & +1.03 & 0.50 & 4/\allowbreak 4; 0/\allowbreak 4 & -1.39 (0.50)\\
\midrule
Qwen3-14 billion & +0.06 & +0.03 & -0.45 & +0.67 & 0.40 & 4/\allowbreak 5; 0/\allowbreak 5 & -0.80 (0.60)\\
\midrule
Gemma-3-270 million & +3.49 & +4.04 & +2.54 & +11.21 & 0.70 & 0/\allowbreak 2; 1/\allowbreak 2 & +0.21 (0.90)\\
\midrule
Gemma-3-1 billion & +1.31 & +1.23 & +0.83 & +13.78 & 0.70 & 0/\allowbreak 5; 4/\allowbreak 5 & +0.19 (0.90)\\
\midrule
Gemma-3-4 billion & +0.83 & +1.47 & +0.24 & +7.00 & 0.60 & 1/\allowbreak 3; 2/\allowbreak 3 & +0.05 (0.90)\\
\midrule
Gemma-3-12 billion & +0.68 & +0.46 & +0.54 & +17.10 & 0.60 & 0/\allowbreak 3; 2/\allowbreak 3 & +0.07 (0.90)\\
\midrule
Gemma-3-27 billion & +4.04 & +4.69 & +3.59 & +15.21 & 0.70 & 1/\allowbreak 2; 1/\allowbreak 2 & +0.01 (0.80)\\
\midrule
Gemma-4-31 billion & +0.68 & +0.77 & +2.89 & +16.06 & 0.60 & 0/\allowbreak 3; 3/\allowbreak 3 & +0.22 (0.90)\\
\midrule
Muse-30 billion & +0.17 & +0.14 & +0.20 & +3.53 & 0.55 & 1/\allowbreak 4; 3/\allowbreak 4 & +0.02 (0.90)\\
\midrule
OLMo-3-7 billion & +0.03 & +0.03 & +0.14 & +0.22 & 0.35 & 4/\allowbreak 8; 3/\allowbreak 8 & -0.02 (0.55)\\
\midrule
OLMo-3-32 billion & +0.01 & +0.03 & +0.02 & +0.09 & 0.20 & 7/\allowbreak 10; 0/\allowbreak 10 & -0.24 (0.30)\\
\bottomrule\end{tabular*}
\par\smallskip\begin{minipage}{\linewidth}\footnotesize Threshold density is the largest measured density with $\Delta L_{\mathrm{math}}\ge 1$; the Question answering ranking columns count, over the pre-cliff densities, how often Question answering is the least and the most affected capability; the last column gives the most negative Question answering response and its density.\end{minipage}
\end{table}

\begin{table}[tb]
\normalfont\centering\footnotesize\setlength{\tabcolsep}{3.5pt}
\caption{The table compares the quantization loss change across models and capabilities. Loss changes are in nats per native token, and bit widths are in bits. The last column lists all measured widths.}
\label{tab:panel_quant}
\begin{tabular*}{\textwidth}{@{\extracolsep{\fill}}>{\raggedright\arraybackslash\hspace{0pt}}p{\dimexpr 0.202127660\textwidth-3.234042553\tabcolsep\relax}>{\raggedright\arraybackslash\hspace{0pt}}p{\dimexpr 0.127659574\textwidth-2.042553191\tabcolsep\relax}>{\raggedright\arraybackslash\hspace{0pt}}p{\dimexpr 0.085106383\textwidth-1.361702128\tabcolsep\relax}>{\raggedright\arraybackslash\hspace{0pt}}p{\dimexpr 0.085106383\textwidth-1.361702128\tabcolsep\relax}>{\raggedright\arraybackslash\hspace{0pt}}p{\dimexpr 0.085106383\textwidth-1.361702128\tabcolsep\relax}>{\raggedright\arraybackslash\hspace{0pt}}p{\dimexpr 0.085106383\textwidth-1.361702128\tabcolsep\relax}>{\raggedright\arraybackslash\hspace{0pt}}p{\dimexpr 0.085106383\textwidth-1.361702128\tabcolsep\relax}>{\raggedright\arraybackslash\hspace{0pt}}p{\dimexpr 0.085106383\textwidth-1.361702128\tabcolsep\relax}>{\raggedright\arraybackslash\hspace{0pt}}p{\dimexpr 0.159574468\textwidth-2.553191489\tabcolsep\relax}@{}}
\toprule
Model & Series & \multicolumn{3}{>{\centering\arraybackslash\hspace{0pt}}p{\dimexpr 0.255319149\textwidth-0.085106383\tabcolsep\relax}}{Loss change at 4 bits} & \multicolumn{3}{>{\centering\arraybackslash\hspace{0pt}}p{\dimexpr 0.255319149\textwidth-0.085106383\tabcolsep\relax}}{Loss change at 3 bits} & All measured bit widths \\
\cmidrule(lr){3-5}\cmidrule(lr){6-8} & & Math & Code & QA & Math & Code & QA & \\
\midrule
Qwen3-0.6 billion & Qwen3 & +0.56 & +0.73 & +0.21 & +8.83 & +10.83 & +6.61 & 8,6,5,4,3\\
\midrule
Qwen3-1.7 billion & Qwen3 & +0.40 & +0.16 & -0.21 & +9.53 & +9.46 & +8.10 & 8,6,5,4,3\\
\midrule
Qwen3-4 billion & Qwen3 & +0.13 & +0.19 & +0.39 & +10.73 & +11.85 & +5.99 & 8,6,5,4,3\\
\midrule
Qwen3-8 billion & Qwen3 & +0.28 & +0.48 & -0.80 & +12.86 & +13.55 & +9.45 & 8,6,5,4,3\\
\midrule
Qwen3-14 billion & Qwen3 & +0.09 & +0.31 & -0.23 & +9.30 & +10.79 & +6.45 & 8,6,4,3\\
\midrule
Gemma-3-270 million & Gemma-3 & +1.99 & +2.35 & +0.88 & +26.15 & +27.94 & +25.93 & 8,6,5,4,3\\
\midrule
Gemma-3-1 billion & Gemma-3 & +0.95 & +0.59 & +0.88 & +15.33 & +14.12 & +9.80 & 8,6,5,4,3\\
\midrule
Gemma-3-4 billion & Gemma-3 & +0.72 & +0.50 & -0.27 & +21.73 & +19.24 & +15.30 & 8,6,5,4,3\\
\midrule
Gemma-3-12 billion & Gemma-3 & +0.40 & +0.32 & -0.42 & +20.19 & +22.92 & +20.75 & 8,6,5,4,3\\
\midrule
Gemma-3-27 billion & Gemma-3 & +0.33 & +0.23 & -0.67 & +18.46 & +17.15 & +16.10 & 8,6,5,4,3\\
\midrule
Gemma-4-31 billion & Gemma-4 & +0.94 & +0.58 & +1.28 & +20.35 & +19.85 & +14.90 & 8,6,5,4,3\\
\midrule
Muse-30 billion & Muse & +1.58 & +2.06 & -0.11 & +11.14 & +11.18 & +8.36 & 8,6,5,4,3\\
\midrule
OLMo-3-7 billion & OLMo-3 & +0.16 & +0.21 & -0.07 & +4.70 & +4.63 & +2.76 & 8,6,5,4,3\\
\midrule
OLMo-3-32 billion & OLMo-3 & +0.04 & +0.06 & +0.02 & +0.68 & +1.22 & +0.38 & 8,6,5,4,3\\
\bottomrule\end{tabular*}
\par\smallskip\begin{minipage}{\linewidth}\footnotesize Per-capability loss change under per-output-channel symmetric round-to-nearest weight quantization, in nats per native token, at the displayed bit widths; 8 and 6 bits lie within 0.01 nats of dense for each model and are omitted from the table as indistinguishable from dense.\end{minipage}
\end{table}

Under pruning, Table~\ref{tab:panel_prune} gives per-capability
damage at two densities, the density at which math damage first exceeds
one nat, and how QA ranks before that cliff. Magnitudes differ by two orders across series at the same density: OLMo-3-32B loses 0.01 nats on math at $d=0.7$ where Gemma-3-27B loses 4.0. The ordering of capabilities is
series-specific: in the Qwen3 sizes and in OLMo-3-32B, QA is the
least-damaged capability at each pre-cliff density and improves, by up
to $-0.79$ nats at Qwen3-4B; in each Gemma-3 size, in Gemma-4-31B,
and in Muse-30B, QA is the most-damaged capability at the mildest densities, by a factor of four on Gemma-4-31B;
OLMo-3-7B lies in between with differences of at most $0.15$ nats. A
measurement-noise floor sets the scale: the median $|\Delta L|$ at $d=0.9$ is
$0.013$ nats, and 29 of the 62 nominal improvement cells across all
arms exceed twice that median; the rest are treated as zero. Cliffs occur between $d=0.4$ and $0.7$ for Qwen3 and Gemma, at $0.55$ for Muse-30B, $0.35$ for
OLMo-3-7B and $0.2$ for OLMo-3-32B. Robustness is not monotone in size within a series: Gemma-3-12B is the most robust Gemma-3 size and 27B collapses earlier.
A predictor has to reproduce this heterogeneity, which no single property of a model family, size or training length included, determines by itself.

\paragraph{A second pruning criterion.} Wanda \citep{sun2023simple} scores each weight by its
magnitude times the norm of its input activation and prunes each output row of the block linear
layers, leaving embeddings dense, so its density counts a different scope from the panel's global
magnitude criterion. Table~\ref{tab:wanda-panel} measures it on Qwen3 4B and Gemma 3 4B at seven
densities. The criterion moves the amplitude of the response by an order of magnitude: at density
0.3 Gemma 3 4B loses 20.6 nats on mathematics under magnitude pruning and 2.1 under Wanda, and
Qwen3 4B 12.3 against 1.7. The shape in density travels better than the amplitude: with the exponent
of the power form pooled from the magnitude cells and only the scale refit, the Wanda responses are
matched to 0.15 and 0.08 nats on mathematics and 0.23 and 0.29 on code over the seven densities,
against 0.18, 0.14, 0.26 and 0.17 with a free exponent. A relation fitted under one criterion thus
transfers its shape and not its scale, and the scale has to be measured under the new criterion; a
single calibration at the mildest density does not supply it, for the reason given in
Section~\ref{sec:efficiency}, that the response there lies within measurement noise. Question
answering keeps its own sign: under Wanda it improves on both models at every density from 0.5
down, where magnitude pruning improves it only on Qwen3 and only above density 0.5.
\begin{table}[tb]
\normalfont
\centering\footnotesize
\caption{Loss change in nats per native token from the dense model under two pruning criteria at the same retained density, on the V6 probes: global magnitude pruning of every weight matrix (the panel's criterion) and Wanda, which scores each weight by its magnitude times the norm of its input activation over 128 C4 documents and prunes each output row of the block linear layers, leaving embeddings dense. A slash marks a density the magnitude panel did not measure.}
\label{tab:wanda-panel}
\begin{tabular*}{\textwidth}{@{\extracolsep{\fill}}>{\raggedright\arraybackslash\hspace{0pt}}p{\dimexpr 0.160000000\textwidth-2.240000000\tabcolsep\relax}>{\raggedright\arraybackslash\hspace{0pt}}p{\dimexpr 0.120000000\textwidth-1.680000000\tabcolsep\relax}>{\raggedright\arraybackslash\hspace{0pt}}p{\dimexpr 0.120000000\textwidth-1.680000000\tabcolsep\relax}>{\raggedright\arraybackslash\hspace{0pt}}p{\dimexpr 0.120000000\textwidth-1.680000000\tabcolsep\relax}>{\raggedright\arraybackslash\hspace{0pt}}p{\dimexpr 0.120000000\textwidth-1.680000000\tabcolsep\relax}>{\raggedright\arraybackslash\hspace{0pt}}p{\dimexpr 0.120000000\textwidth-1.680000000\tabcolsep\relax}>{\raggedright\arraybackslash\hspace{0pt}}p{\dimexpr 0.120000000\textwidth-1.680000000\tabcolsep\relax}>{\raggedright\arraybackslash\hspace{0pt}}p{\dimexpr 0.120000000\textwidth-1.680000000\tabcolsep\relax}@{}}
\toprule
 & & \multicolumn{2}{>{\centering\arraybackslash\hspace{0pt}}p{\dimexpr 0.240000000\textwidth-1.360000000\tabcolsep\relax}}{Math} & \multicolumn{2}{>{\centering\arraybackslash\hspace{0pt}}p{\dimexpr 0.240000000\textwidth-1.360000000\tabcolsep\relax}}{Code} & \multicolumn{2}{>{\centering\arraybackslash\hspace{0pt}}p{\dimexpr 0.240000000\textwidth-1.360000000\tabcolsep\relax}@{}}{QA} \\
\cmidrule(lr){3-4}\cmidrule(lr){5-6}\cmidrule(lr){7-8}
Model & Density & Magnitude & Wanda & Magnitude & Wanda & Magnitude & Wanda \\
\midrule
Qwen3, 4B & 0.9 & +0.01 & +0.01 & +0.02 & +0.01 & -0.00 & -0.00 \\
 & 0.8 & +0.00 & +0.02 & -0.01 & +0.00 & -0.29 & +0.04 \\
 & 0.7 & -0.00 & +0.05 & -0.09 & +0.04 & -0.79 & +0.08 \\
 & 0.6 & +0.15 & +0.10 & -0.00 & +0.13 & -1.76 & -0.01 \\
 & 0.5 & +1.31 & +0.19 & +0.90 & +0.24 & -1.48 & -0.41 \\
 & 0.4 & +5.00 & +0.61 & +5.83 & +0.67 & +1.37 & -1.22 \\
 & 0.3 & +12.34 & +1.70 & +11.95 & +2.51 & +6.84 & -1.88 \\
\midrule
Gemma 3, 4B & 0.9 & +0.02 & +0.00 & +0.00 & -0.00 & +0.05 & -0.00 \\
 & 0.8 & +0.15 & +0.01 & +0.11 & +0.01 & +0.22 & -0.04 \\
 & 0.7 & +0.83 & +0.04 & +1.47 & +0.06 & +0.24 & -0.09 \\
 & 0.6 & +2.21 & +0.17 & +3.16 & +0.16 & +0.07 & -0.20 \\
 & 0.5 & +7.00 & +0.48 & +11.06 & +0.50 & +2.84 & -0.39 \\
 & 0.4 & +19.77 & +1.50 & +10.50 & +1.92 & +18.92 & -0.45 \\
 & 0.3 & +20.61 & +2.10 & +16.17 & +3.03 & +16.63 & -0.70 \\
\bottomrule
\end{tabular*}
\end{table}

Under quantization, Table~\ref{tab:panel_quant} covers
14 models. Int8 and int6 stay within $0.01$ nats of dense. Int4 increases the loss of small models by one to two nats and changes large models by a few tenths; a QA
improvement under int4 recurs in five models and exceeds the measurement-noise floor. Int3 collapses 12 of the 14 models by $4.6$ to $28$ nats;
OLMo-3-7B loses $2.8$ to $4.7$ and OLMo-3-32B only $0.4$ to $1.2$, so
the collapse is common although not universal, and OLMo-3 is the robust
series under both interventions. Within the tested bit-widths a learned
exponential $q_c e^{-kb}$, with amplitude $q_c$ per capability and rate $k$, fits held-out bits better than the fixed $q_c4^{-b}$ shape, 6.7 against 7.6 nats pooled, which shows only
that the fixed base does not describe this quantizer; the squared
step-size ratio between 4 and 5 bits is $(7/15)^2\approx0.22$
(Appendix~\ref{sec:app_details}), and a loss ratio would follow only under assumptions about the error distribution and the local response that lie outside the scope of this
study.

Under distillation, black-box traces from two API teachers, 600 per domain, train Gemma-3 students under a full-trace and an
answer-only recipe in a paired design. The answer-only recipe increases the math loss about twice as much as the full recipe, while QA improves by 0.6 to 1.4 nats under both; because the recipes differ in supervised tokens and update
counts, this pairing does not isolate reasoning coverage, and the
controlled comparisons that do isolate it are the ones reported in Section~\ref{sec:law_distill} under the controlled recipe.

The loss endpoint supports these comparisons within limits. The same
heterogeneity appears on independent secondary benchmarks for math and
QA, where the primary-benchmark response predicts the secondary one
better than the response of any other capability does, with intervals
excluding zero in most hold-out protocols; for code the
cross-capability predictor is as good or better
(Appendix~\ref{sec:app_measure}, Fig.~\ref{fig:measurement}). A lower
QA loss does not by itself mean better QA behavior: in the four pruned
OLMo-3-32B cells whose QA loss lies below dense, exact-match accuracy
does not rise. ``QA improves'' in this paper refers to the loss and to nothing else.

\subsection{Scale and family structure}
\label{sec:family}

Within a family, robustness is not monotonic in size, and this holds in two families
independently: Gemma-3 robustness peaks at 12B and 27B collapses far earlier, in line with the earlier
ability-space observation of a Qwen3 14B peak. A mid-scale robustness peak appears in both families, in the present grid and in the earlier one.
One candidate predictor of this structure was retired. The
top-concentration statistic $\kappa_c$ of the capability gradient tracked robustness within the
Qwen3 ladder and was pre-specified as a predictor. The full grid does not support it: across families the
ordering inverts, since OLMo, the most robust family, has the lowest concentration, and within
Gemma-3 the largest size violates the within-family trend, so the statistic was retired.
At matched scale, on the panel of models near 30B parameters, the
family coefficient dominates: $\Delta\Lc{\text{math}}$ at $\dens=0.7$
spans two orders of magnitude, from 0.02 nats on OLMo-3-32B through 0.17 on Muse-30B and 0.68 on Gemma-4-31B to 4.04 on Gemma-3-27B. OLMo-3-32B survives $70\%$ pruning with
$\le+0.63$ nats on each capability. Family identity is a first-order determinant of prunability on this panel; which training-data or recipe properties carry it was not measured and is left open on this panel.

\section{Fitted Models and Coefficients}
\label{sec:app_coef}

The delivered pruning form (Eq.~\ref{eq:power}) and its standardization are in
Table~\ref{tab:round3_coef}. The coefficients of the grouped-quantization surface, which is not a
delivered relation, are in Section S2 of the supplement beside the analysis of its rank. The per-bit
source regressions, the distillation source and pool forms, and the pre-specified multi-student forms are
stored with their standardization constants in the development files listed in
Appendix~\ref{sec:app_registration}.
\begin{table}[tb]
\normalfont\centering\footnotesize
\caption{The table gives the pre-specified pruning exponent and coefficients for each capability. Coefficients act on standardized inputs; the resulting loss change is in nats per native token. The note gives the prediction formula and the constants used to standardize each input.}
\label{tab:round3_coef}
\begin{tabular*}{\textwidth}{@{\extracolsep{\fill}}>{\raggedright\arraybackslash\hspace{0pt}}p{\dimexpr 0.180000000\textwidth-1.800000000\tabcolsep\relax}>{\raggedright\arraybackslash\hspace{0pt}}p{\dimexpr 0.140000000\textwidth-1.400000000\tabcolsep\relax}>{\raggedright\arraybackslash\hspace{0pt}}p{\dimexpr 0.140000000\textwidth-1.400000000\tabcolsep\relax}>{\raggedright\arraybackslash\hspace{0pt}}p{\dimexpr 0.180000000\textwidth-1.800000000\tabcolsep\relax}>{\raggedright\arraybackslash\hspace{0pt}}p{\dimexpr 0.180000000\textwidth-1.800000000\tabcolsep\relax}>{\raggedright\arraybackslash\hspace{0pt}}p{\dimexpr 0.180000000\textwidth-1.800000000\tabcolsep\relax}@{}}
\toprule
Capability & Exponent & Intercept & Source size & Initial loss & Pretraining tokens \\
\midrule
Math & 3.75 & +3.275 & +0.319 & +3.850 & +0.250 \\
Code & 3.05 & +4.499 & +0.273 & +5.357 & +0.347 \\
QA & 3.05 & -7.409 & -0.481 & +5.486 & +0.106 \\
\bottomrule\end{tabular*}
\par\smallskip\begin{minipage}{\linewidth}\footnotesize Pre-specified coefficients of the selected pruning form $\widehat{\Delta L}_c=(\beta_c\cdot\phi)((1-d)/0.3)^{\gamma_c}$ with $\phi=[1,z(\log N_0),z(L_{0,c}),z(\log D_0)]$, standardized with centers 20.248, 2.661, 25.614 and scales 1.407, 1.691, 0.739 for $(\log N_0, L_{0,c}, \log D_0)$.\end{minipage}
\end{table}

\section{Relation Selection, Boundaries and Calibration}
\label{sec:app_forms}

This section records how each delivered relation was selected, where its boundary lies, which
capabilities it serves, and what structure the relations share.

\subsection{Selection and boundary of the delivered relations}
\label{sec:app_selbound}

The pruning relation was selected on the development panel. Fit on nine states at $d\in\{0.9,0.8,0.7,0.6\}$
and scored at the unseen 0.65 and 0.55, Eq.~\ref{eq:power} reduces the
strength-only error by a factor of two to three, and A2 matches it; refit on 17
states under a pre-specified leave-one-source-out rule, A2 and the power form tie within 0.02 nats, at 0.479 and 0.494, and the rule prefers the continuous form that vanishes at $d=1$; the coefficients are in Appendix~\ref{sec:app_coef}, and the exponent $\gamma_c$ is a fitted constant whose value is supported among the same-complexity alternatives tested in Table~\ref{tab:a14-baselines}: forms with the exponent fixed at one or replaced by a quadratic are worse by 0.04 to 0.20 nats on mathematics and code at both budgets.

Its boundary was then established. Evaluated as a pre-specified candidate on three new checkpoints, the power form reaches the math error of A2 with five parameters against twenty and does not beat the median curve on code or QA; on a 2.8B source the median curve is three to seven times better than either source form (Section~\ref{sec:general}). The compact form trades accuracy for compactness within a tested range; the median curve is the relation for new states, a rule fixed after these tests and reused without change in Section~\ref{sec:selection} for the selection rule and its confirmation.

For quantization, two parametric candidates were tested first, a
separable amplitude-times-shape form and a low-order surface in
$\log_2 q_{\max}$ and $\log_2 g$; the surface beat zero change at the
unseen $b=4$ and $g=128$ on the two-bit-level development set, but its quadratic term is unidentified there, with design rank 16 of 20, as Section S2 of the supplement shows. That earlier round contains the only test of a genuinely unseen bit width, and it is
reported here rather than in Table~\ref{tab:main-prediction-v2}. On the six development states at
bit width 4 with group sizes 64 and 256, fitted on 24 development configuration measurements per
capability, the pre-specified surface reaches 0.19, 0.21 and 0.44 nats on mathematics, code and question
answering, against 0.55, 0.73 and 0.54 for the development-selected median. The surface is not the
delivered relation, and the relation that is delivered can show no error of its own on these cells,
because they were added to the development grid of the round that selected it; the interpolation
variant scored on them in the earlier round uses different models and boundary rules and does not
stand in for it. Table~\ref{tab:main-prediction-v2} therefore carries only tasks on which the
delivered predictor has an independent error. In a confirmation round with all 54
unblinded cells as development set and 21 cells pre-specified at unseen
group sizes and on a new state, the interpolation won the pre-specified cells
for math and code and the median won the new state, so the delivered
rule was fixed after that round and is marked Retrospective in
Table~\ref{tab:main-prediction-v2} and in the tables of
Appendix~\ref{sec:app_results}. Table~\ref{tab:main-prediction-candidates} lists the pre-specified
candidate of each task beside the relation delivered after its test.
\begin{table}[t]\normalfont
\centering\footnotesize\fontsize{8}{9.5}\selectfont\linespread{0.85}\selectfont
\setlength{\tabcolsep}{1.5pt}
\renewcommand{\arraystretch}{0.88}
\setlength{\abovecaptionskip}{4pt}
\newcommand{\TableOneErrors}[3]{\setbox0=\hbox{#1 / #2 / #3}%
\ifdim\wd0>\linewidth #1\newline #2\newline #3\else\box0\fi}
\begin{tabular*}{\textwidth}{@{\extracolsep{\fill}}>{\raggedright\arraybackslash\hspace{0pt}}p{\dimexpr .24\textwidth-1.92\tabcolsep\relax}>{\raggedright\arraybackslash\hspace{0pt}}p{\dimexpr .22\textwidth-1.76\tabcolsep\relax}>{\raggedright\arraybackslash\hspace{0pt}}p{\dimexpr .12\textwidth-0.96\tabcolsep\relax}>{\raggedright\arraybackslash\hspace{0pt}}p{\dimexpr .22\textwidth-1.76\tabcolsep\relax}>{\raggedright\arraybackslash\hspace{0pt}}p{\dimexpr .20\textwidth-1.60\tabcolsep\relax}@{}}
\toprule
Prediction task & Pre-specified candidate & Candidate error (nats) & Delivered relation & Strongest development baseline \\
\midrule
Three unseen checkpoints pruned to densities 0.575, 0.675 and 0.85 & Five-parameter power form & \TableOneErrors{0.243}{0.236}{0.678} & Source-free median density curve\newline Retrospective & Per-density regression; median for QA \\
\midrule
Pythia 410 million and 1.4 billion at unseen group sizes 32 and 512, bit widths 3 to 5 & Source regression, the median and no change & \TableOneErrors{0.215}{0.557}{0.457} & Interpolate math and code; use the median for question answering\newline Retrospective & Bilinear regression; no change; median \\
\midrule
An unseen 1.4 billion stage at bit widths 3 to 5, group sizes 32 to 512 & Source regression, the median and no change & \TableOneErrors{0.303}{0.153}{0.222} & Development median\newline Retrospective & Bilinear regression; no change; median \\
\midrule
Gemma 270 million distilled on six new pools at 50 to 200 thousand tokens & Reuse forms, with a budget and pool form for question answering & \TableOneErrors{0.074}{0.019}{0.515} & The same predictor & Budget regression; reuse regression for Code and QA \\
\midrule
Gemma 1 billion distilled on six new pools at 50 to 200 thousand tokens & Reuse forms, with a budget and pool form for question answering & \TableOneErrors{0.057}{0.049}{0.463} & The same predictor & Loss regression; reuse regression; size regression \\
\bottomrule
\end{tabular*}
\caption{\footnotesize\fontsize{8.5}{10}\selectfont Pre-specified candidates for Table~\ref{tab:main-prediction-v2}, their errors on the same cells, delivered relations, and strongest development baselines. Errors are mean absolute errors in nats per token; errors and baseline names follow math, code and question answering order unless qualified. The distillation rows are the 270 million and 1 billion students.}
\label{tab:main-prediction-candidates}
\end{table}

The boundary of the quantization relation follows from the same tests: the interpolation holds at unseen group sizes of
seen states; on the new state the source-free median is as good or
better except for code, so source conditioning adds nothing there,
and the surface is not delivered, as the tests of Section~\ref{sec:unseen_settings} confirm on each
held-out cell of that panel.

For distillation, eight forms in $T$, $E$, and a source term were
fixed on twelve development runs and scored on nine unseen-pool
trajectories with the headline chosen after the test, and two
descriptor-carrying forms were compared retrospectively
(Section S2 of the supplement); the delivered forms were then
re-selected on all 25 trajectories by a pre-specified
leave-one-trajectory-out rule and pre-specified for six new pools of
$U{=}200$ at three new budgets. A linear form in the student state, fit on nine Pythia states, serves only as the math predictor for the distillation candidates of the selection rule in
Section~\ref{sec:selection}, and enters no other relation.

\begin{figure}[tb]
\centering
\includegraphics[width=\textwidth]{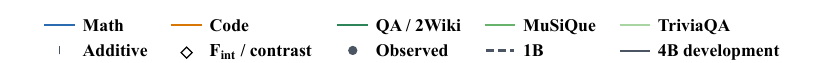}\\[1pt]
\begin{subfigure}[t]{2.7in}\centering\includegraphics[width=\linewidth]{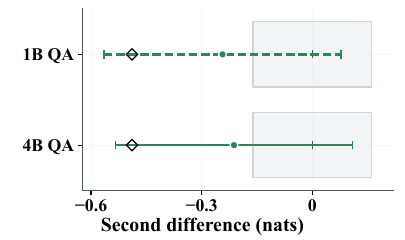}\caption{QA readout}\end{subfigure}\hfill
\begin{subfigure}[t]{2.7in}\centering\includegraphics[width=\linewidth]{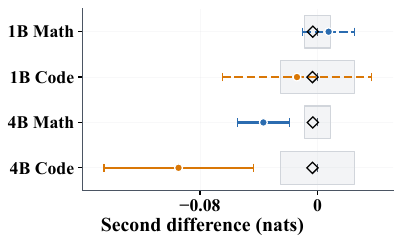}\caption{Math and code readouts}\end{subfigure}
\caption{The four-corner second difference per student and readout on its own scale: the measured
value with its pre-specified interval, the additive prediction of zero and the pre-specified interaction prediction. Both predictions lie inside the QA intervals, so the design did not separate the two
structures; the math and code readouts were specified in advance as underpowered before the trajectories
were trained.}
\label{fig:corner_test}
\end{figure}

\begin{figure}[tb]
\centering
\includegraphics[width=\textwidth]{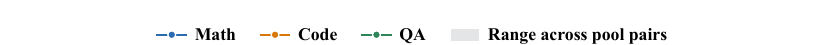}\\[1pt]
\begin{subfigure}[t]{2.7in}\centering\includegraphics[width=\linewidth]{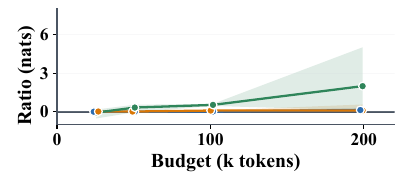}\caption{1B student}\end{subfigure}\hfill
\begin{subfigure}[t]{2.7in}\centering\includegraphics[width=\linewidth]{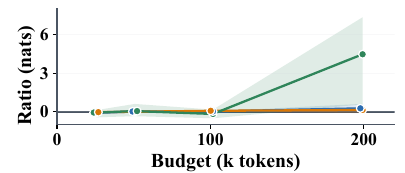}\caption{4B student}\end{subfigure}
\caption{Falsification of the logarithmic reuse term. The pool-change effect divided by its logarithmic bracket against the supervised budget, per capability, for the (a) 1B and (b) 4B students; the heavy line is the per-budget median over pool pairs and the band their range. A logarithmic reuse term would predict a flat line in budget.}
\label{fig:drift}
\end{figure}

Two pre-specified checks bound the joint distillation response. A form with separate logarithmic pool
and budget terms implies a pool-change effect flat in budget, yet for code on the 1B student that
effect drifts by a factor of twenty (Fig.~\ref{fig:drift}). A four-trajectory rectangle replaying the
trainer's update boundaries puts the second difference, zero under additivity, inside the
pre-specified band on question answering for both students (Fig.~\ref{fig:corner_test}), neither
rejecting nor establishing additivity. The delivered forms therefore rest on the leave-one-run-out
comparison of Table~\ref{tab:distill_confirm} and claim no additive joint structure.

The boundary of the distillation relation is set by the new-pool test. At the new pool size the
selected form beats the
pre-specified strongest baseline for code on both students, where that
baseline is the same $\log(1+E)$ form with an intercept, so the gain
supports the zero-budget boundary and the parsimony of the form and does not
establish a full student-data-training law; QA gains on 270M only, math loses to intercept baselines, no selected form is the post-hoc best of the sixteen pre-specified forms, a minimum selection favours by construction (Section S4 of the supplement); a student 6.6 times the reference size lies outside the tested range.

\subsection{Applicability across capabilities}
\label{sec:app_applic}

Where the initial-state inputs help is method- and range-specific:
they carry math and code at unseen pruning densities of seen sizes
and at unseen group sizes of seen states, add nothing on new
quantization states or in distillation beyond the reference anchor,
and over-predict the QA pruning response on each new checkpoint. What capability conditioning adds is measurable on the pre-specified
confirmation panels (Appendix~\ref{sec:app_shared}): per-capability
pruning curves beat a shared response curve with a signed
per-capability scale by 0.06 nats macro-averaged, largely through QA at 0.17 nats, whose aggregate curve is not proportional to the tested shared curve:
it changes sign across density, with a loss increase at $d=0.55$ and an improvement from 0.6 to 0.9, so no scale of that curve, including a negative one, represents it, and its direction also varies across sources. For math and
code a shared curve is as good or better, for grouped quantization the two are indistinguishable, and for distillation the
comparison is empty because the one-coefficient forms are
algebraically equivalent. Conditioning matters where the shape of the
response differs across capabilities.
\begin{table}[tb]
\normalfont
\centering
\small
\setlength{\tabcolsep}{3pt}
\caption{The table compares separate capability responses with shared responses on held-out confirmation panels. Mean absolute errors and gains are in nats per native token. Positive gains favor separate responses; brackets give paired confidence intervals for the error differences.}
\label{tab:cap_conditioning}
\begin{tabular*}{\textwidth}{@{\extracolsep{\fill}}>{\raggedright\arraybackslash\hspace{0pt}}p{\dimexpr 0.170000000\textwidth-2.040000000\tabcolsep\relax}>{\raggedright\arraybackslash\hspace{0pt}}p{\dimexpr 0.160000000\textwidth-1.920000000\tabcolsep\relax}>{\raggedright\arraybackslash\hspace{0pt}}p{\dimexpr 0.120000000\textwidth-1.440000000\tabcolsep\relax}>{\raggedright\arraybackslash\hspace{0pt}}p{\dimexpr 0.120000000\textwidth-1.440000000\tabcolsep\relax}>{\raggedright\arraybackslash\hspace{0pt}}p{\dimexpr 0.120000000\textwidth-1.440000000\tabcolsep\relax}>{\raggedright\arraybackslash\hspace{0pt}}p{\dimexpr 0.120000000\textwidth-1.440000000\tabcolsep\relax}>{\raggedright\arraybackslash\hspace{0pt}}p{\dimexpr 0.190000000\textwidth-2.280000000\tabcolsep\relax}@{}}
\toprule
Compression method & Capability & Separate capability response & Shared response with scale & Shared response with offset & Shared response only & Gain over scaled shared response [95\% confidence interval] \\
\midrule
Pruning & Math & 0.2279 & 0.2328 & 0.2249 & 0.1677 & {0.0048 [-0.0495, 0.0591]} \\
 & Code & 0.1736 & 0.1776 & 0.1750 & 0.1098 & {0.0040 [-0.0181, 0.0228]} \\
 & QA & 0.1818 & 0.3500 & 0.4626 & 0.6165 & {0.1683 [0.1054, 0.2434]} \\
 & Capability mean & 0.1944 & 0.2535 & 0.2875 & 0.2980 & {0.0590 [0.0206, 0.1025]} \\
\midrule
Grouped quantization & Math & 0.3234 & 0.3343 & 0.3380 & 0.3439 & {0.0109 [0.0024, 0.0160]} \\
 & Code & 0.3838 & 0.3744 & 0.4485 & 0.4312 & {-0.0094 [-0.0234, 0.0226]} \\
 & QA & 0.3222 & 0.3306 & 0.3126 & 0.3356 & {0.0084 [-0.0177, 0.0254]} \\
 & Capability mean & 0.3431 & 0.3465 & 0.3664 & 0.3703 & {0.0033 [-0.0119, 0.0145]} \\
\midrule
Distillation$^{\dagger}$ & Math & 0.0656 & 0.0656 & 0.0654 & 0.1313 & {0.0000 [0.0000, 0.0000]} \\
 & Code & 0.0339 & 0.0339 & 0.0437 & 0.0942 & {0.0000 [0.0000, 0.0000]} \\
 & QA & 1.2353 & 1.2353 & 0.6316 & 1.1196 & {0.0000 [0.0000, 0.0000]} \\
 & Capability mean & 0.4449 & 0.4449 & 0.2469 & 0.4483 & {0.0000 [0.0000, 0.0000]} \\
\bottomrule
\end{tabular*}
\par\smallskip\begin{minipage}{\linewidth}\footnotesize A positive gain favors the separate capability response; brackets give 95\% paired cluster-bootstrap intervals.\end{minipage}
\par\smallskip
\begin{minipage}{0.99\linewidth}\footnotesize $^{\dagger}$Distillation gains are identically zero: the per-capability reuse form and the shared reuse curve with a per-capability scale are the same function family, so A and B coincide. Development: pruning 17 states (84 cells); quantization 54 cells; distillation 25 trajectories. Confirmation: pruning 12 cells in 4 states, the two identical 2.8B labels counted once; quantization 21 cells in 3 states; distillation 12 trajectories sharing 6 pools. Bootstrap: 5,000 resamples, seed 0, whole states or shared pools with all budgets retained; fixed development fits. Distillation A and B are algebraically identical: $a_c=a s_c$ for nonzero $a$. This retrospective ablation does not constitute a new pre-specification of the comparison it reports.\end{minipage}
\end{table}

Shared structure and parameter ranges complete the picture. One function family across
methods holds for pruning by construction and fails for quantization
and distillation; within pruning one exponent shared across
capabilities costs nothing, whereas quantization coefficients must be
capability-specific; parameter intervals and 80\% prediction intervals that under-cover are reported in Appendix~\ref{sec:app_shared} together with the one-point calibration and the single measurement that it costs.

\begin{figure}[tb]
\centering
\includegraphics[width=5.5in]{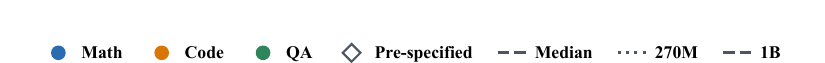}\\[1pt]
\begin{subfigure}[t]{1.8in}\centering\includegraphics[width=\linewidth]{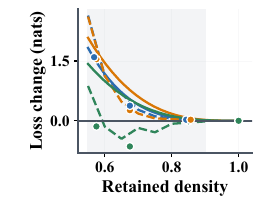}\subcaption{Pruning}\end{subfigure}\hfill
\begin{subfigure}[t]{1.8in}\centering\includegraphics[width=\linewidth]{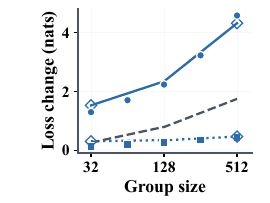}\subcaption{Math, 3 bit}\end{subfigure}\hfill
\begin{subfigure}[t]{1.8in}\centering\includegraphics[width=\linewidth]{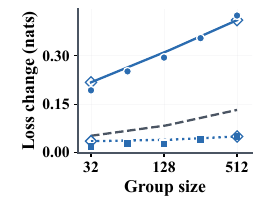}\subcaption{Math, 4 bit}\end{subfigure}\\[3pt]
\begin{subfigure}[t]{1.8in}\centering\includegraphics[width=\linewidth]{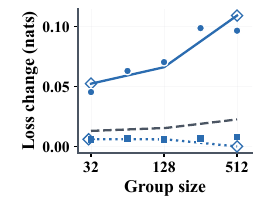}\subcaption{Math, 5 bit}\end{subfigure}\hfill
\begin{subfigure}[t]{1.8in}\centering\includegraphics[width=\linewidth]{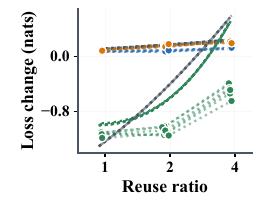}\subcaption{270M}\end{subfigure}\hfill
\begin{subfigure}[t]{1.8in}\centering\includegraphics[width=\linewidth]{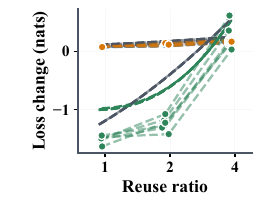}\subcaption{1B}\end{subfigure}

\caption{Measured responses and pre-specified relations on native configuration axes. (a) Pythia-1.4B at step 112k: measured densities, the delivered power form (solid, Eq.~\ref{eq:power}) and the pre-specified median development curves (dashed); shading marks the refit range $d\in[0.55,0.9]$. (b)--(d) Math responses to group size at 3, 4 and 5 bits on Pythia-410M at 143k (solid) and 1.4B at 16k (dotted): pre-specified piecewise interpolation, per-configuration medians (dashed), and predictions at the unmeasured $g=32,512$ (hollow) with their later measurements (filled). (e)--(f) Gemma-3 270M and 1B trajectories at $U=200$: measured responses at the recorded reuse ratio, the pre-specified selected forms (capability colours) and the strongest pre-specified baseline (grey), all on the native configuration axes.}
\label{fig:responses_relations}
\end{figure}

\subsection{Shared structure, parameter ranges, and one-point calibration}
\label{sec:app_shared}

Fig.~\ref{fig:responses_relations} plots the measured responses against the delivered relations on
each method's own configuration axis. One function
family for all three methods, with method- and capability-specific
parameters, holds for pruning by construction and fails for the other two arms on
the same held-out splits. Within pruning one exponent shared across capabilities costs nothing, within 0.02 nats of per-capability exponents, whereas the quantization coefficients must be capability-specific. Three ranges are kept apart: the bootstrap intervals of the pruning coefficients exclude zero for
the dense-anchor term on math and code and not on QA; per-source exponents spread widely, from 3.0 to 6.0 for math; and 80\% prediction intervals from leave-one-source residuals cover only 67 to 78 percent of held-out cells at a mean full width near 1.5 nats. One-point calibration of the pruning amplitude at the mildest density
amplifies noise and offers no saving over the uncalibrated predictor.

Table~\ref{tab:shared_structure} reports the shared-family test, the parameter-sharing tests and
the spread of exponents fitted separately per source. What one target measurement would buy, a
budget no delivered relation uses, is in Section S2 of the supplement.
\begin{table}[tb]
\normalfont
\centering\footnotesize\setlength{\tabcolsep}{3pt}
\caption{The table compares a shared response family with capability-specific responses, and reports the spread of exponents fitted separately per source. Errors are mean absolute errors in nats per native token, pooled equally over capabilities. Distillation carries no pretraining-token input, so that term is inactive.}
\label{tab:shared_structure}
\begin{tabular*}{\textwidth}{@{\extracolsep{\fill}}>{\raggedright\arraybackslash\hspace{0pt}}p{\dimexpr 0.170000000\textwidth-1.700000000\tabcolsep\relax}>{\raggedright\arraybackslash\hspace{0pt}}p{\dimexpr 0.130000000\textwidth-1.300000000\tabcolsep\relax}>{\raggedright\arraybackslash\hspace{0pt}}p{\dimexpr 0.140000000\textwidth-1.400000000\tabcolsep\relax}>{\raggedright\arraybackslash\hspace{0pt}}p{\dimexpr 0.180000000\textwidth-1.800000000\tabcolsep\relax}>{\raggedright\arraybackslash\hspace{0pt}}p{\dimexpr 0.180000000\textwidth-1.800000000\tabcolsep\relax}>{\raggedright\arraybackslash\hspace{0pt}}p{\dimexpr 0.200000000\textwidth-2.000000000\tabcolsep\relax}@{}}
\toprule
Law & What is shared & What varies & \mbox{Held-out} error, shared against capability-specific \\
\midrule
Pruning power & Power family & Source coefficients and exponent & 0.494 against 0.494\\
\midrule
Grouped quantization & Separable family & Source coefficients and exponents & 2.962 against 2.398\\
\midrule
Distillation exposure & $\log(1+E)$ family & Source coefficients; inactive $D_0$ & 0.542 against 0.275\\
\midrule
Pruning sharing & One Exponent & Source coefficients & 0.491 against 0.494\\
\midrule
Quantization sharing & two-dimensional term vectors & Capability offsets & 3.999 against 2.398\\
\midrule
Pruning Math & Power family & Source amplitudes, Exponent & Exponent minimum, median, maximum: 3.00, 4.60, 6.00\\
\midrule
Pruning Code & Power family & Source amplitudes, Exponent & Exponent minimum, median, maximum: 2.30, 5.95, 6.00\\
\midrule
Pruning QA & Power family & Source amplitudes, Exponent & Exponent minimum, median, maximum: 0.50, 3.40, 6.00\\
\bottomrule
\end{tabular*}
\end{table}

The value of capability conditioning can be measured on the same confirmation panels. At equal
development information, Table~\ref{tab:cap_conditioning}
compares the delivered per-capability source-free predictors, variant A, with three shared alternatives: one shared response curve times a signed per-capability scale, variant B; the shared curve plus a per-capability offset, variant C; and the shared curve alone, variant D, all fit on development data only and scored on the
held-out confirmation cells. With the two identical 2.8B labels merged into one state, the gain of A over B is 0.059 nats macro-averaged for pruning, 0.168 for QA, and within noise for math and code; 0.003 for quantization; and zero for distillation, where the single-coefficient forms of A and B coincide, which
supports no conclusion about conditioning there. The audit of this
comparison (Section S2 of the supplement) shows that negative scales were allowed
and none occurred, and that the QA gain arises because the aggregate
QA curve reverses sign across density, a shape no proportional
baseline can represent; the direction of QA also varies across
development sources, which neither source-free variant captures. The ablation is retrospective and is reported as such.

\section{Pre-specified Prediction Results}
\label{sec:app_results}

This section collects the pre-specified predictions and their measured outcomes, cell by cell, in the
order in which the tests were run and the predictions were pre-specified.

\begin{figure}[tb]
\centering
\begin{subfigure}[t]{2.7in}\centering\includegraphics[width=\linewidth]{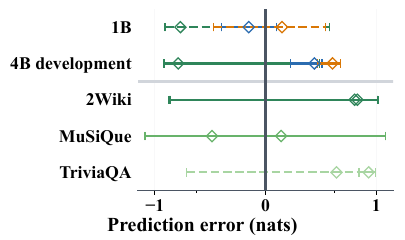}\caption{Prediction minus measurement}\end{subfigure}
\caption{Corner residuals for the two students and the pre-specified additive corner contrasts on the
fresh 2Wiki, MuSiQue and TriviaQA distributions, whose pre-specified additive prediction is zero; the
legend of Fig.~\ref{fig:corner_test} applies to every panel, with the same colours and markers.}
\label{fig:corner_contrasts}
\end{figure}

Section S4 of the supplement gives each pre-specified test in three lines, the candidate, the strongest same-input baseline and the strongest simple baseline, with improvements and provenance codes,
the per-stage and per-protocol results of the Pythia-1B and 6.9B pairs, the confirmation panel and the
grouped-quantization tests, and the paired improvements from adding each input and from each
functional form. Fig.~\ref{fig:corner_contrasts} gives the corner residuals on the fresh
distributions.
The same comparison in one line per test, each relation against its strongest same-input baseline
on the same cells, is in that section as well, with the same per-cell definition of error.
\begin{retargetreferences}{\retargetreference{Appendix Tables~\ref{tab:pred_source}--\ref{tab:pred_config_qd}}{Section S4 of the supplement}}

\begin{figure}[tb]
\centering
\includegraphics[width=\textwidth]{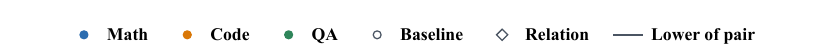}\\[1pt]
\begin{subfigure}[t]{\textwidth}\centering\includegraphics[width=\linewidth]{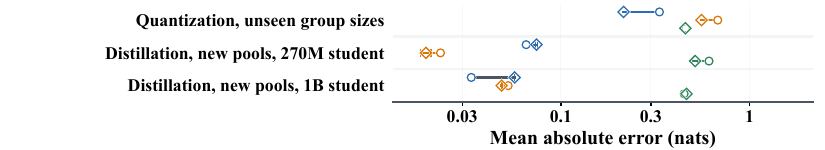}\caption{New configurations of model states that entered the fit}\end{subfigure}\\[2pt]
\begin{subfigure}[t]{\textwidth}\centering\includegraphics[width=\linewidth]{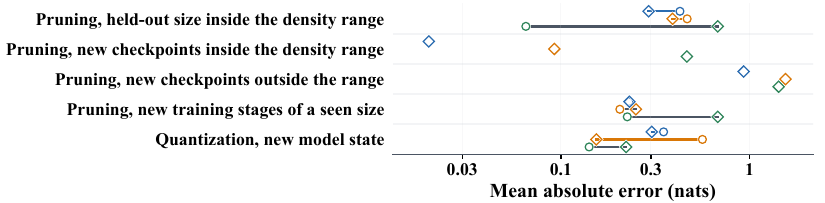}\caption{Model states that entered no fit}\end{subfigure}
\caption{Generalization of the pre-specified relations by kind of novelty: each row compares the error of
the relation, a hollow diamond, with the development-selected baseline, a hollow circle, on the
same cells, one sub-row per capability; the segment is coloured where the relation is lower.}
\label{fig:generalization}
\end{figure}

\end{retargetreferences}

The second confirmation round added seven pruning densities from 0.55 to 0.9, of which 0.55 is an extrapolation, and five bit-widths from 3 to 8 on Pythia-1B at steps 32k and 112k, with protocol A fit
on the full nine-state development panel and protocol B on states at step 64k
or below. The 6.9B pair was a size extrapolation of about five times:
under protocol A at 112k, the source regression reaches 0.23, 0.19 and 0.96 nats on math, code and QA against 0.05, 0.07 and 0.15 for the median curve;
at $d=0.55$ the source-free curves win under both protocols, with
source-conditioned errors between 0.55 and 2.2 nats. For quantization under
protocol A the class-indicator candidate is worst or near worst above 3 bits, an order of magnitude behind the per-bit median, and overshoots
by about four nats at 3 bits; protocol B compresses the differences above
3 bits to the 0.02 to 0.05 range and leaves the candidate ahead at 3 bits only for the late stage. The full two-protocol results are tabulated per stage and protocol in Section S4 of the supplement.

The distillation confirmation round re-selected forms on all 25 trajectories of the uniform protocol under leave-one-trajectory-out folds, with ties within 0.02 nats resolved toward fewer parameters, which gave a reuse form for math and code and the joint form for QA, fixed predictions for six new pools
of $U{=}200$ at budgets of 50k, 100k, and 200k supervised tokens, and
fixed the strongest baseline per student and capability from the same development pass, a development-designated baseline; ranked post hoc over all sixteen pre-specified forms, a source-conditioned variant has the lowest test error for each student and capability, as Section S4 of the supplement records. Table~\ref{tab:distill_confirm} gives the paired differences with pool-cluster intervals over six independent pools. Dense losses
re-measured on a different GPU drift by at most 0.001 nats from the
pre-specified descriptor and were tolerated at 0.01; shifting each response
by the drift of its trajectory in either direction changes no interval sign in any of the 12 comparisons. Per-example losses are
not stored, so an item-level interval cannot be computed; each
evaluation scores 64 items per capability, and the pool-cluster intervals therefore cover the sampling of pools and nothing else.
\begin{table}[tb]
\normalfont
\centering
\small
\caption{The table compares distillation relations with baselines selected before confirmation. Mean absolute errors and paired improvements are in nats per native token. Improvement is baseline error minus relation error; brackets give paired confidence intervals. Pool sizes count traces per domain; budgets count supervised tokens.}
\label{tab:distill_confirm}
\begin{tabular*}{\textwidth}{@{\extracolsep{\fill}}>{\raggedright\arraybackslash\hspace{0pt}}p{\dimexpr 0.130000000\textwidth-1.560000000\tabcolsep\relax}>{\raggedright\arraybackslash\hspace{0pt}}p{\dimexpr 0.120000000\textwidth-1.440000000\tabcolsep\relax}>{\raggedright\arraybackslash\hspace{0pt}}p{\dimexpr 0.170000000\textwidth-2.040000000\tabcolsep\relax}>{\raggedright\arraybackslash\hspace{0pt}}p{\dimexpr 0.170000000\textwidth-2.040000000\tabcolsep\relax}>{\raggedright\arraybackslash\hspace{0pt}}p{\dimexpr 0.100000000\textwidth-1.200000000\tabcolsep\relax}>{\raggedright\arraybackslash\hspace{0pt}}p{\dimexpr 0.100000000\textwidth-1.200000000\tabcolsep\relax}>{\raggedright\arraybackslash\hspace{0pt}}p{\dimexpr 0.210000000\textwidth-2.520000000\tabcolsep\relax}@{}}
\toprule
Student & Capability & Selected & Baseline & Relation error & Baseline error & Improvement [95\% confidence interval] \\
\midrule
Gemma-3-270 million & Math & Reuse response & Budget only & 0.0743 & 0.0654 & {-0.0088 [-0.0098, -0.0081]} \\
 & Code & Reuse response & Reuse only & 0.0193 & 0.0230 & {0.0038 [0.0025, 0.0050]} \\
 & QA & {Joint budget-pool response} & Reuse only & 0.5150 & 0.6097 & {0.0947 [0.0774, 0.1093]} \\
\midrule
Gemma-3-1 billion & Math & Reuse response & {Response surface, initial loss} & 0.0569 & 0.0335 & {-0.0234 [-0.0235, -0.0233]} \\
 & Code & Reuse response & Reuse only & 0.0485 & 0.0532 & {0.0047 [0.0042, 0.0050]} \\
 & QA & {Joint budget-pool response} & {Response surface, student size} & 0.4634 & 0.4502 & {-0.0132 [-0.0198, -0.0098]} \\
\bottomrule
\end{tabular*}
\par\smallskip\begin{minipage}{\linewidth}\footnotesize Pre-specified distillation confirmation at pool size 200 and supervised tokens 50,100,200 thousand. Columns give the selected and baseline forms, chosen by development leave-one-run-out before confirmation, their mean absolute errors in nats, and the paired improvement (baseline minus selected) with 95\% percentile intervals from 5,000 paired pool-cluster bootstrap resamples.\end{minipage}
\end{table}

\subsection{Measurement efficiency of the compact pruning form}
\label{sec:app_a11}

\begin{figure}[tb]
\centering
\includegraphics[width=\textwidth]{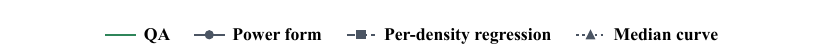}\\[1pt]
\begin{subfigure}[t]{1.8in}\centering\includegraphics[width=\linewidth]{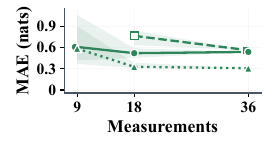}\caption{Error against budget}\end{subfigure}\hfill
\begin{subfigure}[t]{1.8in}\centering\includegraphics[width=\linewidth]{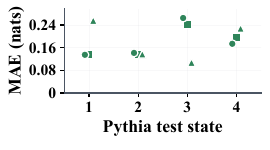}\caption{Four new Pythia states}\end{subfigure}\hfill
\begin{subfigure}[t]{1.8in}\centering\includegraphics[width=\linewidth]{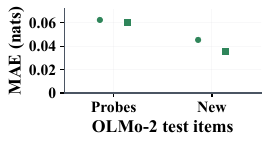}\caption{OLMo-2, half against full grid}\end{subfigure}
\caption{Measurement efficiency for question answering, the counterpart of
Fig.~\ref{fig:efficiency}. The median density curve is the better predictor at every development
budget, and on the four confirmation states the three predictors are within 0.006 nats of each
other, 0.177 for the per-density regression, 0.179 for the power form and 0.183 for the median
curve, so nothing here displaces the median curve as the delivered rule for this capability; on
OLMo-2 the reduced-grid form and the full-grid regression are within 0.002 nats on the probes and
0.009 apart on the new items, both far below the zero-change baseline of 0.13 and 0.16.}
\label{fig:efficiency_qa}
\end{figure}

Fig.~\ref{fig:efficiency_qa} gives the question-answering counterpart of the main-text figure. A retrospective subsampling study of the nine-state development panel, with twenty fixed replicates at 25, 50 and 100 percent of the measurements that reduce either the number of sources or the number of densities per source, found that across the replicates the five-parameter power form fitted on 18
measurements attains median errors of 0.20 nats on mathematics and 0.17 on code, against 0.18 and
0.13 for the per-density regression fitted on all 36, and that the
median curve is the better predictor for QA at every budget. Table~\ref{tab:a9} gives its error at each budget. A pre-specified confirmation then fixed
five predictors before four Pythia states in no earlier fit or test, listed in Table~\ref{tab:a11}, were pruned to densities 0.65 to 0.9;
Table~\ref{tab:a11} gives the outcome. The pre-specified rule, that the reduced-budget power form lies within 0.05 nats of the full-budget regression on math and code, is met.

\begin{table}[tb]
\centering\footnotesize
\caption{Pre-specified measurement-efficiency confirmation on four Pythia states outside every fit, 160M at step 80000, 410M at step 112000, 1.4B at step 48000 and 1B at step 48000, with 24 cells per capability: mean absolute error in nats per token of each pre-specified predictor. The
reduced-budget power form used the densities 0.9 and 0.7 of each development state, 18 measurements per capability.}
\label{tab:a11}
\begin{tabular*}{\textwidth}{@{\extracolsep{\fill}}lrrr@{}}
\toprule
Predictor (development measurements per capability) & Math & Code & QA \\
\midrule
Power form (18) & 0.073 & 0.104 & 0.179 \\
Power form (36) & 0.062 & 0.109 & 0.223 \\
Per-density regression (36) & 0.060 & 0.084 & 0.177 \\
Median density curve (36) & 0.112 & 0.170 & 0.183 \\
Median density curve (18) & 0.145 & 0.212 & 0.186 \\
\bottomrule
\end{tabular*}
\end{table}

\begin{table}[tb]
\centering\footnotesize
\caption{Subsampling study on the nine-state development panel: median mean absolute error in nats
on the new-state cells over twenty fixed replicates, when the nine-state development panel retains
9, 18 or all 36 density measurements per capability in total. Where fewer than twenty replicates could be fitted, the count
is given; a slash means no replicate could be fitted.}
\label{tab:a9}
\begin{tabular*}{\textwidth}{@{\extracolsep{\fill}}rlrrr@{}}
\toprule
Measurements per capability & Predictor & Math & Code & QA \\
\midrule
9 & Power form & 0.25 & 0.30 & 0.61 \\
9 & Per-density regression & / & / & / \\
9 & Median curve & 0.33 & 0.47 & 0.58 \\
18 & Power form & 0.20 & 0.17 & 0.52 \\
18 & Per-density regression & 0.24 (6/20 fits) & 0.28 (6/20 fits) & 0.76 (6/20 fits) \\
18 & Median curve & 0.21 & 0.21 & 0.33 \\
36 & Power form & 0.19 & 0.17 & 0.54 \\
36 & Per-density regression & 0.18 & 0.13 & 0.56 \\
36 & Median curve & 0.19 & 0.18 & 0.31 \\
\bottomrule
\end{tabular*}
\end{table}

\paragraph{Second family.} The measurement-efficiency result of Section~\ref{sec:efficiency}
rests on Pythia, so a pre-registered replication repeats the reduced-grid comparison on OLMo-2, whose
stage-1 checkpoints carry their pretraining token counts in their names and differ in weights by
construction; the shards of every checkpoint were hashed before measurement. Six development states,
OLMo-2 1B after 399, 1993 and 3608 billion tokens and 7B after 391, 1947 and 3507 billion, were pruned
at densities 0.9, 0.8, 0.7 and 0.6, the full grid of 24 configurations; the reduced grid keeps the same
six states at 0.9 and 0.7, 12 configurations. Four test states, 1B after 1196 and 2811 billion tokens
and 7B after 1167 and 2727 billion, enter no fit and were pruned at three densities fixed in advance,
0.85, 0.75 and 0.65, after every predictor had been frozen. The five predictors that share the compact
form's inputs were refitted on each grid with the same labels, the same exponent and curvature grids
and one estimator for every linear-in-feature form, ridge on the standardized inputs with the weight
chosen by leave-one-state-out; this estimator was fixed before any 7B or test state entered a fit,
after an unregularised dry fit on the three 1B states alone had produced code predictions of tens of
nats. Losses were measured on the development probes and on 64 new items per capability drawn from no
earlier probe, sample, teacher trace or few-shot exemplar. Table~\ref{tab:second-family} gives the
outcome over the four states and twelve cells per capability. Fitted on the reduced grid, the compact
form errs by 0.017 nats on mathematics and 0.020 on code against 0.019 and 0.027 for the per-density
regression fitted on the full grid, and by 0.062 against 0.060 on question answering; on the new items
the pairs are 0.012 against 0.013, 0.029 against 0.036 and 0.045 against 0.036. A paired item
bootstrap with the four states fixed puts the difference at $-0.002$ [$-0.005$, $+0.001$] on
mathematics, $-0.007$ [$-0.010$, $-0.004$] on code and $+0.006$ [$+0.000$, $+0.014$] on question
answering. Per state, the compact form's code error is below the regression's on three of the four
states and 0.001 above on the fourth, and its mathematics error is below on two and at most 0.004
above on the other two. Among the five reduced-grid predictors the compact form is the most accurate
on mathematics and code and within 0.001 nats of the best on question answering. Question answering is
the noisiest endpoint on this family: in two of the twelve test cells the measured response is smaller
at the harsher density than at the milder one, and every predictor's error there is two to three
times its mathematics error. The reduced grid took 59 percent of the full grid's GPU seconds and 60
percent of its evaluation tokens with dense anchors and model loading included, so half the
configuration measurements is not half the compute. A zero-change baseline errs by 0.080, 0.091 and 0.132 nats on the probes and by 0.087, 0.108 and
0.161 on the new items, so the compact form's error is four to five times smaller on mathematics and
code and half as large on question answering: the panel's responses are not trivial. The Pythia
coefficients applied to OLMo-2 without refitting err by 0.72, 1.88 and 0.38 nats. The pre-specified reading, the reduced-grid compact form within
0.020 nats of the full-grid regression on mathematics and code, is met with the form below the
regression on both: the structure and its measurement saving replicate across families, and the
coefficients do not transfer without refitting.
\begin{table}[tb]
\normalfont
\centering\footnotesize
\caption{Second-family confirmation on OLMo-2: mean absolute error in nats per token over 4 test states (1B and 7B intermediate checkpoints used in no fit) at three densities fixed before the freeze, 0.85, 0.75 and 0.65, on the 64 development probes per capability and on 64 new items per capability. Every predictor is fitted once on the six development states, on the reduced grid of two densities per state or on the full grid of four, with the same labels, the same tuning grids and the same leave-one-state-out ridge rule, and pre-specified before any test state was pruned. The reduced grid costs 59 percent of the full grid's GPU seconds and 60 percent of its evaluation tokens, dense anchors and model loading included.}
\label{tab:second-family}
\begin{tabular*}{\textwidth}{@{\extracolsep{\fill}}>{\raggedright\arraybackslash\hspace{0pt}}p{\dimexpr 0.260000000\textwidth-3.640000000\tabcolsep\relax}>{\raggedright\arraybackslash\hspace{0pt}}p{\dimexpr 0.140000000\textwidth-1.960000000\tabcolsep\relax}>{\raggedright\arraybackslash\hspace{0pt}}p{\dimexpr 0.100000000\textwidth-1.400000000\tabcolsep\relax}>{\raggedright\arraybackslash\hspace{0pt}}p{\dimexpr 0.100000000\textwidth-1.400000000\tabcolsep\relax}>{\raggedright\arraybackslash\hspace{0pt}}p{\dimexpr 0.100000000\textwidth-1.400000000\tabcolsep\relax}>{\raggedright\arraybackslash\hspace{0pt}}p{\dimexpr 0.100000000\textwidth-1.400000000\tabcolsep\relax}>{\raggedright\arraybackslash\hspace{0pt}}p{\dimexpr 0.100000000\textwidth-1.400000000\tabcolsep\relax}>{\raggedright\arraybackslash\hspace{0pt}}p{\dimexpr 0.100000000\textwidth-1.400000000\tabcolsep\relax}@{}}
\toprule
& & \multicolumn{3}{>{\centering\arraybackslash\hspace{0pt}}p{\dimexpr 0.300000000\textwidth-0.200000000\tabcolsep\relax}}{Probes} & \multicolumn{3}{>{\centering\arraybackslash\hspace{0pt}}p{\dimexpr 0.300000000\textwidth-0.200000000\tabcolsep\relax}@{}}{New items} \\
\cmidrule(lr){3-5}\cmidrule(lr){6-8}
Predictor & Development grid & Math & Code & QA & Math & Code & QA \\
\midrule
Compact power form & Reduced (12) & 0.017 & 0.020 & 0.062 & 0.012 & 0.029 & 0.045 \\
Quadratic strength form & Reduced (12) & 0.026 & 0.045 & 0.061 & 0.030 & 0.056 & 0.041 \\
Strength only & Reduced (12) & 0.018 & 0.029 & 0.076 & 0.019 & 0.037 & 0.050 \\
Median density curve & Reduced (12) & 0.026 & 0.039 & 0.074 & 0.030 & 0.051 & 0.048 \\
Per-density regression & Reduced (12) & 0.025 & 0.035 & 0.062 & 0.030 & 0.046 & 0.043 \\
\midrule
Compact power form & Full (24) & 0.012 & 0.028 & 0.055 & 0.010 & 0.037 & 0.051 \\
Quadratic strength form & Full (24) & 0.045 & 0.062 & 0.057 & 0.038 & 0.064 & 0.053 \\
Strength only & Full (24) & 0.019 & 0.031 & 0.069 & 0.018 & 0.037 & 0.060 \\
Median density curve & Full (24) & 0.022 & 0.036 & 0.069 & 0.021 & 0.037 & 0.053 \\
Per-density regression & Full (24) & 0.019 & 0.027 & 0.060 & 0.013 & 0.036 & 0.036 \\
\midrule
Zero change & None & 0.080 & 0.091 & 0.132 & 0.087 & 0.108 & 0.161 \\
Pythia coefficients, no refit & None & 0.724 & 1.883 & 0.382 & 0.717 & 1.864 & 0.412 \\
\bottomrule
\end{tabular*}
\end{table}

\paragraph{A second pruning criterion.} Both confirmations above prune by global magnitude, so a
pre-registered replication repeats the Pythia design with the pruning criterion changed to Wanda
\citep{sun2023simple}: every weight of the block linears is scored by its magnitude times the norm of
its input activation over 128 calibration documents of 512 tokens from the C4 validation split, and
the scoring is applied per output row at the target density. The nine development states and the four
test states, the four development densities and the six test densities, the grids, the six predictors,
the estimator rule and the freeze are those of the two earlier confirmations; the activation norms
are computed once per state and reused across its densities, and their time is charged to every grid.
Table~\ref{tab:wanda-efficiency} gives the outcome. Fitted on the reduced grid of 18 configurations,
the compact form errs by 0.023 nats on mathematics and 0.026 on code against 0.025 and 0.027 for the
per-density regression on all 36, and by 0.058 against 0.061 on question answering; a paired item
bootstrap with the states fixed puts the differences at $-0.001$ [$-0.002$, $-0.001$], $-0.001$
[$-0.002$, $-0.000$] and $-0.003$ [$-0.009$, $+0.002$], and on the new items at $-0.002$, $-0.001$ and
$-0.002$. The pre-specified reading, the reduced-grid compact form within 0.020 nats of the full-grid
regression on mathematics and code, is met. Two features of this panel bound what the result shows.
Wanda's responses at these densities are a fraction of magnitude pruning's: at density 0.65 the
mathematics and code losses rise by 0.12 and 0.23 nats on the 160M state and 0.27 and 0.40 on the 410M
state, by 0.02 to 0.06 on the 1B and 1.4B states, and question answering falls by 0.03 to 0.15 nats
on three of the four states. A zero-change baseline therefore errs by only 0.043, 0.058 and 0.036 nats,
so the compact form halves the baseline's error on mathematics and code and has no question-answering
response to predict; on the two largest test states, where every response stays within 0.06 nats,
no predictor improves on zero change, and the comparison that carries information is the 160M and
410M states, where the reduced-grid compact form errs by 0.024 and 0.042 on mathematics and 0.036 and
0.040 on code against 0.029 and 0.037 and 0.046 and 0.031 for the full-grid regression and 0.044 and
0.098 and 0.079 and 0.125 for zero change. The magnitude-pruning coefficients applied to the Wanda
measurements without refitting err by 0.19, 0.20 and 0.27 nats: they predict responses of
magnitude-pruning size. The reduced grid took 70 percent of the full grid's GPU seconds, more than on
the larger models because loading, the dense anchor and the calibration pass are a larger share of a
small model's cost, and 60 percent of its evaluation tokens. The structure and its measurement saving
therefore hold under a second pruning criterion where the criterion produces a response; the
coefficients are specific to the criterion.
\begin{table}[tb]
\normalfont
\centering\footnotesize
\caption{Measurement-efficiency confirmation under Wanda pruning on Pythia: mean absolute error in nats per token over 4 test states (the four A11 states, in no fit) at the six A11 densities, three of them unseen in development, on the 64 development probes per capability and on 64 new items per capability. Every predictor is fitted once on the nine development states, on the reduced grid of two densities per state or on the full grid of four, with the same labels, the same tuning grids and the same leave-one-state-out ridge rule, and pre-specified before any test state was pruned. The reduced grid costs 70 percent of the full grid's GPU seconds, the Wanda calibration pass counted once per state, and 60 percent of its evaluation tokens, dense anchors and model loading included in both grids, as in the two earlier confirmations.}
\label{tab:wanda-efficiency}
\begin{tabular*}{\textwidth}{@{\extracolsep{\fill}}>{\raggedright\arraybackslash\hspace{0pt}}p{\dimexpr 0.260000000\textwidth-3.640000000\tabcolsep\relax}>{\raggedright\arraybackslash\hspace{0pt}}p{\dimexpr 0.140000000\textwidth-1.960000000\tabcolsep\relax}>{\raggedright\arraybackslash\hspace{0pt}}p{\dimexpr 0.100000000\textwidth-1.400000000\tabcolsep\relax}>{\raggedright\arraybackslash\hspace{0pt}}p{\dimexpr 0.100000000\textwidth-1.400000000\tabcolsep\relax}>{\raggedright\arraybackslash\hspace{0pt}}p{\dimexpr 0.100000000\textwidth-1.400000000\tabcolsep\relax}>{\raggedright\arraybackslash\hspace{0pt}}p{\dimexpr 0.100000000\textwidth-1.400000000\tabcolsep\relax}>{\raggedright\arraybackslash\hspace{0pt}}p{\dimexpr 0.100000000\textwidth-1.400000000\tabcolsep\relax}>{\raggedright\arraybackslash\hspace{0pt}}p{\dimexpr 0.100000000\textwidth-1.400000000\tabcolsep\relax}@{}}
\toprule
& & \multicolumn{3}{>{\centering\arraybackslash\hspace{0pt}}p{\dimexpr 0.300000000\textwidth-0.200000000\tabcolsep\relax}}{Probes} & \multicolumn{3}{>{\centering\arraybackslash\hspace{0pt}}p{\dimexpr 0.300000000\textwidth-0.200000000\tabcolsep\relax}@{}}{New items} \\
\cmidrule(lr){3-5}\cmidrule(lr){6-8}
Predictor & Development grid & Math & Code & QA & Math & Code & QA \\
\midrule
Compact power form & Reduced (18) & 0.023 & 0.026 & 0.058 & 0.025 & 0.030 & 0.048 \\
Quadratic strength form & Reduced (18) & 0.026 & 0.048 & 0.038 & 0.028 & 0.052 & 0.042 \\
Strength only & Reduced (18) & 0.031 & 0.046 & 0.039 & 0.033 & 0.050 & 0.039 \\
Median density curve & Reduced (18) & 0.029 & 0.046 & 0.033 & 0.029 & 0.050 & 0.049 \\
Per-density regression & Reduced (18) & 0.024 & 0.032 & 0.060 & 0.026 & 0.034 & 0.049 \\
\midrule
Compact power form & Full (36) & 0.023 & 0.029 & 0.051 & 0.025 & 0.034 & 0.039 \\
Quadratic strength form & Full (36) & 0.037 & 0.049 & 0.036 & 0.038 & 0.049 & 0.045 \\
Strength only & Full (36) & 0.031 & 0.044 & 0.048 & 0.033 & 0.049 & 0.038 \\
Median density curve & Full (36) & 0.029 & 0.045 & 0.034 & 0.028 & 0.049 & 0.048 \\
Per-density regression & Full (36) & 0.025 & 0.027 & 0.061 & 0.027 & 0.031 & 0.048 \\
\midrule
Zero change & None & 0.043 & 0.058 & 0.036 & 0.043 & 0.065 & 0.039 \\
Magnitude-pruning coefficients, no refit & None & 0.186 & 0.199 & 0.265 & 0.186 & 0.192 & 0.244 \\
\bottomrule
\end{tabular*}
\end{table}

\paragraph{Same-complexity baselines.} Table~\ref{tab:a14-baselines} refits, on the same development
rows and with the same estimator as the compact power form, every form that shares its inputs and a
similar parameter count: the strength relation with the exponent fixed at one, a quadratic strength
relation, the strength-only curve, the median curve and the per-density regression, and the
pruning-law relation $L=L_0P_0d^{\alpha}$ of \citet{sengupta2025compression} fitted on capability
loss with one exponent and one scale per capability. At 18 measurements the compact form is the most accurate on mathematics and code, 0.036 and 0.033
nats ahead of the per-density regression on the same rows; at 36 that regression leads by 0.002 and
0.025, the comparison of Section~\ref{sec:efficiency}, so the compact form's advantage is the
measurements it saves. The fixed-exponent and quadratic forms trail it by 0.04 to 0.20 nats and the
pruning-law form by 0.04 to 0.06 at either budget. Calibrating one number
on each target's mildest pruned measurement, the power form's amplitude or the pruning law's $P_0$,
helps neither form on the remaining densities. These forms were fitted after the confirmation.
\begin{table}[tb]
\normalfont
\centering\footnotesize
\caption{Same-complexity baselines for the measurement-efficiency confirmation. Every form is fitted on the same development rows as the compact power form, at 18 and at 36 measurements per capability, with the same inputs and estimator, and scored on the same 24 confirmation cells per capability; entries are mean absolute errors in nats per token. The pruning-law form is the literature relation $L=L_0P_0d^{\alpha}$ \citep{sengupta2025compression} fitted on capability loss, with one exponent and one scale per capability and no calibration on the target. The last two rows calibrate one number on each target's mildest pruned measurement and are scored on the remaining five densities, twenty cells per capability, so they are not comparable with the rows above. All forms other than the compact power form, the median curve and the per-density regression at 36 measurements were fitted after the confirmation, on its stored development rows.}
\label{tab:a14-baselines}
\begin{tabular*}{\textwidth}{@{\extracolsep{\fill}}>{\raggedright\arraybackslash\hspace{0pt}}p{\dimexpr 0.280000000\textwidth-3.920000000\tabcolsep\relax}>{\raggedright\arraybackslash\hspace{0pt}}p{\dimexpr 0.200000000\textwidth-2.800000000\tabcolsep\relax}>{\raggedright\arraybackslash\hspace{0pt}}p{\dimexpr 0.086700000\textwidth-1.213800000\tabcolsep\relax}>{\raggedright\arraybackslash\hspace{0pt}}p{\dimexpr 0.086700000\textwidth-1.213800000\tabcolsep\relax}>{\raggedright\arraybackslash\hspace{0pt}}p{\dimexpr 0.086700000\textwidth-1.213800000\tabcolsep\relax}>{\raggedright\arraybackslash\hspace{0pt}}p{\dimexpr 0.086700000\textwidth-1.213800000\tabcolsep\relax}>{\raggedright\arraybackslash\hspace{0pt}}p{\dimexpr 0.086600000\textwidth-1.212400000\tabcolsep\relax}>{\raggedright\arraybackslash\hspace{0pt}}p{\dimexpr 0.086600000\textwidth-1.212400000\tabcolsep\relax}@{}}
\toprule
 & & \multicolumn{3}{>{\centering\arraybackslash\hspace{0pt}}p{\dimexpr 0.260100000\textwidth--0.358600000\tabcolsep\relax}}{18 measurements} & \multicolumn{3}{>{\centering\arraybackslash\hspace{0pt}}p{\dimexpr 0.259900000\textwidth--0.361400000\tabcolsep\relax}@{}}{36 measurements} \\
\cmidrule(lr){3-5}\cmidrule(lr){6-8}
Form & Parameters & Math & Code & QA & Math & Code & QA \\
\midrule
Compact power form & 5 & 0.073 & 0.104 & 0.179 & 0.062 & 0.109 & 0.223 \\
Linear strength (fixed exponent one) & 4 & 0.151 & 0.183 & 0.227 & 0.250 & 0.310 & 0.359 \\
Quadratic strength, same inputs & 5 & 0.115 & 0.142 & 0.206 & 0.187 & 0.238 & 0.317 \\
Strength only, no source inputs & 2 & 0.098 & 0.166 & 0.243 & 0.084 & 0.153 & 0.239 \\
Median density curve & one per density & 0.145 & 0.212 & 0.186 & 0.112 & 0.170 & 0.183 \\
Per-density regression & four per density & 0.108 & 0.136 & 0.194 & 0.060 & 0.084 & 0.177 \\
Pruning-law form on loss & 2 & 0.116 & 0.164 & 0.233 & 0.114 & 0.155 & 0.265 \\
\midrule
Compact power form, amplitude calibrated & one target measurement & 0.125 & 0.221 & 0.437 & 0.160 & 0.151 & 0.724 \\
Pruning-law form, $P_0$ calibrated & one target measurement & 0.129 & 0.183 & 0.250 & 0.184 & 0.228 & 0.349 \\
\bottomrule
\end{tabular*}
\end{table}

\paragraph{Prediction of task accuracy.} The pruning law's own endpoint is task accuracy, and
Table~\ref{tab:accuracy-comparison} meets it there: on Gemma 3 1B and 4B and Qwen3 1.7B and 4B
pruned to densities 0.8, 0.7 and 0.6, with accuracy measured by the protocol of
Appendix~\ref{sec:app_probes} on 64 probes per capability, the law's form fitted leave-one-model-out
predicts accuracy with a mean absolute error of 0.145 without calibration and
0.119 with one measurement, whereas the loss route, the final rule's development
median curve followed by a logistic link from loss to accuracy fitted on the other three models, errs by
0.264 and 0.312. Each relation is the better instrument on its own
endpoint: on capability loss the law's form trails the compact power form in every column of
Table~\ref{tab:a14-baselines}, and on accuracy the loss route trails the law, so neither replaces the
other and the paper's claims stay on loss. Two features of the accuracy endpoint explain the
ordering. In 17 of the 36 scored cells the measured accuracy is at most 0.03, a floor that a
collapsing power form in density reaches by construction and that the loss endpoint still resolves,
Gemma 3 1B moving from 3.4 to 10.1 nats on question answering across the same densities; and the link
applied to the measured loss itself errs by 0.147, so a large part of the loss
route's error is the coarseness of accuracy at this probe count rather than the loss prediction.
Where accuracy stays informative the picture reverses in places: Qwen3 4B keeps 0.84 on
mathematics at density 0.8, which the law puts at 0.24 and the loss route at 0.33.
\begin{table}[tb]
\normalfont
\centering\footnotesize
\caption{Prediction of task accuracy after magnitude pruning, the endpoint of the pruning law, on Gemma 3 1B and 4B and Qwen3 1.7B and 4B at densities 0.8, 0.7 and 0.6, leave-one-model-out: mean absolute error in accuracy points over 36 cells per column, 64 probes per capability. The pruning-law form fits one exponent per capability on the other three models; the loss route predicts the loss with the final rule's development median curve and maps it to accuracy with a logistic link fitted on the other three models; the last row applies that link to the measured loss. Calibration uses one measurement of the target at density 0.9. Paired over cells, the loss route minus the law is +0.119 [+0.042, +0.200] without calibration and +0.193 [+0.104, +0.285] with it.}
\label{tab:accuracy-comparison}
\begin{tabular*}{\textwidth}{@{\extracolsep{\fill}}>{\raggedright\arraybackslash\hspace{0pt}}p{\dimexpr 0.340000000\textwidth-3.400000000\tabcolsep\relax}>{\raggedright\arraybackslash\hspace{0pt}}p{\dimexpr 0.260000000\textwidth-2.600000000\tabcolsep\relax}>{\raggedright\arraybackslash\hspace{0pt}}p{\dimexpr 0.100000000\textwidth-1.000000000\tabcolsep\relax}>{\raggedright\arraybackslash\hspace{0pt}}p{\dimexpr 0.100000000\textwidth-1.000000000\tabcolsep\relax}>{\raggedright\arraybackslash\hspace{0pt}}p{\dimexpr 0.100000000\textwidth-1.000000000\tabcolsep\relax}>{\raggedright\arraybackslash\hspace{0pt}}p{\dimexpr 0.100000000\textwidth-1.000000000\tabcolsep\relax}@{}}
\toprule
Predictor & Target measurement used & Math & Code & QA & Pooled \\
\midrule
Pruning-law form & none & 0.167 & 0.151 & 0.117 & 0.145 \\
Pruning-law form & one accuracy at density 0.9 & 0.139 & 0.128 & 0.090 & 0.119 \\
Loss route: median curve and link & none & 0.264 & 0.253 & 0.274 & 0.264 \\
Loss route: median curve and link & one loss at density 0.9 & 0.323 & 0.313 & 0.301 & 0.312 \\
Link on the measured loss & the target's own loss & 0.159 & 0.180 & 0.102 & 0.147 \\
\bottomrule
\end{tabular*}
\end{table}

\subsection{Measured data requirements under distillation}
\label{sec:app_dreq}

\begin{figure}[tb]
\centering
\includegraphics[width=\textwidth]{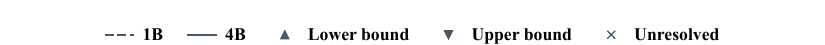}\\[1pt]
\begin{subfigure}[t]{1.8in}\centering\includegraphics[width=\linewidth]{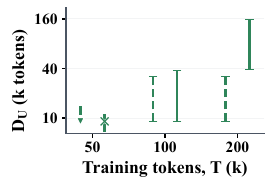}\caption{Fresh 2Wiki}\end{subfigure}\hfill
\begin{subfigure}[t]{1.8in}\centering\includegraphics[width=\linewidth]{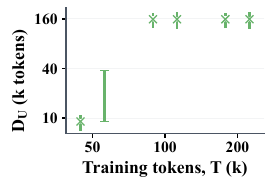}\caption{MuSiQue}\end{subfigure}\hfill
\begin{subfigure}[t]{1.8in}\centering\includegraphics[width=\linewidth]{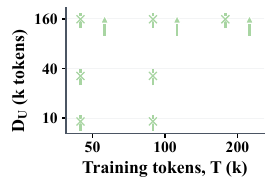}\caption{TriviaQA}\end{subfigure}
\caption{Measured data requirements of distillation at a tolerance of 0.25 nats: for each student
and budget, the range of independent tokens between the largest failing and the smallest passing
pool; an upward marker at the largest pool means the constraint is not yet met, a downward marker
at the smallest pool that it is met everywhere, and a cross that the seeds disagree. The largest
pool varies the training seed, and the smaller pools vary which traces are drawn into the pool.}
\label{fig:dreq}
\end{figure}

The pre-specified experiment behind Fig.~\ref{fig:dreq} trained eighteen trajectories on the 1B and 4B
students, three pool tiers each with three seeds, and evaluated every trajectory at 50, 100 and 200
thousand supervised tokens on six readouts. Table~\ref{tab:dreq-main} lists, for each readout, student and budget at the primary tolerance,
whether the constraint was crossed inside the tested pool range, the crossing interval or the
bound, and the distance of each predictor's recommended pool size from the measured interval; the
supplement gives the full accounting over all tolerances. The boundary relation pre-specified before the run did not predict these intervals better than a
per-student monotone interpolation, which solved the most requests over the same denominator on
the QA distributions and on math and code alike.
\begin{table}[tb]
\centering\scriptsize
\captionsetup{font=scriptsize}
\setlength{\tabcolsep}{2pt}
\renewcommand{\arraystretch}{0.95}
\caption{Measured data requirements at the primary loss tolerance $\tau=0.25$ nats are shown for both Gemma 3 students, identified by parameter count. Budgets count supervised training tokens, including reuse; k means thousands. The largest pool varies the training seed; smaller pools vary the pool draw. Intervals are observed brackets, not confidence intervals. Each boundary-rule recommendation gives a pool size followed by inside or its natural-log distance to the interval. Inside includes bracket endpoints and does not verify success. For a one-sided bound, distance is only a lower bound. Unresolved means no measured interval; none means the rule abstains.}
\label{tab:dreq-main}
\begin{tabular*}{\textwidth}{@{\extracolsep{\fill}}>{\raggedright\arraybackslash\hspace{0pt}}p{\dimexpr 0.100000000\textwidth-1.400000000\tabcolsep\relax}>{\raggedright\arraybackslash\hspace{0pt}}p{\dimexpr 0.070000000\textwidth-0.980000000\tabcolsep\relax}>{\raggedright\arraybackslash\hspace{0pt}}p{\dimexpr 0.070000000\textwidth-0.980000000\tabcolsep\relax}>{\raggedright\arraybackslash\hspace{0pt}}p{\dimexpr 0.230000000\textwidth-3.220000000\tabcolsep\relax}>{\raggedright\arraybackslash\hspace{0pt}}p{\dimexpr 0.140000000\textwidth-1.960000000\tabcolsep\relax}>{\raggedright\arraybackslash\hspace{0pt}}p{\dimexpr 0.130000000\textwidth-1.820000000\tabcolsep\relax}>{\raggedright\arraybackslash\hspace{0pt}}p{\dimexpr 0.130000000\textwidth-1.820000000\tabcolsep\relax}>{\raggedright\arraybackslash\hspace{0pt}}p{\dimexpr 0.130000000\textwidth-1.820000000\tabcolsep\relax}@{}}
\toprule
Readout & Student & Budget (k tokens) & Outcome & Measured interval (k independent tokens) & \multicolumn{3}{c@{}}{Recommended pool (k independent tokens)} \\
 & & & & & Boundary relation & Fixed reuse & Per-student interpolation \\
\midrule
Math & 1 billion & 50 & non-monotone & unresolved & 32.2, unresolved & 9.0, unresolved & 158.4, unresolved \\
Math & 1 billion & 100 & Crossed in range & $[9.0,\allowbreak\,32.5]$ & 32.2, inside & 32.2, inside & 32.2, inside \\
Math & 1 billion & 200 & Crossed in range & $[9.0,\allowbreak\,32.5]$ & 32.2, inside & 32.2, inside & 32.2, inside \\
Math & 4 billion & 50 & Upper bound & $\leq 9.1$ & 38.1, $\geq 1.43$ & 9.0, inside & 158.4, $\geq 2.86$ \\
Math & 4 billion & 100 & Crossed in range & $[8.9,\allowbreak\,38.7]$ & 38.1, inside & 38.1, inside & none \\
Math & 4 billion & 200 & Crossed in range & $[37.5,\allowbreak\,158.4]$ & 38.1, inside & 38.1, inside & 38.1, inside \\
\addlinespace[2pt]
Code & 1 billion & 50 & non-monotone & unresolved & 32.2, unresolved & 9.0, unresolved & 158.4, unresolved \\
Code & 1 billion & 100 & Crossed in range & $[9.0,\allowbreak\,32.5]$ & 32.2, inside & 32.2, inside & 32.2, inside \\
Code & 1 billion & 200 & Crossed in range & $[9.0,\allowbreak\,32.5]$ & 158.4, $1.58$ & 32.2, inside & 32.2, inside \\
Code & 4 billion & 50 & Upper bound & $\leq 9.1$ & none & none & 158.4, $\geq 2.86$ \\
Code & 4 billion & 100 & Crossed in range & $[8.9,\allowbreak\,38.7]$ & none & none & none \\
Code & 4 billion & 200 & seeds disagree & unresolved & none & none & 38.1, unresolved \\
\addlinespace[2pt]
Original 2Wiki & 1 billion & 50 & Upper bound & $\leq 9.1$ & 32.2, $\geq 1.26$ & 32.2, $\geq 1.26$ & 158.4, $\geq 2.86$ \\
Original 2Wiki & 1 billion & 100 & Crossed in range & $[9.0,\allowbreak\,32.5]$ & 32.2, inside & 32.2, inside & 32.2, inside \\
Original 2Wiki & 1 billion & 200 & seeds disagree & unresolved & 32.2, unresolved & 158.4, unresolved & 32.2, unresolved \\
Original 2Wiki & 4 billion & 50 & seeds disagree & unresolved & 38.1, unresolved & 38.1, unresolved & 158.4, unresolved \\
Original 2Wiki & 4 billion & 100 & Crossed in range & $[8.9,\allowbreak\,38.7]$ & 38.1, inside & 38.1, inside & none \\
Original 2Wiki & 4 billion & 200 & Crossed in range & $[37.5,\allowbreak\,158.4]$ & 38.1, inside & 158.4, inside & 158.4, inside \\
\addlinespace[2pt]
Fresh 2Wiki & 1 billion & 50 & Upper bound & $\leq 9.1$ & 32.2, $\geq 1.26$ & 9.0, inside & 32.2, $\geq 1.26$ \\
Fresh 2Wiki & 1 billion & 100 & Crossed in range & $[9.0,\allowbreak\,32.5]$ & 32.2, inside & 32.2, inside & 158.4, $1.58$ \\
Fresh 2Wiki & 1 billion & 200 & Crossed in range & $[9.0,\allowbreak\,32.5]$ & 158.4, $1.58$ & 32.2, inside & 32.2, inside \\
Fresh 2Wiki & 4 billion & 50 & seeds disagree & unresolved & 38.1, unresolved & 9.0, unresolved & 38.1, unresolved \\
Fresh 2Wiki & 4 billion & 100 & Crossed in range & $[8.9,\allowbreak\,38.7]$ & 38.1, inside & 38.1, inside & 38.1, inside \\
Fresh 2Wiki & 4 billion & 200 & Crossed in range & $[37.5,\allowbreak\,158.4]$ & 38.1, inside & 38.1, inside & 158.4, inside \\
\addlinespace[2pt]
MuSiQue & 1 billion & 50 & seeds disagree & unresolved & 32.2, unresolved & 32.2, unresolved & 32.2, unresolved \\
MuSiQue & 1 billion & 100 & seeds disagree & unresolved & 158.4, unresolved & 158.4, unresolved & none \\
MuSiQue & 1 billion & 200 & seeds disagree & unresolved & none & 158.4, unresolved & none \\
MuSiQue & 4 billion & 50 & Crossed in range & $[8.9,\allowbreak\,38.7]$ & 38.1, inside & 38.1, inside & 38.1, inside \\
MuSiQue & 4 billion & 100 & seeds disagree & unresolved & 158.4, unresolved & 38.1, unresolved & 158.4, unresolved \\
MuSiQue & 4 billion & 200 & seeds disagree & unresolved & 158.4, unresolved & 158.4, unresolved & none \\
\addlinespace[2pt]
TriviaQA & 1 billion & 50 & non-monotone & unresolved & 158.4, unresolved & 32.2, unresolved & 32.2, unresolved \\
TriviaQA & 1 billion & 100 & non-monotone & unresolved & 158.4, unresolved & 158.4, unresolved & none \\
TriviaQA & 1 billion & 200 & seeds disagree & unresolved & 158.4, unresolved & 158.4, unresolved & none \\
TriviaQA & 4 billion & 50 & Lower bound & $>158.4$ & none & none & none \\
TriviaQA & 4 billion & 100 & Lower bound & $>158.4$ & none & none & none \\
TriviaQA & 4 billion & 200 & Lower bound & $>158.4$ & none & none & none \\
\bottomrule
\end{tabular*}
\end{table}

\section{Mechanisms}
\label{sec:app_mech}

Four analyses develop the mechanism summary of Section~\ref{sec:resp_capabilities}: a pruning
law in the deleted capability mass, the sign of its first-order term, the geometry of the cliff, and
the logit-displacement account of the loss change, in that order.

\subsection{The pruning law}
\label{sec:law}

\begin{figure}[tb]
\centering
\includegraphics[width=\textwidth]{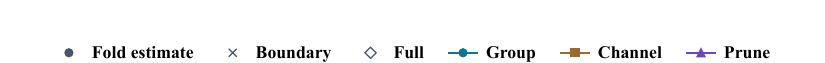}\\[1pt]
\begin{subfigure}[t]{2.7in}\centering\includegraphics[width=\linewidth]{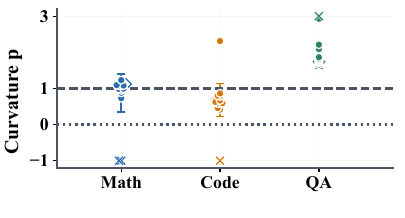}\caption{Reuse-term curvature}\end{subfigure}\hfill
\begin{subfigure}[t]{2.7in}\centering\includegraphics[width=\linewidth]{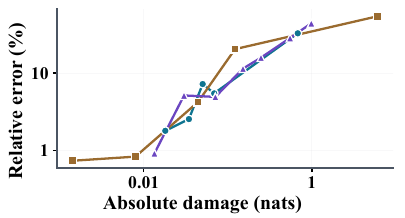}\caption{Displacement account}\end{subfigure}
\caption{Determinants of the distillation response and the displacement account of the compression loss change. (a) The reuse
exponent $p$ across training folds per capability with the full-development value and its
conditional interval; $p=0$ is the logarithmic term and $p=1$ linear. (b) Median relative error of
the second-order displacement account as a function of loss-change magnitude under pruning, grouped quantization and
per-channel quantization.}
\label{fig:explanation}
\end{figure}

The mechanism form below fits the full pre-cliff grid.
With the measured alignment term subtracted, the curvature residual
follows a power law in the deleted capability mass $m_c(\dens)$:
\begin{equation}
\Delta\Lc{c}(\dens)\;=\;\underbrace{\gbar{c}^{\top}\dw_\dens}_{\text{measured}}
\;+\;C\,\big[\underbrace{m_c(\dens)}_{\text{measured}}\big]^{\eta},
\label{eq:fitlaw}
\end{equation}
where $\gbar{c}$ is the mean gradient of the capability loss at the dense weights, $\dw_\dens$ the
weight change from pruning to density $\dens$, $C$ a fitted amplitude and $\eta$ a fitted exponent,
with in-sample $R^2 = 0.82$--$0.996$ over 24 capability$\times$model fits. The form is tested on the held-out battery with a 1000-resample cell bootstrap: in \emph{leave-one-family-out} transfer the
shared-$\eta$ mechanism form attains MAE $2.17$ nats versus $10.3$
for per-capability raw sparsity and $8.2$ for the direct-damage baseline, with both intervals excluding zero, so the deleted-mass coordinate
transfers across families; it also beats the direct-damage baseline in fit-shallow/predict-deep and raw sparsity in leave-largest-model-out, while its edge over raw
sparsity within-model does not reach significance; without the first-order term QA is unfittable, and with it QA joins the law. Fig.~\ref{fig:explanation} shows the fitted reuse exponent and the accuracy of the displacement account against the size of the loss change. Two structural results follow. First, in the deleted-mass coordinate the exponent concentrates at
$\eta\in[1.2,2.2]$ across all families, versus $3.2$--$8.7$ in the
raw $(1-\dens)$ coordinate, which makes the form portable. Second, \emph{within a model, $\eta$ is shared across capabilities}: capability differences appear to be carried
by the two measured quantities, and the fitted structure is two
parameters per model. This mechanism form uses gradient
information of the dense model and is not a K0 predictor; it
is reported as an upper comparator for the reduced forms of the main
text and not as a bound on what the initial state can predict.

\subsection{The signed first-order term predicts, in both directions}
\label{sec:mech}

The alignment term $\gbar{c}^{\top}\dw$ is the family
coefficient that carries sign. Its sign predicts the presence or absence of the
capability-selective improvement in each case measured:
the QA term of Qwen3 is negative and QA improves under pruning, with a predicted $-0.83$ against a measured $-0.79$ at 4B and $\dens=0.7$; the QA term of Gemma is
positive and no improvement exists at any size; the term for OLMo-3-7B is positive, a mild degradation predicted and measured at $+0.015$ on code at $\dens=0.8$, while the term for OLMo-3-32B is negative, and
QA improvement under deep pruning appears. At the level of individual model$\times$capability$\times$density
cells in the mild regime, sign prediction accuracy is $67.6\%$ over 108 cells, above chance with a Wilson interval of 58 to 76 percent, and the family-level pattern of Qwen and OLMo-32B QA improvements, none for Gemma and an OLMo size flip was stated in advance. Beyond the mild regime, first- plus second-order
predictions fail by orders of magnitude at the cliff in each family:
the collapse is non-perturbative in loss space, consistent with the
two-regime structure of \eqref{eq:fitlaw}.

\subsection{The cliff is a geometric threshold crossing}
\label{sec:orderparam}

Measuring the capability geometry \emph{of the pruned models}
accounts for the location of the cliff on the three models tested. The capability-region mass retention, the fraction of the capability-exclusive top coordinates of the dense model still prominent after pruning, declines smoothly from the first densities with no cliff in geometry, while loss is still flat;
the behavioral cliff occurs where retention crosses a critical value.
The critical value is family-specific, near 0.04 to 0.10 for Gemma across sizes and capabilities and 0.13 to 0.23 for Qwen, but consistent
within a family across sizes and capabilities: the cliff boundary
$\dens_c^{*}(N_0,f)$, for size $N_0$ and family $f$, reduces to one critical-retention constant per
family, so the fitted cliff parameter of the two-regime
law becomes a measurable order parameter that varies smoothly within each family.
\preliminary{three models, two families; a multi-model extension is left to future work}

\subsection{The logit-displacement account of the compression loss change}
\label{sec:app_disp}

For a compressed model whose logits are displaced by $r$ from those of the reference, the change
in cross-entropy on a supervised token $y$ under the reference next-token distribution $\pi$ is
exactly
\begin{equation}
\Delta L \;=\; \big(\mathbb{E}_{\pi}[r]-r_y\big)\;+\;\tfrac12\,\mathrm{Var}_{\pi}(r)\;+\;O(\lVert r\rVert^3),
\label{eq:displacement}
\end{equation}
so the first-order term is the signed alignment of the displacement with the target and the
second-order term is coefficient-free. Evaluated on 339 cells across pruning, grouped quantization
and per-channel quantization on the development states, the truncation to second order reproduces
the measured loss change with a median relative error that depends on the loss-change magnitude and not on the
family or the knob (Table~\ref{tab:disp_boundary}): under 5\% wherever the loss change is below about
0.1 nats, and rising past 20\% by 0.25 nats. Grouped four-bit quantization with fine groups, at 0.06 nats, sits on the same error curve as per-channel four-bit quantization at 0.24 nats of damage.

\begin{table}[tb]
\normalfont
\centering\small
\caption{Median relative error of the coefficient-free second-order account, by family and
configuration. Rows identify compression families; the second column pairs each setting with its median relative error in percent, from a \mbox{post-hoc} analysis of the measured damage.}
\label{tab:disp_boundary}
\begin{tabular*}{\textwidth}{@{\extracolsep{\fill}}>{\raggedright\arraybackslash\hspace{0pt}}p{\dimexpr 0.250000000\textwidth-0.500000000\tabcolsep\relax}>{\raggedright\arraybackslash\hspace{0pt}}p{\dimexpr 0.750000000\textwidth-1.500000000\tabcolsep\relax}@{}}
\toprule
Family & Configuration: median relative error \\
\midrule
Pruning & $d=0.9$: 0.9\%; 0.85: 4.3\%; 0.8: 5.3\%; 0.75: 11.1\%; 0.7: 16.0\%; 0.65: 28.2\%; 0.6: 44.4\% \\
Grouped quantization & 5 bits: 1.6--2.7\% at each group size; 4 bits: 3.3\% to 9.2\% by group size; 3 bits: 35.9\% \\
Per-channel quantization & 8 bits: 2.5\%; 6 bits: 3.0\%; 5 bits: 4.2\%; 4 bits: 23.8\%; 3 bits: 53.9\% \\
\bottomrule
\end{tabular*}
\end{table}

Two readings of this account are excluded. Projecting the displacement onto the reference logits $\ell$
with a shift-invariant estimator, $\epsilon=-\mathrm{Cov}_{\pi}(r,\ell)/\mathrm{Var}_{\pi}(\ell)$, since the uncentred version moves by 85\% under a constant logit shift, the pure-shrinkage account
fails on magnitude: about three percent of the response lies in the shrinkage direction, and the
variance of the orthogonal residual carries the rest for math and code. That residual variance is
not shown to be random noise, and it is not a pre-compression predictor, because computing it needs
the compressed model. The account explains the loss change after the fact and enters no
relation as an input. Variance dominance is per capability: seventeen cells with negative loss
change are carried by the first-order term.

\section{Capability-Specific Selection}
\label{sec:app_selection}

\subsection{Independent cross-method validation}
\label{sec:app_s3}

The confirmation of Section~\ref{sec:selection} used four references, none of which supplied a
measurement to the final rule. Table~\ref{tab:s3-identity} gives the exact releases, revisions and prior use, and
separates a new model state from a new intervention on familiar weights. Each reference carried
four pruning densities, six quantization settings, the dense model and one student distilled in
this round under the recipe on which the delivered relation was fitted. Two questions were settled
before any outcome was observed, against two different pre-specified sets. The first is checkpoint
identity: the learned values of all six snapshots were compared with the thirty Pythia revisions in
the earlier records, 180 comparisons in which no snapshot matched any of them. The second is
prior use, which Table~\ref{tab:s3-identity} records: the two Pythia step-120000 states
appear in no earlier record, whereas the Gemma releases were measured in earlier development, so
for them the round tests new candidates on familiar weights. The dense reference and pristine
student losses were then measured on the fixed probes, and 144 capability predictions, 48
largest-increase predictions and both selection maps were fixed before any outcome was observed. Table~\ref{tab:s3-selection} reports the outcome for each reference and objective. Question answering is the only objective with an opportunity to capture, and the rule captures it;
on the other three the opportunity is negligible for every policy.
\begin{table}[tb]
\centering\footnotesize
\setlength{\tabcolsep}{4pt}
\renewcommand{\arraystretch}{1.12}
\caption{Identity of the eight roles in the selection validation. All are pretrained base releases, none instruction tuned, and the eight roles draw on six distinct snapshots because two serve twice. The two Pythia states appear nowhere in the earlier records; the Gemma releases were measured in earlier development, so for them this round is a new set of candidates on familiar weights. Every candidate in the round is a fresh run, and no measurement on any of these weights entered the fitted selection rule. Parameter counts cover the text-decoder matrices. Exact revisions and their identifiers, the evidence for each entry, and the times at which the predictions were fixed are recorded in the repository audit.}
\label{tab:s3-identity}
\begin{tabular*}{\textwidth}{@{\extracolsep{\fill}}>{\raggedright\arraybackslash}p{\dimexpr 0.16\textwidth-1.28\tabcolsep\relax}>{\raggedright\arraybackslash}p{\dimexpr 0.21\textwidth-1.68\tabcolsep\relax}>{\raggedright\arraybackslash}p{\dimexpr 0.12\textwidth-0.96\tabcolsep\relax}>{\raggedright\arraybackslash}p{\dimexpr 0.17\textwidth-1.36\tabcolsep\relax}>{\raggedright\arraybackslash}p{\dimexpr 0.30\textwidth-2.40\tabcolsep\relax}@{}}
\toprule
Role & Model and stage & Matrix parameters & Weights seen in earlier work & What is new in this round \\
\midrule
Reference 1 & Pythia 410M, training step 120000 & 302M & No & No earlier use; first compression \\
Student for reference 1 & Pythia 160M, training step 120000 & 85M & No & No earlier use; first training \\
\midrule
Reference 2 & Pythia 1.4B, training step 120000 & 1.21B & No & No earlier use; first compression \\
Student for reference 2 & Pythia 410M, training step 120000 & 302M & No & No earlier use; first training \\
\midrule
Reference 3 & Gemma 3, 1B & 698M & Yes & New candidates, familiar weights \\
Student for reference 3 & Gemma 3, 270M & 100M & Yes & Retrained from familiar weights \\
\midrule
Reference 4 & Gemma 3, 4B & 3.21B & Yes & New candidates, familiar weights \\
Student for reference 4 & Gemma 3, 1B & 698M & Yes & Retrained from familiar weights \\
\bottomrule
\end{tabular*}
\end{table}

\begin{table}[tb]
\centering\footnotesize
\caption{Independent cross-method selection validation. None of the four references supplied an intervention outcome to the rule, and each carries twelve candidates at seventeen storage budgets, including a distillation candidate trained in this round. Opportunity is the loss of the best feasible quantization candidate minus that of the best feasible candidate across all methods; policy regret is the loss of the candidate that policy selected minus that same all-method optimum. All values are means in nats per native token of that family over the sixteen budgets at which both policies find a feasible candidate, and are never pooled across families; no quantization setting fits the smallest budget. All four references meet the pre-specified criterion, which compares the two policies per reference after averaging over the four objectives. Read by objective, the rule is ahead or level throughout except code on Gemma 3, 1B, where it is 0.0001 nats behind.}
\label{tab:s3-selection}
\begin{tabular*}{\textwidth}{@{\extracolsep{\fill}}llrrr@{}}
\toprule
Reference & Objective & Opportunity & Rule regret & Quantization-only regret \\
\midrule
Pythia 410M, step 120000 & Math & 0.000 & 0.000 & 0.000 \\
 & Code & 0.000 & 0.000 & 0.000 \\
 & QA & 0.117 & 0.027 & 0.592 \\
 & Largest increase & 0.001 & 0.000 & 0.001 \\
\midrule
Pythia 1.4B, step 120000 & Math & 0.000 & 0.000 & 0.000 \\
 & Code & 0.000 & 0.000 & 0.000 \\
 & QA & 0.572 & 0.000 & 0.572 \\
 & Largest increase & 0.000 & 0.000 & 0.000 \\
\midrule
Gemma 3, 1B & Math & 0.047 & 0.000 & 0.047 \\
 & Code & 0.018 & 0.019 & 0.018 \\
 & QA & 0.886 & 0.000 & 1.835 \\
 & Largest increase & 0.041 & 0.040 & 0.041 \\
\midrule
Gemma 3, 4B & Math & 0.000 & 0.015 & 0.015 \\
 & Code & 0.000 & 0.010 & 0.010 \\
 & QA & 0.220 & 0.000 & 0.413 \\
 & Largest increase & 0.000 & 0.015 & 0.015 \\
\bottomrule
\end{tabular*}
\end{table}

A retrospective control, outside the pre-specified analysis, separates the selection opportunity that an
initial smaller student supplies by itself from the further opportunity associated with the
distillation. Table~\ref{tab:s3-retrospective-control} compares the best feasible loss when
quantization is supplemented with the pristine student against the same quantity when it is
supplemented with the distilled student, at the same nominal storage ratio. Its entries are
therefore differences in candidate-set opportunity, and not the loss change
$L(S_{\mathrm{KD}})-L(S_0)$ of a student measured against its own starting point. The final rule
never saw the pristine candidate, so the original pre-specified selections and the pre-specified results are
unchanged.
\begin{table}[tb]
\centering\footnotesize
\caption{This analysis is retrospective. The pristine student $S_0$ was not in the pre-specified candidate set, and the rule's selections remain unchanged.}
\label{tab:s3-retrospective-control}
\begin{tabular*}{\textwidth}{@{\extracolsep{\fill}}llrrrr@{}}
\toprule
Reference & Objective & \shortstack{Pristine\\opportunity} & \shortstack{Distillation opportunity\\change} & \shortstack{Other\\methods} & \shortstack{Pre-specified\\opportunity} \\
\midrule
Pythia 410M & Math & 0.000000 & 0.000000 & 0.000000 & 0.000000 \\
 & Code & 0.000000 & 0.000000 & 0.000227 & 0.000227 \\
 & QA & 0.005770 & 0.083639 & 0.027125 & 0.116534 \\
 & Largest increase & 0.000000 & 0.000000 & 0.000939 & 0.000939 \\
\midrule
Pythia 1.4B & Math & 0.000276 & -0.000276 & 0.000000 & 0.000000 \\
 & Code & 0.007068 & -0.007068 & 0.000037 & 0.000037 \\
 & QA & 0.034696 & 0.537548 & 0.000000 & 0.572243 \\
 & Largest increase & 0.001173 & -0.001173 & 0.000056 & 0.000056 \\
\midrule
Gemma 3, 1B & Math & 0.055018 & -0.007796 & 0.000073 & 0.047295 \\
 & Code & 0.028921 & -0.010511 & 0.000000 & 0.018410 \\
 & QA & 0.235258 & 0.650243 & 0.000000 & 0.885501 \\
 & Largest increase & 0.050995 & -0.010511 & 0.000716 & 0.041200 \\
\midrule
Gemma 3, 4B & Math & 0.000086 & -0.000086 & 0.000125 & 0.000125 \\
 & Code & 0.005650 & -0.005650 & 0.000182 & 0.000182 \\
 & QA & 0.000000 & 0.220057 & 0.000000 & 0.220057 \\
 & Largest increase & 0.000086 & -0.000086 & 0.000126 & 0.000126 \\
\bottomrule
\end{tabular*}
\par\smallskip\begin{minipage}{\linewidth}\footnotesize Both student states use the pre-specified nominal student-to-reference matrix storage ratio. Pristine opportunity is the best measured quantization loss minus the best loss after adding the pristine student. Distillation opportunity change is the signed difference between the best loss with the pristine student and the best loss with the distilled student, with quantization available in both cases; it is a change in the opportunity available to the candidate set, not the loss change of a student against its own starting point. A negative change means lost opportunity. Other methods measure the further gain from pre-specified pruning and the dense reference. These three columns sum to pre-specified opportunity before rounding. All means use the same sixteen of seventeen budgets, from 25\% to 100\%; 20\% is excluded because quantization is infeasible. Sets containing either student are feasible at all seventeen budgets only for Gemma 3, 1B; all other sets are feasible at sixteen. Budgets are operating points, not independent replicates. Losses are in nats per native token within each family. Largest increase takes the maximum capability loss increase relative to the dense reference before selecting a candidate. Both Pythia references use step 120000.\end{minipage}
\end{table}

\paragraph{Policy ablation.} A retrospective ablation scores three policies, defined after the
round from development data alone, on the same candidates, budgets and outcomes
(Table~\ref{tab:s3-ablation}): a development-fixed method priority with no numeric prediction,
which ranks methods by their oracle win rate on the development panel and takes the largest
feasible storage within the method; development-median predictions for every candidate, with the
development constant for the student; and the source-conditioned relations of
Section~\ref{sec:laws} where their inputs exist, which excludes the two Gemma references without a
disclosed pretraining token count, with the selected distillation forms for the student. Writing
$\widehat Q$ for the quantization-only choice, $Q^*$ for the best quantization candidate and $A^*$
for the best candidate of any method, the quantization-only regret decomposes as
$L(\widehat Q)-L(A^*)=[L(\widehat Q)-L(Q^*)]+[L(Q^*)-L(A^*)]$, the selection error inside the family
plus the opportunity; the last column of the table reports $L(Q^*)-L(\widehat A)$ for the rule's
choice $\widehat A$, the gain that survives a perfect quantization selection. On these references the heuristic matches the rule wherever the opportunity sits in one method, and
the medians add value on mathematics for Gemma 3 1B, where the rule's choice outside quantization is
what the heuristic misses; the numeric value of prediction is thus carried by the development
medians the rule already uses, and the contribution of the source-conditioned relations lies in
measurement efficiency (Section~\ref{sec:efficiency}) rather than in selection.
\begin{table}[tb]
\normalfont
\centering\footnotesize
\caption{Retrospective policy ablation on the independent selection validation. Every policy chooses from the same twelve candidates at the same seventeen budgets; entries are mean regret against the all-method optimum in nats per native token over the sixteen budgets at which both the rule and quantization find a candidate, with the opportunity of Table~\ref{tab:s3-selection} repeated. The development-priority policy ranks methods by their oracle win rate on the development panel and takes the largest feasible storage within the method, with no numeric prediction; the median policy predicts every candidate from development medians and the student from the development constant; the relations policy uses the source-conditioned relations of Section~\ref{sec:laws} where their inputs exist and the selected distillation forms; the final rule is the policy fixed before the round, and its quantization-only counterpart is in Table~\ref{tab:s3-selection}. The rule gain is the loss of the best quantization candidate minus that of the rule's choice, the gain the rule realises even against a perfect quantization selection. The three added policies were defined after the round from development data alone.}
\label{tab:s3-ablation}
\begin{tabular*}{\textwidth}{@{\extracolsep{\fill}}>{\raggedright\arraybackslash\hspace{0pt}}p{\dimexpr 0.220000000\textwidth-3.080000000\tabcolsep\relax}>{\raggedright\arraybackslash\hspace{0pt}}p{\dimexpr 0.140000000\textwidth-1.960000000\tabcolsep\relax}>{\raggedright\arraybackslash\hspace{0pt}}p{\dimexpr 0.130000000\textwidth-1.820000000\tabcolsep\relax}>{\raggedright\arraybackslash\hspace{0pt}}p{\dimexpr 0.090000000\textwidth-1.260000000\tabcolsep\relax}>{\raggedright\arraybackslash\hspace{0pt}}p{\dimexpr 0.090000000\textwidth-1.260000000\tabcolsep\relax}>{\raggedright\arraybackslash\hspace{0pt}}p{\dimexpr 0.110000000\textwidth-1.540000000\tabcolsep\relax}>{\raggedright\arraybackslash\hspace{0pt}}p{\dimexpr 0.080000000\textwidth-1.120000000\tabcolsep\relax}>{\raggedright\arraybackslash\hspace{0pt}}p{\dimexpr 0.140000000\textwidth-1.960000000\tabcolsep\relax}@{}}
\toprule
Reference & Objective & Opportunity & Priority & Median & Relations & Rule & Rule gain \\
\midrule
Pythia 410M, step 120000 & Math & 0.000 & 0.000 & 0.000 & 0.000 & 0.000 & 0.000 \\
 & Code & 0.000 & 0.000 & 0.000 & 0.000 & 0.000 & 0.000 \\
 & QA & 0.117 & 0.027 & 0.027 & 0.059 & 0.027 & 0.089 \\
 & Largest increase & 0.001 & 0.000 & 0.000 & 0.000 & 0.000 & 0.001 \\
\midrule
Pythia 1.4B, step 120000 & Math & 0.000 & 0.000 & 0.000 & 0.045 & 0.000 & 0.000 \\
 & Code & 0.000 & 0.000 & 0.000 & 0.005 & 0.000 & 0.000 \\
 & QA & 0.572 & 0.000 & 0.000 & 0.000 & 0.000 & 0.572 \\
 & Largest increase & 0.000 & 0.000 & 0.000 & 0.005 & 0.000 & 0.000 \\
\midrule
Gemma 3, 1B & Math & 0.047 & 0.047 & 0.000 & 0.047 & 0.000 & 0.047 \\
 & Code & 0.018 & 0.018 & 0.019 & 0.019 & 0.019 & 0.000 \\
 & QA & 0.886 & 0.000 & 0.000 & 0.000 & 0.000 & 0.886 \\
 & Largest increase & 0.041 & 0.040 & 0.040 & 0.040 & 0.040 & 0.001 \\
\midrule
Gemma 3, 4B & Math & 0.000 & 0.015 & 0.015 & 0.015 & 0.015 & -0.015 \\
 & Code & 0.000 & 0.010 & 0.010 & 0.010 & 0.010 & -0.009 \\
 & QA & 0.220 & 0.000 & 0.000 & 0.000 & 0.000 & 0.220 \\
 & Largest increase & 0.000 & 0.015 & 0.015 & 0.015 & 0.015 & -0.015 \\
\bottomrule
\end{tabular*}
\end{table}

\paragraph{Fresh-item check of the chosen models.} After the round, every model that a policy chose
for question answering, with each reference's dense model and initial student, was scored on 384
2WikiMultihopQA validation items that no probe, sample, teacher trace or few-shot exemplar had used
(Table~\ref{tab:p1-behaviour}). Losses and their differences are token-weighted means in the sense
of Eq.~\ref{eq:taskloss}, the differences bootstrapped over items. The loss endpoint replicates: the
distilled student the rule chose has a lower loss than the quantization-only choice by 0.55, 0.45 and
1.82 nats on Pythia 410M, Pythia 1.4B and Gemma 3 1B, with paired 95\% intervals excluding zero, and
by 0.19 on Gemma 3 4B with an interval that includes zero; against its own initial student it is lower
by 0.37, 1.05 and 0.91 nats on Pythia 1.4B, Gemma 3 1B and Gemma 3 4B and unchanged on Pythia 410M,
whose initial student already carries the advantage, so the initial student separates the part of the
gain that comes from the smaller model from the part that distillation adds. The same items were
then read out by greedy generation, an endpoint the selection never targeted, under two protocols.
The first, the 32-token protocol of Appendix~\ref{sec:app_probes}, scores the whole block before
the next question marker; under it the students, which answer in the teacher's longer style, reach
exact match of 0.029, 0.065, 0.000 and 0.003 against 0.044, 0.167, 0.036 and 0.086 for the
quantization-only choices. The second gives 96 tokens and applies one gold-blind extraction rule
to every model, the first sentence of the first line with markdown removed, and records whether a
generation reached the cap without a stop marker. Under that readout the students score 0.089,
0.091, 0.115 and 0.172, ahead of the quantization-only choice by 0.04, 0.07 and 0.09 on Pythia
410M, Gemma 3 1B and Gemma 3 4B, with paired intervals excluding zero, and behind it by 0.06 on
Pythia 1.4B; against their own initial students they are level on three references and 0.09 behind
on Pythia 1.4B, so the loss gains over the initial student do not convert into exact-match gains.
The students reach the 96-token cap on 63 to 99 percent of items, since they rarely emit a stop
marker: a protocol audit, recorded with the results, found that their training targets carry no
end-of-answer marker, while the stop handling, few-shot format and extraction rule are the same for
every candidate. The shared procedure makes the candidates comparable under one readout; it does not
exclude that the missing end-of-answer supervision itself shapes the students' generations. A target
without an end marker leaves the teacher-forced loss, the endpoint the selection optimises, intact,
and limits how far the generation readout can be read and carried to other distillation protocols;
Appendix~\ref{sec:app_eos} reports a paired control that adds the marker. A reading of 20 sampled items per model, recorded with the results, found the students'
first answer correct on 4, 7, 5 and 6 items against 3, 4, 2 and 5 for the quantization-only
choices, with the difference between correct content and exact match confined to run-on sentences
and initials that the rule now handles; the remaining student errors are wrong answers. The two
readouts disagree by up to 0.14 in exact match on the same models, so generation accuracy is
sensitive to the protocol, and its ordering of the candidates does not follow the loss ordering
on every reference. Reference-answer likelihood and generation correctness are therefore different
endpoints, and which of them a selector optimises decides which model it recommends; the rule of
Section~\ref{sec:selection} optimises the former, and a deployment that needs a particular answer
format or accuracy enters that endpoint as its own objective.
\begin{table}[tb]
\normalfont
\centering\scriptsize
\caption{Fresh-item check of the models the policies chose for question answering, on 384 unused 2WikiMultihopQA items: completion loss in nats per native token; exact match under the 32-token readout and under the 96-token readout with the gold-blind extraction rule, with that readout's token F1 and the percentage of generations reaching the cap without a stop marker; and the difference of token-weighted mean losses from the quantization-only choice, the statistic of the loss column, with its 95\% bootstrap interval over items, negative favouring the row. Parentheses name the policies that chose the model.}
\label{tab:p1-behaviour}
\begin{tabular*}{\textwidth}{@{\extracolsep{\fill}}>{\raggedright\arraybackslash\hspace{0pt}}p{\dimexpr 0.130000000\textwidth-1.820000000\tabcolsep\relax}>{\raggedright\arraybackslash\hspace{0pt}}p{\dimexpr 0.280000000\textwidth-3.920000000\tabcolsep\relax}>{\raggedright\arraybackslash\hspace{0pt}}p{\dimexpr 0.070000000\textwidth-0.980000000\tabcolsep\relax}>{\raggedright\arraybackslash\hspace{0pt}}p{\dimexpr 0.080000000\textwidth-1.120000000\tabcolsep\relax}>{\raggedright\arraybackslash\hspace{0pt}}p{\dimexpr 0.080000000\textwidth-1.120000000\tabcolsep\relax}>{\raggedright\arraybackslash\hspace{0pt}}p{\dimexpr 0.080000000\textwidth-1.120000000\tabcolsep\relax}>{\raggedright\arraybackslash\hspace{0pt}}p{\dimexpr 0.080000000\textwidth-1.120000000\tabcolsep\relax}>{\raggedright\arraybackslash\hspace{0pt}}p{\dimexpr 0.200000000\textwidth-2.800000000\tabcolsep\relax}@{}}
\toprule
Reference & Model (policies) & Loss & Exact match, 32 tokens & Exact match, 96 tokens & Token F1, 96 tokens & Truncated (percent) & Loss difference from the quantization-only choice \\
\midrule
Pythia 410M, step 120000 & pruned to density 0.8 (oracle) & 5.138 & 0.109 & 0.117 & 0.140 & 32 & -0.66 [-0.75, -0.57] \\
 & grouped 4 bits, group 32 (quantization oracle) & 5.388 & 0.094 & 0.102 & 0.122 & 38 & -0.41 [-0.51, -0.30] \\
 & per-channel 4 bits (rule, quantization only, oracle, quantization oracle) & 5.797 & 0.044 & 0.049 & 0.069 & 42 & 0.00 (reference) \\
 & per-channel 6 bits (quantization oracle) & 5.224 & 0.174 & 0.143 & 0.183 & 23 & -0.57 [-0.69, -0.46] \\
 & distilled student (rule, oracle) & 5.246 & 0.029 & 0.089 & 0.126 & 73 & -0.55 [-0.68, -0.42] \\
 & initial student & 5.219 & 0.070 & 0.078 & 0.110 & 51 & -0.58 [-0.70, -0.46] \\
 & dense reference & 5.228 & 0.193 & 0.182 & 0.219 & 22 & -0.57 [-0.68, -0.45] \\
\midrule
Pythia 1.4B, step 120000 & per-channel 4 bits (quantization only, quantization oracle) & 5.306 & 0.167 & 0.146 & 0.189 & 13 & 0.00 (reference) \\
 & distilled student (rule, oracle) & 4.860 & 0.065 & 0.091 & 0.138 & 63 & -0.45 [-0.57, -0.33] \\
 & initial student & 5.228 & 0.193 & 0.182 & 0.219 & 22 & -0.08 [-0.17, +0.01] \\
 & dense reference & 5.386 & 0.198 & 0.182 & 0.226 & 9 & +0.08 [+0.00, +0.16] \\
\midrule
Gemma 3, 1B & grouped 3 bits, group 32 (quantization oracle) & 6.795 & 0.010 & 0.026 & 0.069 & 87 & -0.48 [-0.63, -0.34] \\
 & grouped 4 bits, group 128 (quantization oracle) & 6.579 & 0.188 & 0.180 & 0.238 & 17 & -0.70 [-0.84, -0.56] \\
 & per-channel 4 bits (quantization only) & 7.279 & 0.036 & 0.042 & 0.057 & 88 & 0.00 (reference) \\
 & distilled student (rule, oracle) & 5.458 & 0.000 & 0.115 & 0.171 & 99 & -1.82 [-2.06, -1.60] \\
 & initial student & 6.511 & 0.128 & 0.133 & 0.200 & 35 & -0.77 [-0.93, -0.60] \\
 & dense reference & 6.621 & 0.211 & 0.195 & 0.270 & 7 & -0.66 [-0.82, -0.50] \\
\midrule
Gemma 3, 4B & grouped 3 bits, group 32 (quantization oracle) & 6.086 & 0.135 & 0.141 & 0.178 & 24 & +0.18 [+0.05, +0.33] \\
 & per-channel 4 bits (quantization only) & 5.902 & 0.086 & 0.081 & 0.167 & 27 & 0.00 (reference) \\
 & distilled student (rule, oracle) & 5.710 & 0.003 & 0.172 & 0.249 & 97 & -0.19 [-0.39, +0.02] \\
 & initial student & 6.621 & 0.211 & 0.190 & 0.269 & 8 & +0.72 [+0.52, +0.92] \\
 & dense reference & 6.200 & 0.276 & 0.234 & 0.313 & 4 & +0.30 [+0.14, +0.47] \\
\bottomrule
\end{tabular*}
\end{table}

\paragraph{Storage in bytes.} The selection states its budget in nominal storage, the fraction of
stored matrix parameters: the retained density for pruning, $b/16$ plus the 16-bit group scales for
quantization, and the ratio of block-matrix parameter counts for a student.
Table~\ref{tab:storage-accounting} prices the same candidates in bytes on one base, every weight
matrix of the model in 16-bit precision. Grouped quantization is exact and per-channel quantization
is within 0.1 percent of its nominal ratio, since its row scales are few. Unstructured pruning needs
an index: a bitmap adds one sixteenth at every density, and a compressed sparse row format doubles
the cost, so it pays only below density 0.5. A student's bytes exceed its nominal ratio by its
embedding rows, which a small model carries in larger proportion: 0.40 against 0.28 for Pythia 160M
under Pythia 410M, and 0.27 against 0.14 for Gemma 3 270M under Gemma 3 1B. Read as a byte budget, the grid changes the feasible set in 176 of the 272 selection cells, 128 of
them only through a pruned candidate's index, and moves the rule's choice out of it in 10, all
student choices at budgets 0.20 to 0.35; in 25 more the choice exceeds its budget by at most 0.32
percent, 24 per-channel candidates and one student. Re-running the selection with the feasible sets counted
in bytes, and scoring against the byte-feasible oracle, leaves the picture intact within each family,
native nats never pooled across tokenizers: on question answering the Pythia pair's opportunity is
0.31 nats against 0.34 nominally and the rule's regret 0.050 against 0.014, the difference being the
Pythia 410M student, infeasible in bytes at budgets 0.30 and 0.35, where the rule falls back to
quantization, with the quantization-only regret 0.54 against 0.58; the Gemma pair's opportunity is
0.51 against 0.55, the rule's regret zero under both units and the quantization-only regret 1.05
against 1.12. On mathematics, code and the largest increase the Pythia pair has no opportunity under
either unit, and the Gemma pair's rule regret of 0.008, 0.014 and 0.028 nats nominally falls to zero
in bytes, where the per-channel candidates that sat exactly at their budgets drop out of the
feasible sets. The rule
itself does not depend on the unit: it ranks whatever candidates the budget admits, and a deployment
that stores sparse weights or counts embeddings enters its own byte costs in place of the nominal ones.
\begin{table}[tb]
\normalfont
\centering\footnotesize
\caption{Storage of the selection candidates in bytes, as a fraction of the dense reference stored in 16-bit weights, beside the nominal storage ratio the selection uses. Pruning is priced in a bitmap format, 16-bit kept weights plus one bit per position, and in compressed sparse row format, 16-bit kept weights with 16-bit column indices and 32-bit row pointers; quantization packs its bits exactly and adds one 16-bit scale per output row or per group; the distilled student is stored in 16-bit weights. Byte ratios agree across the four references to three decimals except for the student, whose four values follow the order Pythia 410M, Pythia 1.4B, Gemma 3 1B, Gemma 3 4B, the four references of the validation.}
\label{tab:storage-accounting}
\begin{tabular*}{\textwidth}{@{\extracolsep{\fill}}>{\raggedright\arraybackslash\hspace{0pt}}p{\dimexpr 0.260000000\textwidth-1.040000000\tabcolsep\relax}>{\raggedright\arraybackslash\hspace{0pt}}p{\dimexpr 0.260000000\textwidth-1.040000000\tabcolsep\relax}>{\raggedright\arraybackslash\hspace{0pt}}p{\dimexpr 0.480000000\textwidth-1.920000000\tabcolsep\relax}@{}}
\toprule
Candidate & Nominal storage ratio & Bytes as a fraction of the dense reference \\
\midrule
Dense reference & 1.000 & 1.000 \\
Pruning, density 0.9 & 0.900 & bitmap 0.963; compressed sparse row 1.802 \\
Pruning, density 0.8 & 0.800 & bitmap 0.863; compressed sparse row 1.602 \\
Pruning, density 0.7 & 0.700 & bitmap 0.762; compressed sparse row 1.402 \\
Pruning, density 0.6 & 0.600 & bitmap 0.662; compressed sparse row 1.202 \\
Per-channel 8 bits & 0.500 & 0.501 \\
Per-channel 6 bits & 0.375 & 0.376 \\
Per-channel 4 bits & 0.250 & 0.251 \\
Grouped 4 bits, group 32 & 0.281 & 0.281 \\
Grouped 4 bits, group 128 & 0.258 & 0.258 \\
Grouped 3 bits, group 32 & 0.219 & 0.219 \\
Distilled student & 0.281, 0.250, 0.144, 0.217 & 0.400, 0.286, 0.268, 0.258 \\
\bottomrule
\end{tabular*}
\end{table}

\begin{figure}[tb]
\centering
\includegraphics[width=5.5in]{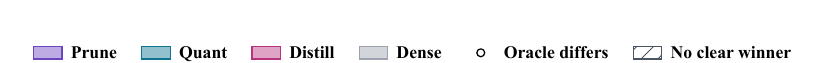}\\[1pt]
\begin{subfigure}[t]{1.8in}\centering\includegraphics[width=\linewidth]{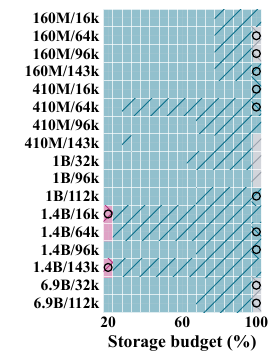}\subcaption{Math}\end{subfigure}\hfill
\begin{subfigure}[t]{1.8in}\centering\includegraphics[width=\linewidth]{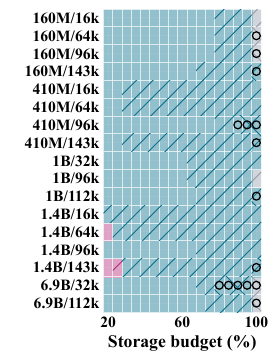}\subcaption{Code}\end{subfigure}\hfill
\begin{subfigure}[t]{1.8in}\centering\includegraphics[width=\linewidth]{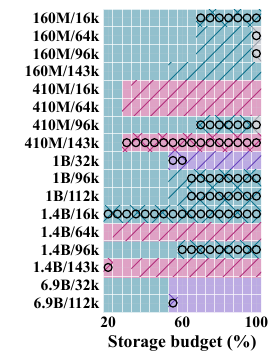}\subcaption{QA on 2Wiki}\end{subfigure}

\caption{Capability-specific selection maps on held-out Pythia states under the earlier laws, retrospectively and leave-one-state-out. For each state and nominal storage budget the colour is the method with the lowest predicted loss among feasible measured configurations; grey cross-hatching marks cells with no feasible configuration, diagonal hatching the no-clear-winner heuristic, and a hollow circle with cross-hatching a cell in which the measured oracle chooses another method.}
\label{fig:selection}
\end{figure}

This section gives the selection rule behind Section~\ref{sec:selection}, its retrospective validation
and its prospective confirmation. Section S6 of the supplement tabulates the per-policy results behind Fig.~\ref{fig:selection}: mean regret against the measured oracle, agreement with the oracle's method, the share of no-clear-winner cells, and the share of cells in which a fixed-method policy cannot meet the budget; such cells are infeasible, have no fallback and enter no mean.

The rule was first validated retrospectively. On the controlled panel of 17 states and budgets from 0.2 to 1, with each law refit with the target held out, the law-based choice is within 0.19, 0.17 and 0.42 nats of
the oracle for math, code and QA and picks the oracle's method in 96\%, 95\% and 73\% of 289 cells;
quantization-only is close behind.

\begin{figure}[tb]
\centering
\includegraphics[width=5.5in]{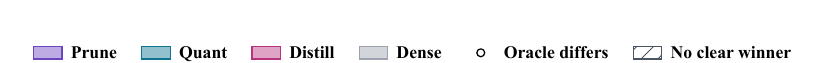}\\[1pt]
\begin{subfigure}[t]{2.7in}\centering\includegraphics[width=\linewidth]{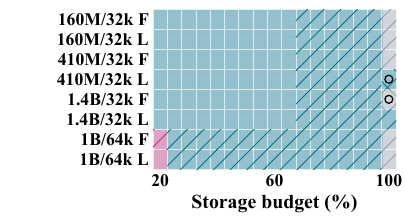}\subcaption{Math}\end{subfigure}\hfill
\begin{subfigure}[t]{2.7in}\centering\includegraphics[width=\linewidth]{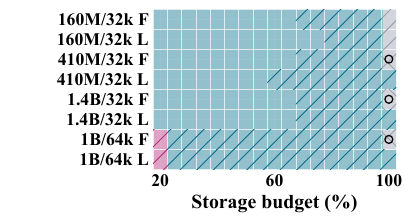}\subcaption{Code}\end{subfigure}\\[3pt]
\begin{subfigure}[t]{2.7in}\centering\includegraphics[width=\linewidth]{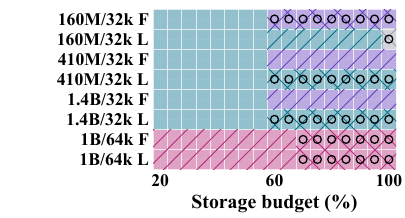}\subcaption{QA on 2Wiki}\end{subfigure}\hfill
\begin{subfigure}[t]{2.7in}\centering\includegraphics[width=\linewidth]{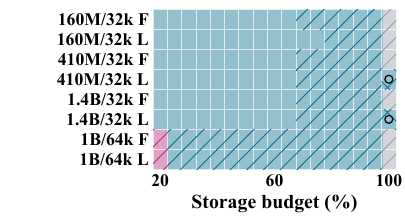}\subcaption{Largest loss increase}\end{subfigure}

\caption{Full selection maps on the four fresh states under the final rule and under the earlier source-conditioned laws, in row pairs, against nominal storage budget; colour: predicted method; hollow circle with cross-hatching: the oracle differs; diagonal hatching: no clear winner. The last panel uses the largest loss increase across capabilities as the objective.}
\label{fig:rule_maps}
\end{figure}

It was then fixed and confirmed on independent states. The rule fixed
before the confirmation panel was measured is as follows:
pruning uses the power form for math and code and the median curve
for QA at unseen densities of seen sizes, and the median development
curve for each capability on new sizes or stages; per-channel
quantization on new states uses the per-bit development median;
grouped quantization uses the same-input piecewise interpolation for
math and code and the per-configuration median for QA on seen states,
and the per-configuration median for each capability on new states;
distillation with a same-stage smaller student uses the student's
reference loss plus a per-capability constant for code and QA or the linear student-state form for math, QA on the 2Wiki distribution only.
Selection minimizes the predicted absolute loss over feasible measured
configurations, or the largest predicted loss increase across
capabilities for the multi-capability objective
(Table~\ref{tab:locked_rule} lists the rule as coded, including where
it departs from the delivered relations of Section~\ref{sec:laws}); quantization-only
uses the same candidates, predictor, and feasibility rule with no
fallback. Predictions under this rule and under the earlier
source-conditioned laws were pre-specified for four fresh states before
measurement (Table~\ref{tab:rule-confirm}; Fig.~\ref{fig:rule_maps});
the pre-specified criterion confirms a capability when the mean regret of the final rule over the 68 cells is within that of the source-conditioned map and below that of quantization-only; math, code, and QA meet it, while the multi-capability objective misses the first condition by 0.0001 nats, 0.0020 against
0.0019, and is reported as prospective and unconfirmed. The distillation candidates on 1B@64k are the
existing students 160M@64k and 410M@64k, whose post-training
outcomes were purged from the fits that predict them; per-state
regrets and candidate-set sizes are in
Section S6 of the supplement; the coverage columns there
describe whether the no-clear-winner candidate set contains the oracle's method or configuration, and the method agreement of the single chosen
configuration is the figure quoted in the main
text.
\begin{table}[tb]
\normalfont
\centering\small
\setlength{\tabcolsep}{3pt}
\caption{The table compares pre-specified selection policies on the independent confirmation panel. Mean regret is in nats per native token, and method agreement is the percentage of evaluated cells in which the policy's method matches the oracle's.}
\label{tab:rule-confirm}
\begin{tabular*}{\textwidth}{@{\extracolsep{\fill}}>{\raggedright\arraybackslash\hspace{0pt}}p{\dimexpr 0.220000000\textwidth-1.760000000\tabcolsep\relax}>{\raggedright\arraybackslash\hspace{0pt}}p{\dimexpr 0.380000000\textwidth-3.040000000\tabcolsep\relax}>{\raggedright\arraybackslash\hspace{0pt}}p{\dimexpr 0.180000000\textwidth-1.440000000\tabcolsep\relax}>{\raggedright\arraybackslash\hspace{0pt}}p{\dimexpr 0.220000000\textwidth-1.760000000\tabcolsep\relax}@{}}
\toprule
Capability & Policy & Regret & {Method agreement (percent)} \\
\midrule
Math & {Final selection rule} & 0.0000 & 98.5 \\
Math & {Source-conditioned} & 0.0149 & 98.5 \\
Math & Quantization only & 0.0068 & 94.1 \\
Math & Cheapest feasible & 1.7716 & 73.5 \\
\midrule
Code & {Final selection rule} & 0.0043 & 95.6 \\
Code & {Source-conditioned} & 0.1124 & 100.0 \\
Code & Quantization only & 0.0102 & 97.1 \\
Code & Cheapest feasible & 2.4346 & 75.0 \\
\midrule
QA  & {Final selection rule} & 0.1410 & 76.5 \\
QA  & {Source-conditioned} & 0.2136 & 61.8 \\
QA  & Quantization only & 0.2526 & 48.5 \\
QA & Cheapest feasible & 1.3518 & 63.2 \\
\midrule
{Largest increase} & {Final selection rule} & 0.0020 & 100.0 \\
{Largest increase} & {Source-conditioned} & 0.0019 & 97.1 \\
{Largest increase} & Quantization only & 0.0081 & 92.6 \\
{Largest increase} & Cheapest feasible & 2.4263 & 72.1 \\
\bottomrule
\end{tabular*}
\par\smallskip\begin{minipage}{\linewidth}\footnotesize Question answering is restricted to 2Wiki; policy predictions were pre-specified before the new measurements. The two candidate-set policies contain the oracle's method in every cell, except the source-conditioned policy on question answering, which contains it in 97.1 percent of cells.\end{minipage}
\par\smallskip\begin{minipage}{\linewidth}\footnotesize
The maximum-capability objective is $\max_c(L_c-L_{0c})$. Quantization only pools both round-to-nearest quantization arms. All four headline policies share the same 68 feasible cells. Heuristic sets use pre-specified development errors; set coverage is retrospective, not calibrated uncertainty. Verdicts: Mathematics: confirmed; Code: confirmed; Question answering (2Wiki): confirmed. The final selection rule uses the final policy, and the source-conditioned predictor uses the earlier selection laws.
\end{minipage}
\end{table}

\begin{table}[tb]
\normalfont
\centering\small
\setlength{\tabcolsep}{3pt}\renewcommand{\arraystretch}{1.05}
\caption{The table gives the selection predictor for each compression method, capability and source-state status. A prediction adds the listed response, in nats per native token, to a dense anchor: the source loss for pruning and quantization, the initial student's loss for distillation. Each method is fitted on its own development panel, listed below. The rule was pre-specified before confirmation.}
\label{tab:locked_rule}
\begin{tabular*}{\textwidth}{@{\extracolsep{\fill}}>{\raggedright\arraybackslash\hspace{0pt}}p{\dimexpr 0.190000000\textwidth-1.900000000\tabcolsep\relax}>{\raggedright\arraybackslash\hspace{0pt}}p{\dimexpr 0.130000000\textwidth-1.300000000\tabcolsep\relax}>{\raggedright\arraybackslash\hspace{0pt}}p{\dimexpr 0.160000000\textwidth-1.600000000\tabcolsep\relax}>{\raggedright\arraybackslash\hspace{0pt}}p{\dimexpr 0.220000000\textwidth-2.200000000\tabcolsep\relax}>{\raggedright\arraybackslash\hspace{0pt}}p{\dimexpr 0.170000000\textwidth-1.700000000\tabcolsep\relax}>{\raggedright\arraybackslash\hspace{0pt}}p{\dimexpr 0.130000000\textwidth-1.300000000\tabcolsep\relax}@{}}
\toprule
Compression method & Capability & State status & Response predictor \\
\midrule
Pruning & Mathematics & Seen source state; new density & Pruning power form \\
Pruning & Code & Seen source state; new density & Pruning power form \\
Pruning & Question answering & Seen source state; new density & {Pruning development median curve} \\
Pruning & Mathematics & {New size or new stage} & {Pruning development median curve} \\
Pruning & Code & {New size or new stage} & {Pruning development median curve} \\
Pruning & Question answering & {New size or new stage} & {Pruning development median curve} \\
\addlinespace
{Per-channel round-to-nearest quantization} & Mathematics & Seen state & {Per-bit source regression} \\
{Per-channel round-to-nearest quantization} & Code & Seen state & {Per-bit source regression} \\
{Per-channel round-to-nearest quantization} & Question answering & Seen state & {Per-bit source regression} \\
{Per-channel round-to-nearest quantization} & Mathematics & New state & {Per-bit development median} \\
{Per-channel round-to-nearest quantization} & Code & New state & {Per-bit development median} \\
{Per-channel round-to-nearest quantization} & Question answering & New state & {Per-bit development median} \\
\addlinespace
\bottomrule\end{tabular*}
\end{table}
\begin{table}[tb]\ContinuedFloat
\normalfont
\centering\small
\setlength{\tabcolsep}{3pt}\renewcommand{\arraystretch}{1.05}
\begin{tabular*}{\textwidth}{@{\extracolsep{\fill}}>{\raggedright\arraybackslash\hspace{0pt}}p{\dimexpr 0.190000000\textwidth-1.900000000\tabcolsep\relax}>{\raggedright\arraybackslash\hspace{0pt}}p{\dimexpr 0.130000000\textwidth-1.300000000\tabcolsep\relax}>{\raggedright\arraybackslash\hspace{0pt}}p{\dimexpr 0.160000000\textwidth-1.600000000\tabcolsep\relax}>{\raggedright\arraybackslash\hspace{0pt}}p{\dimexpr 0.220000000\textwidth-2.200000000\tabcolsep\relax}>{\raggedright\arraybackslash\hspace{0pt}}p{\dimexpr 0.170000000\textwidth-1.700000000\tabcolsep\relax}>{\raggedright\arraybackslash\hspace{0pt}}p{\dimexpr 0.130000000\textwidth-1.300000000\tabcolsep\relax}@{}}
\toprule
Compression method & Capability & State status & Response predictor \\
\midrule
{Grouped round-to-nearest quantization} & Mathematics & Seen state & {Piecewise source interpolation} \\
{Grouped round-to-nearest quantization} & Code & Seen state & {Piecewise source interpolation} \\
{Grouped round-to-nearest quantization} & Question answering & Seen state & {Per-configuration development median} \\
{Grouped round-to-nearest quantization} & Mathematics & New state & {Per-configuration development median} \\
{Grouped round-to-nearest quantization} & Code & New state & {Per-configuration development median} \\
{Grouped round-to-nearest quantization} & Question answering & New state & {Per-configuration development median} \\
\addlinespace
Distillation & Mathematics & New source & {Linear source regression} \\
Distillation & Code & New source & Pre-specified constant \\
Distillation & Question answering & New source & Pre-specified constant \\
\addlinespace
Dense & Mathematics & Any valid status & 0 (no loss change) \\
Dense & Code & Any valid status & 0 (no loss change) \\
Dense & Question answering & Any valid status & 0 (no loss change) \\
\bottomrule\end{tabular*}
\caption[]{The table continues the pre-specified selection rules. Absolute losses are in nats per native token.}
\end{table}
\begingroup\small
\noindent\textbf{Development data.} Pruning: pruning development set, 17 Pythia states, 84 density responses per capability ($0.55\le d<1$, excluding 2.8 billion). Channel quantization: development selection pre-specified paired channel rows; 16/16/4/17/17 states at 8/6/5/4/3 bits. Grouped quantization: grouped-quantization interpolation development grid, 160 million/410 million/1.4 billion at 16 thousand/143 thousand, $b\in\{3,4,5\}$, $g\in\{64,128,256\}$. Fixed-recipe students: fixed-recipe student fixed-recipe Pythia students pre-specified in development selection: 160 million/410 million/1.4 billion at 16 thousand/64 thousand/143 thousand, excluding both selectable students (160 million at step 64 thousand, 410 million at step 64 thousand), leaving seven. Recipe: teacher gpt-5.6-luna, full pool 600, two epochs, strict low-rank adaptation, seed 0. Full rosters and literal code are in the selection decomposition artifact.
\par\noindent\textbf{Exact forms.} The pruning development power response is $(\beta_c^\top z)[(1-d)/0.3]^{\gamma_c}$, with $z=(1,\allowbreak z(\log N_0),\allowbreak z(L_{0,c}),\allowbreak z(\log D_0))$ and development standardization. Its median curve linearly interpolates adjacent per-density medians. per-bit source regression is per-bit ordinary least squares on $(1,\allowbreak z(\log(N_0/10^9)),\allowbreak z(L_{0,c}),\allowbreak z(\log(D_0/10^9)))$; the new-state branch takes the median signed response at that bit. grouped-quantization interpolation interpolation uses per-configuration source-regression anchors $\beta_{c,b,g}^\top z$; its median branch uses per-configuration development medians. Both pass the anchors through the same piecewise bilinear interpolation in $(\log_2(2^{b-1}-1),\log_2(g/128))$. Outside the group grid, use the nearest boundary pair and floor only the extrapolated response at zero; no bit extrapolation. pruning development power and grouped-quantization interpolation source-anchor fits use ridge $10^{-3}$ including the intercept. Distillation math uses $\beta_0+\beta_1z(\log N_S)+\beta_2z(L_{S0,c})+\beta_3z(\log D_S)$; code and Question answering use the arithmetic mean development response for each capability, rather than its median.
\par\noindent\textbf{Domain and selection.} New status means a new state, a new size, a new stage, or a new source, taking precedence over seen size. Pruning requires $0.6\le d\le0.9$; channel round-to-nearest quantization requires a development bit. Distillation requires a smaller same-stage student ($N_S<N_0$, $D_S=D_0$), and Question answering requires the primary question-answering distribution. Undefined cells are rejected. Dense accepts every valid status. Selection minimizes predicted absolute loss or $\max_c(\widehat L_c-L_{0,c})$ over storage-feasible candidates; multi has no separately fitted predictor. independent selection invokes the new-stage branch for all four states included in the independent confirmation panel.
\par\noindent\textbf{Differences from the delivered-predictor table.} The delivered-predictor table allows power or per-density regression for seen-size pruning; this rule uses power for math/code and medians for Question answering and new size/stage. The channel-quantization branches now agree: both take the per-bit source regression on states in the fit and the per-bit development median on new ones. This rule was pre-specified with that branch before the confirmation panel was measured, and is unchanged. Grouped branches agree, with the boundary rule specified above. Its distillation exposure and joint budget--pool forms are not used here: selection uses the fixed-recipe student-state math form together with the code and Question answering constants fixed at the same time.
\endgroup

\begin{figure}[tb]
\centering
\includegraphics[width=5.5in]{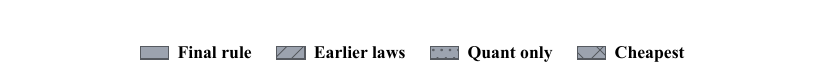}\\[1pt]
\includegraphics[width=5.5in]{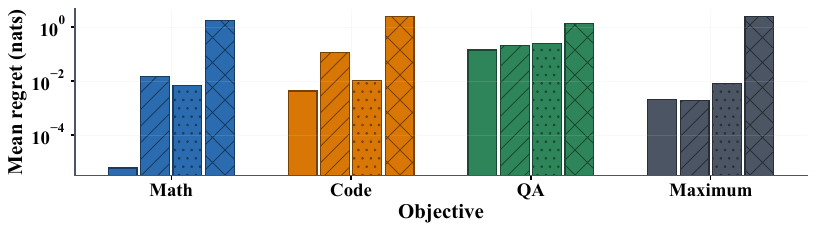}
\caption{Mean regret per objective on the four earlier Pythia states, in nats per native token on
a logarithmic scale, for the final rule, the earlier source-conditioned laws, quantization-only
and the cheapest feasible configuration; the saved pooled means of the confirmation round are
plotted without recomputation.}
\label{fig:rule_regret}
\end{figure}

\subsection{Protocol sensitivity of the generation readout}
\label{sec:app_eos}

The selection-round students were trained on teacher answers with no end-of-answer marker. A token-level
audit, recorded with the results, rebuilt the 1,781 training examples of each of the four students with
the trainer's own functions: no target carries the tokenizer's end-of-sequence token among its
supervised labels, the label mask covers exactly the completion tokens, 1,437 targets end in a sentence
mark and 4 in a newline, and the generation readout stops on the end-of-sequence token and on the
next-question markers for every candidate. To measure what the missing marker does, two students were retrained with one change, the
end-of-sequence token appended to every target and supervised, 1,781 extra supervised tokens per
epoch on the same examples, seed, learning rate and two-epoch schedule: the student of the Pythia
1.4B reference, on which the loss and generation orderings disagree most, and the student of the
Gemma 3 1B reference, which reached the 96-token cap on 99 percent of items.
Table~\ref{tab:eos-control} gives both pairs' readouts on the 384 fresh items. The marker restores
output control on both: the share of generations reaching the cap falls from 0.63 and 0.99 to 0.01
and 0.00, and the mean generation from 64 and 96 tokens to 5 and 8. It leaves the
reference-completion loss where it was, $-0.007$ nats [$-0.060$, $+0.047$] on Pythia and $+0.008$
[$-0.022$, $+0.037$] on Gemma. On generation accuracy the two students differ. The Pythia student's
exact match is unchanged, $+0.018$ [$-0.010$, $+0.049$] at 32 tokens and $0.000$ [$-0.034$,
$+0.031$] at 96; it keeps its loss advantage over the quantization-only choice, $-0.45$ nats, with
its exact-match deficit, $-0.07$ [$-0.11$, $-0.03$], and the same against its initial student,
$-0.38$ nats and $-0.10$. The Gemma student, whose untreated generations never stopped, gains
$+0.120$ [$+0.089$, $+0.154$] at 32 tokens and $+0.044$ [$+0.021$, $+0.068$] at 96, which puts it
ahead of the quantization-only choice on exact match by $0.086$ [$0.049$, $0.122$] as well as on loss
by $1.81$ nats, and level with its initial student on exact match, $-0.003$ [$-0.042$, $+0.036$],
while $1.05$ nats ahead on loss. The missing marker therefore costs generation accuracy where the
untreated student never terminates and nothing where it sometimes does; in neither case does the
loss gain over the initial student convert into an exact-match gain, and on Pythia 1.4B the
disagreement between the loss ordering and the generation ordering stands with the answers properly
terminated. Which endpoint a selector optimises still decides which model it recommends, on either family.
\begin{table}[tb]
\normalfont
\centering\footnotesize
\caption{End-marker control on the 384 fresh question-answering items for the selection-round students of the Pythia 1.4B and Gemma 3 1B references: the original student against a student retrained on the same 1,800 teacher answers, recipe, seed and schedule with the tokenizer's end-of-sequence token appended to every target and supervised, 1,781 extra supervised tokens per epoch. Differences are control minus original, with 95 percent intervals from a paired item bootstrap; the loss is token-weighted as in Eq. 1.}
\label{tab:eos-control}
\begin{tabular*}{\textwidth}{@{\extracolsep{\fill}}>{\raggedright\arraybackslash\hspace{0pt}}p{\dimexpr 0.290000000\textwidth-1.740000000\tabcolsep\relax}>{\raggedright\arraybackslash\hspace{0pt}}p{\dimexpr 0.130000000\textwidth-0.780000000\tabcolsep\relax}>{\raggedright\arraybackslash\hspace{0pt}}p{\dimexpr 0.180000000\textwidth-1.080000000\tabcolsep\relax}>{\raggedright\arraybackslash\hspace{0pt}}p{\dimexpr 0.400000000\textwidth-2.400000000\tabcolsep\relax}@{}}
\toprule
Readout & Original & With end marker & Difference [95\%] \\
\midrule
\multicolumn{4}{@{}>{\raggedright\arraybackslash\hspace{0pt}}p{\dimexpr 1.000000000\textwidth-0.000000000\tabcolsep\relax}@{}}{\emph{Pythia 1.4B reference, student Pythia 410M}} \\
Loss, nats per token & 4.860 & 4.853 & -0.007 [-0.060, +0.047] \\
Exact match, 32 new tokens & 0.065 & 0.083 & +0.018 [-0.010, +0.049] \\
Exact match, 96 new tokens & 0.076 & 0.076 & +0.000 [-0.034, +0.031] \\
Token F1, 96 new tokens & 0.117 & 0.106 & -0.011 [-0.045, +0.023] \\
Reached the 96-token cap & 0.628 & 0.005 & -0.622 [-0.672, -0.573] \\
Generated tokens, mean & 63.8 & 4.6 & -59.1 [-63.2, -54.8] \\
\midrule
\multicolumn{4}{@{}>{\raggedright\arraybackslash\hspace{0pt}}p{\dimexpr 1.000000000\textwidth-0.000000000\tabcolsep\relax}@{}}{\emph{Gemma 3 1B reference, student Gemma 3 270M}} \\
Loss, nats per token & 5.458 & 5.465 & +0.008 [-0.022, +0.037] \\
Exact match, 32 new tokens & 0.000 & 0.120 & +0.120 [+0.089, +0.154] \\
Exact match, 96 new tokens & 0.083 & 0.128 & +0.044 [+0.021, +0.068] \\
Token F1, 96 new tokens & 0.134 & 0.182 & +0.047 [+0.025, +0.071] \\
Reached the 96-token cap & 0.990 & 0.000 & -0.990 [-0.997, -0.979] \\
Generated tokens, mean & 96.0 & 7.9 & -88.1 [-88.6, -87.6] \\
\bottomrule
\end{tabular*}
\end{table}

\section{Reproducibility and Timing of Predictions}
\label{sec:app_registration}

Development and confirmation rounds ended by 11 September 2026; the corner design of
Appendix~\ref{sec:app_selbound} was specified on 14 September, and the measurement-efficiency confirmation of
Section~\ref{sec:efficiency} and the data-requirement experiment of Section~\ref{sec:observed} on 16
September, each before its models were trained or pruned; no measurement taken after a prediction
was fixed enters it or its baseline.
\label{sec:app_register}
\begin{table}[H]
\normalfont
\centering\footnotesize
\caption{Information budgets: K0 uses no compressed-target calibration, K1 one calibration
measurement, and the oracle row is a post-hoc diagnostic.}
\label{tab:budgets}
\begin{tabular*}{\textwidth}{@{\extracolsep{\fill}}>{\raggedright\arraybackslash\hspace{0pt}}p{\dimexpr 0.29\textwidth-1.16\tabcolsep\relax}>{\raggedright\arraybackslash\hspace{0pt}}p{\dimexpr 0.40\textwidth-1.60\tabcolsep\relax}>{\raggedright\arraybackslash\hspace{0pt}}p{\dimexpr 0.31\textwidth-1.24\tabcolsep\relax}@{}}
\toprule
Budget & Information available & Cost to disclose \\
\midrule
Metadata only (K0) & $N_0$, public $D_0$, architecture, configuration & $D_0$ may be undisclosed \\
Plus dense anchor (K0) & Plus the dense losses $L_{0,c}$, measured once & One forward pass per probe \\
One target calibration (K1) & Plus one pre-specified target measurement & Its configuration and cost \\
Oracle diagnostic & Rescaling on measured target outcomes & None; measures form misfit \\
\bottomrule
\end{tabular*}
\end{table}

Prospective tests follow one order: forms, development set, standardization and coefficients are fixed; the reference losses of a new source are measured; each candidate's predictions are recorded with their hash; only then is the outcome measured.

\section{Discussion}
\label{sec:app_learned}

Four lessons bear on how such laws should be built and used. A
rejected form is not a rejected class: the checks of Appendix~\ref{sec:app_selbound} reject the
logarithmic reuse term but not additivity of budget and reuse, so the class stays open to designs
that can resolve it.
An explanation is not a predictor: the displacement account reproduces the loss change across
three families but needs the compressed model to compute, which is why the delivered relations
read only what is known before compression. Source information transfers less than it fits: dense
pre-compression statistics improve a parametric family in four of nine pairs and lose to a
source-free configuration curve in nine of nine, which is why the selection rule calls that
curve's medians on every new state. Identifiability is a property of the design: the one fit-free
statistic that separates additive from interacting distillation structures was uncomputable until
four trajectories were designed to land on a budget-by-reuse rectangle. Each relation holds for the
evaluated distributions, implementations and configuration ranges stated with it, and extending the
set to a new compression algorithm, capability or data source is a matter of the same protocol, one
development panel and one pre-specified test, rather than of a new theory.

\end{document}